\documentclass[acmsmall]{acmart}

\usepackage{fontawesome5}
\usepackage{soul}
\usepackage{color}
\usepackage{amsthm}
\usepackage{tabularx}
\usepackage{enumitem}

\RequirePackage[normalem]{ulem} 
\RequirePackage{color}\definecolor{RED}{rgb}{1,0,0}\definecolor{BLUE}{rgb}{0,0,1} 

\newtheoremstyle{exmpl}
{2pt}
{2pt}
{}
{}
{\normalfont\scshape}
{ --- }
{}
{}

\usepackage{amsmath}
\DeclareMathOperator*{\argminA}{arg\,min}

\newenvironment{myquote}%
  {\list{}{\leftmargin=0.85in\rightmargin=0.85in}\item[]}%
  {\endlist}
\AtBeginDocument{%
  \providecommand\BibTeX{{%
    \normalfont B\kern-0.5em{\scshape i\kern-0.25em b}\kern-0.8em\TeX}}}

\begin{document}

\title{\textit{Periscope}: A Robotic Camera System to Support Remote Physical Collaboration}


  \author{Pragathi Praveena}
\affiliation{%
  \institution{Department of Computer Sciences, University of Wisconsin--Madison}
  \city{Madison}
  \state{WI}
  \country{USA}
  }
\email{pragathi@cs.wisc.edu}

  \author{Yeping Wang}
\affiliation{%
  \institution{Department of Computer Sciences, University of Wisconsin--Madison}
  \city{Madison}
  \state{WI}
  \country{USA}
  }
\email{yeping@cs.wisc.edu}

  \author{Emmanuel Senft}
\affiliation{%
  \institution{Idiap Research Institute}
  \city{Martigny}
  \country{Switzerland}
  }
\email{esenft@idiap.ch}

\author{Michael Gleicher}
\affiliation{%
  \institution{Department of Computer Sciences, University of Wisconsin--Madison}
  \city{Madison}
  \state{WI}
  \country{USA}
  }
\email{gleicher@cs.wisc.edu}

\author{Bilge Mutlu}
\affiliation{%
  \institution{Department of Computer Sciences, University of Wisconsin--Madison}
  \city{Madison}
  \state{WI}
  \country{USA}
  }
\email{bilge@cs.wisc.edu}

\renewcommand{\shortauthors}{Pragathi Praveena et al.}

\begin{abstract}
We investigate how robotic camera systems can offer new capabilities to computer-supported cooperative work through the design, development, and evaluation of a prototype system called \textit{Periscope}\footnote{Code and videos are available at \url{https://project-periscope.github.io} | Demo paper \cite{meng2023demo}}. With \textit{Periscope}, a local worker completes manipulation tasks with guidance from a remote helper who observes the workspace through a camera mounted on a semi-autonomous robotic arm that is co-located with the worker. Our key insight is that the helper, the worker, and the robot should all share responsibility of the camera view---an approach we call \textit{shared camera control}. Using this approach, we present a set of modes that distribute the control of the camera between the human collaborators and the autonomous robot depending on task needs. We demonstrate the system's utility and the promise of shared camera control through a preliminary study where 12 dyads collaboratively worked on assembly tasks. Finally, we discuss design and research implications of our work for future robotic camera systems that facilitate remote collaboration.
\end{abstract}





\begin{teaserfigure}
  \includegraphics[width=\textwidth]{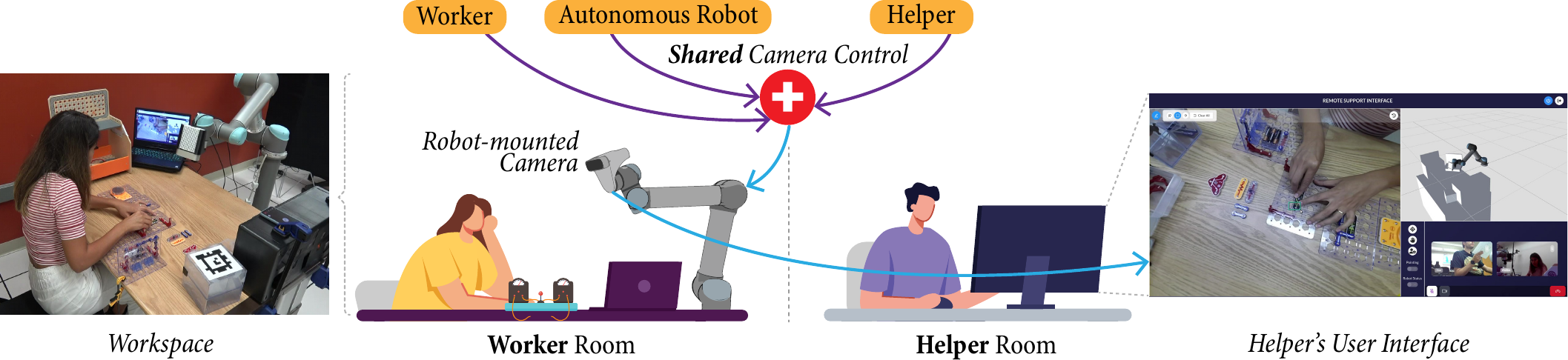}
  \caption{This paper introduces \textit{Periscope}, a robotic camera system that allows two people to collaborate remotely on physical tasks. With \textit{Periscope}, a local worker can complete an assembly task with guidance from a remote helper who views the workspace through a robot-mounted camera. We use a \textit{shared camera control} approach in which the worker, the helper, and the autonomous robot all contribute to camera control and we design a set of modes that uniquely combine inputs from these three sources to move the camera. The \textit{Periscope} system facilitates remote collaboration by providing the worker and the helper with shared visual information that enhances their verbal communication and coordination processes.
  }
  \label{fig:teaser}
\end{teaserfigure}

\received{January 2023}
\received[revised]{April 2023}
\received[accepted]{May 2023}

\maketitle
\section{Introduction}

Remote collaboration on physical tasks is valuable in scenarios such as experts assisting novices with manual assembly or repair tasks, particularly when it is inconvenient, time-consuming, or expensive to travel and assist someone in person. 
For example, a field technician might seek guidance from an expert to repair a wind turbine; an expert might provide training to car mechanics on how to repair a new engine model; or an astronaut might get help from ground control to maintain critical infrastructure on the space station.
Such scenarios typically involve a local ``worker'' manipulating physical artifacts with guidance from a remote ``helper.'' The helper views the workspace through one or more cameras, which may be fixed or movable. Ideally, the helper is able to observe various key sources of information including the worker, the task objects, and the environment \cite{kraut2003visual}. Additionally, the requirements on these views may change over the course of the task \cite{fussell2003helpers}. 
For example, the helper monitors the worker's actions during assembly, recognizes incorrect actions, and intervenes with new instructions, which requires looking at task objects while attempting to identify the component required for the next step. 
Finally, the helper may need to examine artifacts in the workspace from various angles, such as the interior of a drawer or the top of an object, and at varying levels of detail, such as a close-up view to see fine details or a wide-angle view to see more context \cite{kuzuoka1994gesturecam}. A core challenge for technologies that facilitate remote collaboration is \textit{providing the helper with diverse, informative, and task-relevant views}, which is not only critical for the helper to maintain awareness throughout the task but also for the helper and the worker to develop a shared understanding during the collaboration process \cite{fussell2000coordination}. 

The focus of recent research on remote collaboration in human-computer interaction (HCI) and computer-supported cooperative work (CSCW) has been on 
Virtual Reality approaches that give the helper the freedom to independently explore a reconstructed version of the worker's environment using a virtual camera (see \citet{schafer2021survey} for a review). These reconstructed workspaces can afford a high level of immersion and viewpoint flexibility, but they lack the dynamically changing details that are necessary for real-time collaboration. Other approaches involve cameras that stream directly from the real world, providing dynamic information from the task environment. However, these cameras are often limited to fixed viewpoints or viewpoints controlled solely by the worker (e.g., a head-worn camera), which can impede collaborative processes such as monitoring task status, observing worker's actions and comprehension, establishing joint attention, and formulating messages \cite{fussell2000coordination,fussell2003effects}. One potential solution that combines a high level of viewpoint flexibility and real-time, dynamic information through a live stream is the use of \textit{robotic cameras}. 

Modern \textit{collaborative robot}, or \textit{cobot}, platforms, augmented with cameras, can move with many degrees of freedom (DoF), supporting precise camera control for complex tasks and environments while maintaining safety for co-located human interaction. Despite their potential, such robots with high kinematic capabilities have rarely been utilized in robotic camera systems that support remote collaboration \cite{druta2021review}. Giving the helper direct control of a high-DoF robotic camera presents challenges related to designing control schemes that meaningfully link the user's inputs to robot movements. Controlling a low-DoF camera, such as a pan-tilt camera, is relatively simple with 2D controls that are directly mapped to the camera's movement. However, applying such methods to controlling a high-DoF camera in order to obtain precise views, such as looking into a drawer, is not straightforward to implement, as it requires mapping the helper's view intent to the camera's full 6-DoF pose (position and orientation). On the other hand, \textit{autonomous} camera control, particularly determining what the robot should be looking at at any given time during the collaboration, is an open question. In this work, we address the challenge of designing direct and autonomous camera control that enables the use of high-DoF robotic cameras for remote collaboration.

Prior literature suggests that both the helper and the worker may require control of the camera view at different points of the collaboration process, such as to provide guidance or ask questions \cite{mentis2020remotely, lanir2013ownership}. Therefore, a robotic system for remote collaboration must permit both the helper and the worker to modify the camera view. However, moving the camera is only a secondary activity for the helper and the worker, whose primary goal is to complete a collaborative physical task. Offloading some of the camera control to an autonomous robot can allow collaborators to devote more of their attention to the primary goal \cite{rae2014bodies}. Thus, the system should allow the robot to assume part of the workload of camera control by making autonomous adjustments to the camera view as needed while also allowing control of the view by the helper and the worker. We call this approach \textit{shared camera control} (based on a robot control paradigm called \textit{shared control} \cite{losey2018review}) and investigate how robotic camera systems can leverage this approach to offer new capabilities to CSCW through the design, development, and evaluation of a prototype system called \textit{Periscope} (see Figure \ref{fig:teaser}). 


The \textit{Periscope} system supports a worker in completing physical tasks with remote guidance from a helper who observes the workspace through a robot-mounted camera. The camera view is displayed on a screen interface for both the worker and the helper, enabling them to share
task-relevant visual information and develop a mutual understanding during the collaboration process. We design camera controls to empower both the helper and the worker to independently control the view depending on the needs of the task, but also allow the robot to assist and reduce their effort. Our system is centered around five design goals: (1) \textit{versatility} to support camera views for various task activities; (2) \textit{intuitivity} to simplify camera control for users through intuitive mappings and autonomous behaviors; (3) \textit{dual-user interactivity} to allow both the helper and the worker to modify the camera view; (4) \textit{congruity} to arbitrate user interactions and autonomous behaviors to reach consensus; and (5) \textit{usability} to support general communication and functional requirements. To balance these five design goals, we designed three modes that uniquely distribute camera control among the worker, the helper, and the autonomous robot. These modes serve as an initial point of inquiry for understanding the promise of shared camera control for facilitating remote collaboration. \textit{Through shared camera control, we tackle the challenge of simplifying control of a high-DoF robotic camera and providing users with diverse, informative, and task-relevant views.} 

We conducted a preliminary evaluation of the \textit{Periscope} system with 12 dyads in a lab study to understand how the system supports remote collaboration. During a 2-hour session, each dyad collaboratively worked on assembly tasks while physically located in separate rooms. From our analysis of recorded video data of the collaboration, we present use patterns for the system's features that illustrate the individual value of each mode and the rich interactions enabled by transitioning between the modes. Based on these results, 
we present reflections on our design goals and design implications for future robotic camera systems. Our work makes key contributions in four categories, \textit{Design (\S \ref{sec:design})}, \textit{System (\S\ref{sec:primitives}, \S\ref{sec:modes})}, \textit{Data (\S\ref{sec:results}, \S \ref{sec:reflection})}, and \textit{Recommendations (\S\ref{sec:implications})}:

\begin{enumerate}
    \item \textit{Design} --- the shared camera control approach and a set of design goals to realize this approach.
    \item \textit{System} --- \textit{Periscope}, a robotic camera system that is an instantiation of shared camera control.
    \item \textit{Data} --- empirical observations on system use and their contribution to the design goals.
    \item \textit{Recommendations} --- design implications for robotic camera-based CSCW systems.
\end{enumerate}

\section{Related Work}

In this section, we discuss prior research that identifies how \textit{shared visual context} is essential for successful collaboration. Then, we review systems that provide \textit{technological support for remote collaboration}, including robotic systems. Finally, we discuss existing \textit{control frameworks for cameras and robots} that we use to develop our shared-camera-control system. 

\subsection{Shared Visual Context}

During synchronous collaboration (both co-located and remote), verbal communication is the primary medium through which information is exchanged \cite{flor1998side, kraut2003visual}. \textit{Shared visual context} \cite{fussell2000coordination} or task-relevant visual information that the collaborators have in common augments verbal communication and improves collaborative outcomes. Findings from studies \cite{tang1991findings, daly1998some, flor1998side, kraut2003visual} suggest that people use the shared visual context for two coordination processes: \textit{situation awareness} and \textit{conversational grounding}. According to situation awareness theory by \citet{endsley1995measurement}, shared visual information helps people to establish an up-to-date mental model of the state of the task, the environment, and their partner, which can help the pair to plan future actions. According to conversational grounding theory by \citet{clark1981definite}, shared visual information supports verbal communication by providing an alternative and rich source of information that contributes to the development of a mutual understanding between collaborators, resulting in more efficient conversation. When collaborating on physical artifacts, shared visual information can particularly help the pair achieve \textit{joint attention} \cite{bruner1995joint}, where they have a shared focus on an object.

An example of remote collaboration from \citet{kraut2003visual} illustrates the use of shared visual context for situation awareness and conversational grounding. In this example, a helper guides a worker in adjusting the inclination of a bicycle seat during a repair task. The helper uses the shared visual context to gain situation awareness about the current state of the worker, the task, and the environment, allowing them to acknowledge the state (e.g., ``\textit{Cool}'') and plan next steps (e.g., ``\textit{next go on} and adjust it'' and ``angle the nose up \textit{a little bit more}''). The shared visual information also supports conversational grounding and joint attention, as both the helper and the worker use definite articles (e.g., ``\textit{the} bar'' and ``\textit{the} nose'') and deixis \cite{levinson2004deixis} (e.g., ``\textit{this} bar \textit{here}'') that require contextual information to be fully understood. The worker's verbal responses (e.g., ``\textit{Is that good?}'') and actions (e.g., \textit{Adjusts seat}) indicate their understanding of the helper's instructions and further contribute to the grounding process. 

\begin{myquote}
    \item \textit{\textbf{Helper:} Uh- next go on and adjust it so it’s parallel to the bar- the top}
    \item \textit{\textbf{Worker:} This bar here? Is that good?}
    \item \textit{\textbf{Helper:} Uh- angle the nose up a little bit more.}
    \item \textit{\textbf{Worker:} [Adjusts seat]}
    \item \textit{\textbf{Helper:} Cool.}
\end{myquote}

\subsection{Technological Support for Remote Collaboration}

Systems that support remote collaboration facilitate the sharing of visual context to enable effective cooperation and communication between users (see \citet{druta2021review} for a review). Our work draws from design choices made in other systems that support two remote users seeking to accomplish synchronous collaboration over physical artifacts. We divide the review of prior work into four categories --- (1) technologies for visual information capture, (2) technologies for visual information display, (3) technologies for communication cues, and (4) robotic systems for collaboration --- and provide examples of systems that fall under each category. We discuss relevant opportunities and challenges of different technologies for providing a shared visual context between collaborators.

\subsubsection{Technologies for Visual Information Capture:} Prior systems have used fixed-view cameras \cite{fussell2000coordination, fussell2004gestures, kirk2007turn}, head-mounted cameras \cite{fussell2003effects, johnson2015handheld, gupta2016do}, shoulder-mounted cameras \cite{kurate2004remote, piumsomboon2019on}, hand-held cameras \cite{gauglitz2012integrating, sodhi2013bethere, marques2022remote}, multiple cameras \cite{gaver1993one, fussell2003effects, rasmussen2019scenecam}, pan-tilt-zoom (PTZ) cameras \cite{ranjan2007dynamic, palmer2007annotating}, 360$^{\circ}$ cameras \cite{kasahara2014livesphere, piumsomboon2019on} or depth cameras \cite{adcock2013remotefusion, teo2019mixed} to capture visual information about the workspace. These sensors and additional eye-tracking or head-tracking technology may also be used for capturing information about the worker or helper \cite{gupta2016do, tecchia20123d, wang2019head} (see \citet{xiao2020usage} for a review). 

Early remote collaboration systems mostly relied on views from fixed cameras or worker-worn cameras (e.g., head-mounted or hand-held cameras), which the helper could not modify independently. These approaches can disrupt collaboration because the helper has to repeatedly interrupt the worker while they are performing task-related activities and direct them to change the view. Recent research focuses on enabling the helper to independently view the workspace via remote control of physical cameras (e.g., PTZ cameras) or virtual cameras (e.g., in 3D reconstructed workspaces). Although this approach increases the system's complexity, granting the helper control over the view enables them to have diverse and independent views of the workspace.

\subsubsection{Technologies for Visual Information Display:} Prior systems have used 2D views \cite{fussell2000coordination, fussell2004gestures, kirk2007turn, ranjan2007dynamic}, 3D views \cite{adcock2013remotefusion, gauglitz2014touch}, 360$^{\circ}$ views \cite{kasahara2014livesphere, lee2017mixed}, Virtual Reality (VR) \cite{tecchia20123d}, Augmented Reality (AR) \cite{sodhi2013bethere, johnson2015handheld, gupta2016do}, Mixed Reality (MR) \cite{oda2015virtual, thoravi2019loki, piumsomboon2019on, teo2019mixed, bai2020user}, and projected AR \cite{gurevich2012teleadvisor, machino2006remote, speicher2018360anywhere} for the display of shared visual information (see \cite{schafer2021survey} for a review on VR, AR, and MR systems). AR and projected AR are typically used for situated information display to the worker who handles the physical artifacts. 

VR, AR, and MR solutions provide users with a highly immersive experience. However, high-quality virtual reconstructions can be difficult to update in real-time, require significant bandwidth, and may lack the fine and dynamically changing details that are necessary for many physical tasks. In such scenarios, live 2D or 360$^{\circ}$ video may be superior. Additionally, these approaches can be mixed together \cite{teo2019mixed} to leverage the benefits and reduce the drawbacks of each approach. 

\subsubsection{Technologies for Communication Cues:} The primary communication channels in remote collaboration systems are typically visual and verbal. Additionally, \citet{fussell2004gestures} recommend that gestures used by helpers should be captured by collaboration systems to support referential communication. These gestures may be captured through vision-based or IMU-based hand tracking \cite{bai2020user, tecchia20123d, teo2019mixed} or specified through annotations \cite{fussell2003assessing, kim2013comparing, gauglitz2014touch}. The gestures are then relayed to the worker by overlaying graphics on the shared view. This includes 2D graphic overlays on 2D views, 3D graphic overlays in MR, and projections onto the physical world in projected AR. Prior works have found improved collaborative outcomes when gestures are combined with visualizations of the helper's eye gaze \cite{bai2020user, akkil2016gazetorch}, the worker's eye gaze \cite{gupta2016do, sasikumar2019wearable}, or viewing direction \cite{higuch2016can,otsuki2018effects}. Other interesting communication cues include virtual replicas of task objects \cite{oda2015virtual} or human avatars to provide non-verbal cues in MR \cite{piumsomboon2018mini}. 

\subsubsection{Robotic Systems for Collaboration:} Prior work has explored how robots can facilitate remote collaboration. These works mostly focus on enabling the helper to control a robot-mounted camera in low-DoF settings and do not fully explore the possibility of robot autonomy, control of the robot by the worker, or how to manage the complexity of sharing camera control among the helper, the worker, and the robot. Thus, they do not leverage the full potential of a robotic platform for the formation of a co-constructed visual context. Our research aims to address this gap.

Early work by \citet{kuzuoka1994gesturecam} demonstrated that granting the helper independent control of a 3-DoF robotic camera enabled the helper to explore 3D workspaces and examine physical artifacts from various angles. More recently, \citet{feick2018perspective} used a robotic arm to reproduce orientation manipulations on a proxy object at a remote site. While this solution improves the user's spatial understanding of the object, the method is hard to scale beyond one object. \citet{gurevich2012teleadvisor} and \citet{machino2006remote} designed systems that used a robot-mounted camera and projector (to capture the workspace and project on top of it), and showed that the mobility of the system improved collaborative outcomes. These systems allowed the helper to control the robot, but there was no exploration of robot autonomy or worker control of the robot. \citet{sirkin2012consistency} and \citet{onishi2016embodiment} explored the use of a robotic arm to display gestures such as pointing to and touching remote objects but not to capture any visual information about the workspace.  

Telepresence robots make up a special case of robots designed to support collaboration by emulating face-to-face communication in a remote setting. The prototypical telepresence robot is a screen on wheels that is roughly human-sized in height with a camera and microphone. An interface will typically allow the remote user (in rare cases such as \cite{rae2013body}, the local user) control over the movement of the telepresence robot and the positioning of the cameras. These robots improve collaborative outcomes through the provision of a physical embodiment \cite{rae2013body, kuzuoka2000gestureman} that enhances the feeling of presence or ``being there'' for the remote user, and improves the local user's sense of the remote user's presence \cite{choi2017identity}. 
There is a rich literature (e.g., \cite{kratz2016immersed, kiselev2014effect, rae2013influence, rae2015framework, vartiainen2015expert, johnson2015can, stahl2018social}) on how telepresence robots and interfaces should be designed to support communication between remote users. However, these design choices are constrained by the anthropomorphic treatment of the robot as the remote user's surrogate. Non-anthropomorphic form factors, such as a robotic arm, offer a different design space of interaction techniques that can leverage robot autonomy and control by the co-located user to support remote collaborative work. 

Researchers have also explored other form factors for telepresence robots, such as drones \cite{sabet2021designing, zhang2019lightbee} or tabletop robots \cite{adalgeirsson2010mebot, sakashita2022remotecode}. While these systems are typically designed for interpersonal communication, some recent works \cite{villanueva2021robotar, li2022asteroids} have addressed the use of tabletop robots for supporting collaboration in remote physical tasks. \citet{villanueva2021robotar} designed a tabletop robot that can be controlled by a remote instructor to provide in-situ advice on basic electrical circuitry to students. \citet{li2022asteroids} used a swarm of tabletop robots with cameras to allow several remote persons to view physical skill demonstrations by an instructor. The remote audience members can view the workspace through automatic and manual navigation of the robots and the instructor can physically move the robots for camera repositioning. These systems are advancing the possibilities of robotic platforms.  However, the workspaces in these prior works are relatively level and uncluttered where low-DoF tabletop robots are adequate. Our work leverages high-DoF robot arms that can be used for precise camera control to support scenarios with clutter and complex geometry and allow remote users to achieve specific views such as the interior of a drawer or the top of an object. We draw from state-of-the-art camera and robot control frameworks (discussed next in \S \ref{sec:control}) to limit the effort needed to control of high-DoF robots in 3D environments.

\subsection{Control Frameworks for Cameras and Robots}
\label{sec:control}

Relevant to the goals of this paper is prior literature on state-of-the-art camera control methods and control mechanisms for robots when humans and robots work in a collaborative ecosystem, which can inform the design of a robotic camera to support remote collaboration between individuals.

\subsubsection{Camera Control:} \citet{christie2008camera} describe various challenges associated with camera control. Designing control schemes for direct control of the camera by the user is challenging because users can find it difficult to deal simultaneously with all of the camera's degrees of freedom. Consequently, control schemes must provide mappings that meaningfully link the user's actions to the camera parameters. On the other hand, it is also challenging to partially or fully automate camera movement because the geometric specification of the camera pose needs to result in a semantically meaningful view for the user. Thus, our work draws from various manual and automated camera control techniques such as visual servoing \cite{chaumette2016visual, hutchinson1996tutorial}, through-the-lens camera control \cite{gleicher1992through}, assisted camera control in virtual environments \cite{christie2008camera}, and automatic cinematography \cite{christianson1996declarative} to make it easier for the helper and the worker to influence the shared view.  

Visual servoing \cite{hutchinson1996tutorial} is a robot control method using features extracted from vision data (from a camera) to define a target pose for the robot and determine how the robot should move. Through-the-lens camera control \cite{gleicher1992through} is a technique where a camera view is specified through controls in the image plane, essentially mapping visual goals to camera movements. Methods that provide assisted camera control in virtual environments \cite{christie2008camera} use knowledge of the environment to assist the user with camera control. For example, if the camera maintains a fixed distance around an object when it is being inspected, it results in the camera orbiting around the object in response to user inputs. Techniques for automatic cinematography \cite{christianson1996declarative} enable automatic tracking of a person (or their face or hands) to keep them in view. This has been utilized both in research prototypes of remote collaboration systems, for instance, hand tracking in \citet{ranjan2007dynamic}, and commercial video conferencing products such as Apple Center Stage\footnote{\url{https://support.apple.com/en-us/HT212315}} and Lumens Auto Tracking Camera.\footnote{\url{https://www.mylumens.com/en/Products/12/Auto-Tracking-Camera}}

In his paper on remote collaboration systems, \citet{gaver1992affordances} asserts that unless the cost of gaining additional information is low enough, it will not seem worth the additional effort for users. Our work is guided by this idea of allowing both the helper and the worker to move the camera with low cognitive and physical costs.


\subsubsection{Shared Control:} Shared control is a robot control paradigm where robot behavior is determined by multiple different agents (agents may be human or robotic) working together to achieve a common goal \cite{dragan2013policy, losey2018review}. This paradigm is also referred to as collaborative control \cite{macharet2012collaborative} or mixed-initiative human-robot interaction \cite{jiang2015mixed}. One key aspect of shared control systems is the design of \textit{arbitration} or the division of control among agents when completing a task. \citet{losey2018review} suggest that agents assume different roles during task execution. For example, the human agent controls larger robot motions while the robotic agent controls finer robot positioning. Additionally, these roles can shift over time. Thus, arbitration in shared control should allow all agents to contribute and change the type of contribution they make over time. This idea of dynamic roles is central to the arbitration mechanisms we design for our shared camera control system.

Some prior works in the robotics literature \cite{abi2016visual, nicolis2018occlusion, rakita2019remote, senft2022drone} use shared control-based methods for control of a robot-mounted camera to give the remote user a view of another robotic arm used for remote manipulation. There is no local worker in such scenarios, and hence these solutions do not consider the needs of a collaboration setting. In our work, we use an optimization-based shared control method similar to \citet{rakita2019remote, rakita2021collisionik} with adaptations for remote human collaboration where the robot augmented with a camera is co-located with a worker completing manual tasks.




\section{The Periscope System}

In this section, we introduce the design and implementation of the \textit{Periscope} system, which supports remote collaboration by leveraging shared camera control. Our approach provides collaborators with a low-effort means of shaping the shared visual context using a robot-mounted camera. 

\subsection{Design Goals}
\label{sec:design}

Based on prior literature and early feasibility studies, we identified five high-level design goals that guided our design process for developing a robotic camera system to support remote collaboration. The first four design goals are related to the core functionality of camera control: \textit{versatility}, \textit{intuitivity}, \textit{dual-user interactivity}, and \textit{congruity}. The final design goal, \textit{usability}, is related to system functionality that is peripheral (but crucial) to camera control.


\subsubsection{\textbf{Versatility:} Support camera views for various task activities} The visual information necessary for users to maintain awareness and ground their conversation varies depending on task activities (e.g., searching, assembling, inspecting, or correcting). Hence, the system should support these dynamic needs and provide the helper with access to diverse sources of visual information in the workspace (including the worker's face and actions, task objects, and the environment) from various angles and in varying levels of detail. This information should be shared with the worker, so that the pair can use the shared visual context to monitor comprehension, plan future actions, achieve joint attention, and communicate efficiently.
\label{goal:one}

\subsubsection{\textbf{Intuitivity:} Simplify camera control for users through intuitive mappings and autonomous behaviors} Camera movement in response to user input should be clear and familiar. The usage of autonomous behaviors should facilitate the user's ability to provide high-level specifications while the robot handles the low-level details of how to achieve those specifications. Autonomous behaviors should also be used without requiring human input for aspects of robot control that may be difficult and non-intuitive for users. Camera control should be as non-intrusive as possible (i.e., not interrupt the collaboration process).
\label{goal:two}

\subsubsection{\textbf{Dual-user Interactivity:} Allow both the helper and the worker to modify the camera view} Both the helper and the worker require control of the camera view at different points of the collaboration process to gather or exchange information. Hence, they should be able to independently control the camera. The camera control functionality should consider the specific modalities supported by the users' locations (the helper is remote, the worker is co-located with the robot). 
\label{goal:three}

\subsubsection{\textbf{Congruity:} Arbitrate user interactions and autonomous behaviors to reach consensus} The camera's movement can be controlled by three sources of input with potentially conflicting interests: the helper, the worker, and the autonomous robot. Hence, there is a need for arbitration of control authority between the three entities in order to determine which input has priority at what times and to prevent any conflicts. Arbitration should allow all agents (human and robotic) to contribute and change the type of contribution they make over time.
\label{goal:four}

\subsubsection{\textbf{Usability:} Support general communication and functional requirements} The system should support verbal communication since it is a key medium through which information is exchanged during collaboration. Additionally, the system should try to support non-verbal communication (e.g., gestures, visual annotations), especially to facilitate deictic referencing. Finally, users should be informed of the system's internal state in a non-intrusive manner as necessary.
\label{goal:five}

\subsection{System Overview}

\begin{figure}[!h]
    \centering
    \includegraphics[width=\textwidth ]{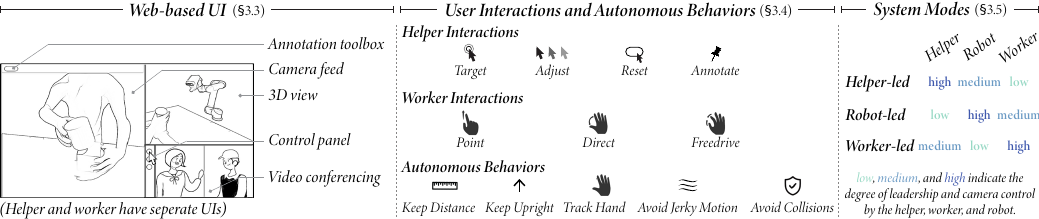}
    \caption{The \textit{Periscope} system consists of three components: user interfaces for the helper and the worker, a set of helper interactions, worker interactions, and autonomous robot behaviors to support establishing a shared visual context, and system modes that arbitrate user interactions and autonomous behaviors in real-time, resulting in camera motion. \textit{(Left)} Each user interface includes a \textit{camera feed} that displays the live video feed from the robot-mounted camera, accepts mouse input commands, and can be annotated using the \textit{annotation toolbox}; a \textit{3D view} that shows a visualization of the robot and its surroundings; a \textit{video conferencing} panel for verbal communication between the helper and the worker; and a \textit{control panel} for mode selection. \textit{(Center)} Helpers use the screen interface to interact with the system via mouse input commands on the camera feed or the control panel. Workers move the camera by engaging directly with the robot arm using physical contact and gestures. Autonomous robot behaviors augment helper and worker interactions by supporting the aspects of camera control that are difficult and non-intuitive for users. \textit{(Right)} \textit{Helper-led Mode}, \textit{Robot-led Mode}, and \textit{Worker-led Mode} are three system modes that arbitrate user interactions and autonomous behaviors differently, offering varying degrees of control to both users for low effort. Users can select from the three available modes via the control panel to support their current needs. 
    }
    \label{fig:system}
    \vspace{-0.1in}
\end{figure}

We developed a prototype system based on the design goals stated in \S \ref{sec:design}. As shown in Figure \ref{fig:system}, the \textit{Periscope} system consists of three components: (1) \textit{user interfaces} for the helper and the worker, (2) a set of \textit{helper interactions}, \textit{worker interactions}, and \textit{autonomous robot behaviors} to support establishing a shared visual context, and (3) \textit{system modes} that arbitrate user interactions and autonomous behaviors in real-time, resulting in camera motion.
To maintain brevity, we include technical details of the implementation in Appendix \ref{sec:apptechnical}, and present high-level descriptions of the system below.

\subsection{User Interfaces}

We designed interfaces for the helper and the worker based on the goals of \textit{versatility}, \textit{dual-user interactivity}, and \textit{usability}. In our remote collaboration setup, the worker is co-located with a robot arm augmented with an RGB-D (color + depth) camera, which is used to capture information about the worker and the workspace. The robot arm has six degrees of freedom, which is the minimum required for reaching any position and orientation in a 3D workspace. The helper is in a remote location and views the workspace on a 2D screen interface\footnote{Although a head-mounted display is a viable option, its interplay with robotic technology for collaboration is unclear and we chose a more established display technology for this work.} through a live video from the RGB camera and a simulated 3D view. The worker can view the visual information shared with the helper on a 2D screen interface that is similar to the helper's interface. 

The screen interface consists of four panels. The \textit{camera feed} panel shows the live video feed from the robot-mounted camera. The camera feed accepts input commands (through mouse clicks and drags) that can be used for camera control. Additionally, the camera feed can be annotated (with a pin, a rectangle, or an arrow) using the annotation toolbox to support referential communication. Overlays on the camera feed provide visual feedback for input commands and annotations. The \textit{3D view} panel shows a simulated visualization of the robot and its surrounding objects, and updates their states in real-time. The \textit{video conferencing} panel allows verbal and visual communication between the helper and the worker. The \textit{control panel} provides options related to camera control (including mode selection), in addition to those accessible through the camera feed. 

\subsection{User Interactions and Autonomous Behaviors}
\label{sec:primitives}

We designed interactions for the helper and the worker that are augmented by autonomous robot behaviors based on the goals of \textit{versatility}, \textit{intuitivity}, and \textit{dual-user interactivity}. Below, we describe helper and worker interactions, and autonomous behaviors afforded by the \textit{Periscope} system.

\begin{figure}[!b]
    \centering
    \includegraphics[width=\textwidth ]{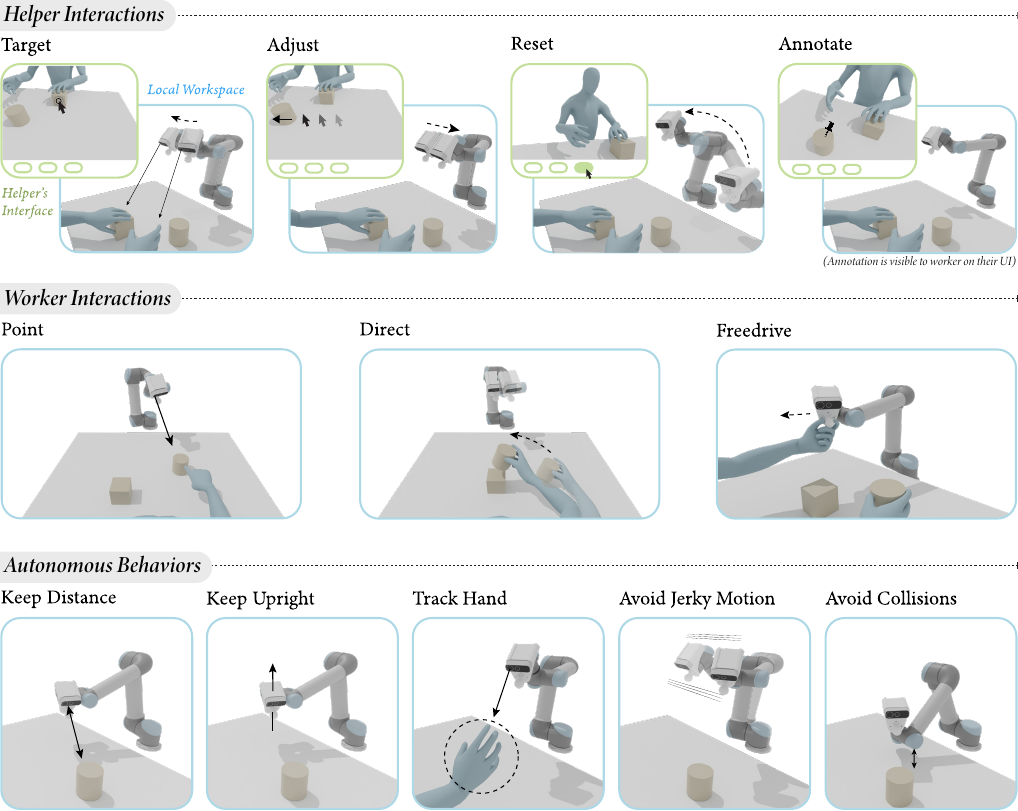}
    \caption{The \textit{Periscope} system supports a variety of interactions for the helper and the worker, assisted by autonomous robot behaviors. Images with a blue border represent the local workspace, where both the robot and the worker are present. Images in the top row with a green border depict the remote helper's UI which enables the helper to interact with the system through mouse input on the camera feed or the control panel. Worker interactions leverage the worker’s proximity to the robot, allowing them to move the camera directly using gestures and physical contact with the robot arm. Autonomous behaviors focus on geometric (rather than semantic) qualities of the view, which are challenging for humans but feasible for robots to achieve.}
    \label{fig:interactions}
\end{figure}

\subsubsection{Helper Interactions} Helpers use the screen interface to interact with the system via mouse input commands on the camera feed or the control panel (see Figure \ref{fig:interactions} for illustrations). 

\paragraph{Target:} The helper can change the viewing direction of the camera by setting a target through a mouse right-click on the camera feed. The camera will point to the specified target such that the target is positioned near the center of the camera's field of view. Visual feedback is displayed on the camera feed in the form of a dot corresponding to the target.

\paragraph{Adjust:} The helper can move the camera in a specific direction based on directional inputs, in order to make adjustments to the view. Through mouse scroll, the helper can move the camera forward or backward in the direction that the camera is currently pointing at, allowing them to see more detail or context depending on task needs. Other directional inputs (mouse left-click + drag up/down/left/right) will result in different behaviors depending on whether \textit{Target} was engaged prior to \textit{Adjust}. If the camera is pointing at a target, then it will perform orbital rotations around the target point.  If there is no target, the camera will linearly move in the direction specified by the helper. We will refer to the three behaviors as \textit{zoom} (move forward/backward), \textit{orbit} (orbital rotation), and \textit{shift} (linear movement) in the remainder of the paper. Visual feedback is displayed on the camera feed in the form of arrow overlays. 

The \textit{Target} + \textit{Adjust} interactions attempt to replicate the behavior of orbital cameras, which are widely used in virtual environments and suitable for object-focused applications.

\paragraph{Reset:} The helper can move the camera from its current state to a pre-defined configuration by clicking a button on the GUI. The pre-defined configuration is identical to the initial configuration that the system enters at startup.

\paragraph{Annotate:} The helper can overlay graphics on the camera feed for referential communication with the worker. The helper can drop a pin to indicate a point, draw a rectangle to indicate an object or an area, or place an arrow to indicate a direction. When the helper engages \textit{Annotate}, the robot motion is automatically stopped to freeze the scene during the interaction. 

\subsubsection{Worker Interactions} Workers move the camera by engaging directly with the robot arm using physical contact and gestures recognized by the camera (see Figure \ref{fig:interactions} for illustrations). These interactions leverage the worker's proximity to the robot.

\paragraph{Point:} The worker can specify the target that the camera should look at using a pointing gesture. The camera will point to the target indicated by the worker's index finger. Additionally, the camera moves to a predetermined distance from the target (40 cm in our system) so that the target is visible in adequate detail in the view. The \textit{Point} interaction is intended to be a discrete input from the worker in contrast with the next interaction, \textit{Direct}, which is intended to be a continuous input.

\paragraph{Direct:} The worker can continuously influence the camera's viewing direction by moving their hand, which can be set as the camera's target. This interaction is augmented by the \textit{Track hand} autonomous behavior, allowing the worker to guide the view without touching the robot. 

\paragraph{Freedrive:} The worker can manually move the robot-mounted camera into desired poses by manipulating the robot joints. The robot arm responds to applied forces, moving in the direction of push or pull from the worker.

\subsubsection{Autonomous Behaviors} Autonomous robot behaviors augment helper and worker interactions by supporting the aspects of camera control that are difficult and non-intuitive for users. These behaviors are typically related to geometric (rather than semantic) qualities of the view, which are challenging for humans but feasible for robots to achieve (see Figure \ref{fig:interactions} for illustrations).

\paragraph{Keep distance:} The robot keeps the camera at a specific distance from the target point. This augments the \textit{Adjust} interaction to enable orbital motions and \textit{Point} interaction to keep the target visible in adequate detail. For the \textit{Adjust} interaction, the distance is determined as the distance between the camera and the target at the time the helper engages adjustment through orbit. 


\paragraph{Keep upright:} The robot maintains the camera in an upright direction and prevents any roll (i.e., rotation along the front-to-back axis of the camera). This is typically done during assisted control of virtual cameras to avoid users from being disoriented.

\paragraph{Track hand:} The robot detects the worker's hand (implemented using MediaPipe \cite{lugaresi2019mediapipe}) and automatically points the camera at the hand. This augments the worker's \textit{Direct} interaction.

\paragraph{Avoid jerky motion:} The robot avoids large and jittery camera motions, and promotes safe operation of the robot by maintaining the robot's range of motion within the limits of the joints. This is essential because the view needs to be stable and not disorienting for viewers. 

\paragraph{Avoid collisions:} The robot automatically avoids collisions with itself and objects in the environment, including the worker. This can be particularly beneficial for the helper, as they may face challenges in avoiding collisions when controlling the robot. Helpers have limited awareness of potential collisions as they only see the workspace from the camera's point of view and may not be aware of the placement of the robot arm's joints and obstacles outside the camera's field of view. 

\subsection{System Modes}
\label{sec:modes}

We developed system modes that arbitrate the user interactions and the autonomous behaviors described in \S \ref{sec:primitives} based on the design goal of \textit{congruity}. To achieve effective arbitration, these interactions and behaviors should work in harmony to generate camera motion. Additionally, there is a trade-off between the degree of control users desire and the amount of effort they are willing to put in. Ideally, users should have high control over the view with low effort, but this is difficult to achieve. Through an iterative design process, we developed three modes that we believe offer varying degrees of control to both users for low effort. Users can select from the three available modes via the control panel to support their current needs. The three modes are: \textit{Helper-led Mode}, \textit{Robot-led Mode}, and \textit{Worker-led Mode}. Each mode is led by one of the three agents, while the other two exert less influence. This leader-follower approach makes the arbitration of control authority more tractable. After arbitration, a motion generation algorithm (detailed in Appendix \ref{sec:apptechnical}) moves the robot's joints to achieve the desired camera pose.



\subsubsection{Helper-led Mode}
This mode is led by the helper who can specify the camera's viewing direction by setting a target and adjusting the view through zoom and orbit. The worker has some influence over the camera's viewing direction via a pointing gesture that can be accepted by the helper. Meanwhile, the robot assists to ensure safe and high-quality camera control by keeping the camera at a constant distance during orbit, keeping the camera upright, avoiding jerky camera motions, and avoiding robot collisions. This mode gives the helper substantial control of the camera. The helper can freely move the camera to observe the workspace, and the worker can participate by pointing to a location of interest. 

\subsubsection{Robot-led Mode}
This mode is led by the robot which tracks the worker's hand while the helper can adjust the view through zoom and orbit. Similar to the \textit{helper-led mode}, the robot also assists by ensuring safe and high-quality camera control. This mode is designed to reduce the workload of camera control for both the helper and the worker. In this mode, the worker can focus on completing the physical task, while the robot captures the worker's activity in the workspace and maintains the worker's hand in the camera view. This mode allows the helper to focus on providing guidance without the need to control the camera to monitor the worker's behaviors. 

\subsubsection{Worker-led Mode}
This mode is led by the worker who can set the camera's pose through freedrive (manually moving the robot) while the helper can adjust the view through zoom and shift (not orbit, since no target is set prior to adjust). This mode gives the worker substantial control of the camera. In fact, robot assistance for safe and high-quality camera control is disabled when the worker moves the camera. We wanted to include a mode in which autonomous behaviors exert less influence, giving more control to the co-located worker to handle these aspects of camera control. However, when the helper adjusts the view, the robot provides moderate assistance by avoiding jerky camera motions and robot collisions. The worker can use this mode to present visual information to the helper, and the helper can adjust the camera pose for a better viewpoint.

\section{User Study}

We conducted an evaluation of the \textit{Periscope} system in a lab study with 12 dyads to understand the utility of our shared camera control approach for remote collaboration.



\subsection{Study Design}

We recruited dyads to participate in our user study. One participant was assigned the role of \textit{worker} and had access to the physical workspace but no instructions on how to carry out the assembly. The other participant was assigned the role of \textit{helper} and was tasked with guiding the worker using the instruction manuals that we provided. During the study, participants collaboratively worked on a training task and a main task, which were both assembly tasks from scientific play kits. These kits were sufficiently complex to make completion without instructions challenging, and their components were sturdy enough to withstand frequent handling by participants. 

The training required for participants to be able to successfully interact with the system was unclear initially. Thus, we iteratively developed a training protocol based on early participant observations and feedback. In our final training protocol, one experimenter guided both participants simultaneously through completion of a training task for around an hour. The training protocol consisted of $\sim$70 steps that introduced all the functionalities available in the \textit{Periscope} system and allowed dyads to try them out. Experimenters solicited feedback throughout the training process to encourage participants to reflect on their use of the system's functionalities. We also made adjustments to the main task protocol based on participant feedback. Below, we describe the final protocol that we developed and clarify the variations of the protocol followed by each dyad in \S \ref{sec:participants}.

\begin{figure}[!t]
    \centering
    \includegraphics[width=\textwidth ]{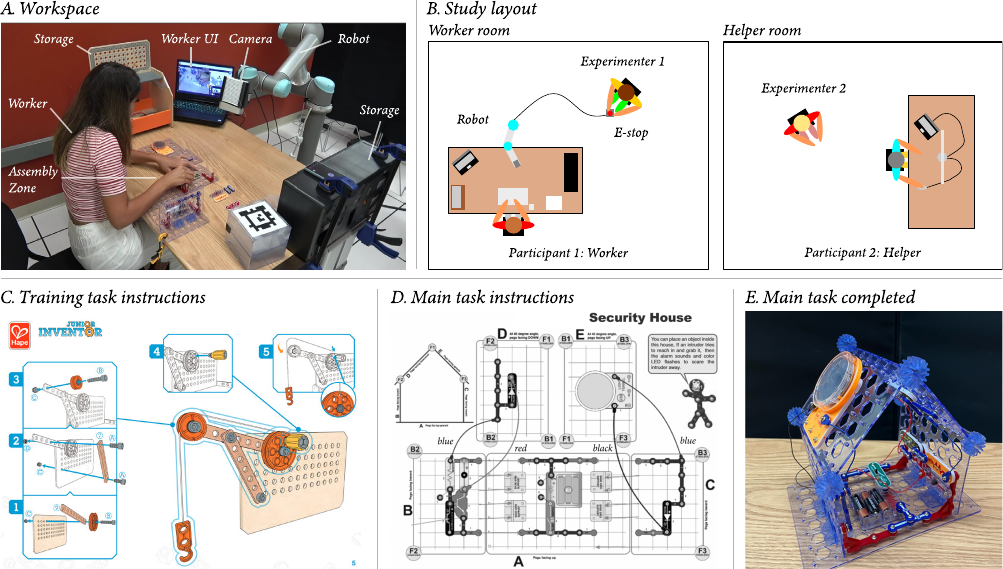}
    \caption{A. One participant (the worker) was located in the same physical space as a robot arm with access to the workspace but no instructions on how to carry out the assembly. The other participant (the helper) was tasked with guiding the worker remotely using the \textit{Periscope} interface. B. The study took place in two rooms with accompanying experimenters. C. Instructions for the training task. D. Instructions for the main task. E. The completed structure that participant dyads were tasked with building collaboratively.
    }
    \label{fig:study}
\end{figure}

\subsection{Tasks}

The training task was to construct a pulley system from a toy workbench kit\footnote{Workbench Kit: \url{https://a.co/d/2zLeQoV}} (see Figure \ref{fig:study}C). The helper was provided with the instruction manual that came with the kit. The workbench comprised of a peg board for assembling the pulley system and a toolbox with storage space. The workbench was clamped to the table to be immobile. The components required for the task were distributed between the toolbox and another storage unit located away from the workbench.

The main task was to build a 3D illumination circuit project\footnote{SnapCircuit Kit: \url{https://a.co/d/34trhAd}} (see Figure \ref{fig:study}E). The helper was provided with a black and white copy of the instruction manual that came with the kit (see Figure \ref{fig:study}D). Some visual features on the manual were deliberately blurred to ensure sufficient task complexity. Participants were tasked with building 3D circuits for a lighting and alarm system in a security house, which consisted of a base grid, two wall grids, and two roof grids. When participants began the task, the house was partially built, with one wall grid connected to the base grid and completed circuitry on the roof grids. Participants had to evaluate the partially assembled house, attach missing components to the existing wall grid, attach and build circuitry on the other wall grid and base grid, attach the roof grids, and finish the wiring.

\subsection{Study Setup}

The study took place in two rooms: the worker room and the helper room (see Figure \ref{fig:study}B). The participant who was assigned the role of the \textit{worker} was located in the same physical space as a robot arm and Experimenter 1 (see Figure \ref{fig:study}A). The worker sat behind a desk, facing the robot that was within arm's reach. The experimenter was nearby, observing the room and had access to the robot's emergency stop button. The worker viewed the screen interface on a laptop and could provide inputs to the interface using a mouse or directly interacting with the robot arm. A workbench kit (from the training task) was adjacent to the laptop. A large immobile organizer and a small movable organizer on the opposite side of the desk provided storage for various task components. The components for the training and main tasks were stored together. The participant used the laptop's camera and microphone for video-conferencing through the interface.

The participant who was assigned the role of the \textit{helper} was located in a different room than the worker, accompanied by Experimenter 2. The helper sat behind a desk with access to a laptop, a monitor, and a mouse for interacting with the interface. The participant used the laptop's camera and microphone for video-conferencing through the interface.

\subsection{Procedure}
\label{sec:procedure}

This protocol was approved by the Institutional Review Board of University of Wisconsin--Madison. We conducted the study in two rooms in a university laboratory. Each session lasted approximately two hours. The first author (Experimenter 1) facilitated the study along with another experimenter (Experimenter 2). Both experimenters individually described the study to the participant and obtained written consent. Experimenter 1 introduced the interface and the physical robot to the worker before connecting to the video conference. In parallel, Experimenter 2 provided the same introduction for the interface and described the virtual robot in the 3D view of the interface to the helper before joining the video conference. Experimenter 1 guided both participants simultaneously through the training protocol. The experimenter familiarized participants with the workspace, outlined the task flow, and initiated test interactions in each mode. Participants were then asked to use their cheat-sheet, which listed all system features, to summarize what they had learned. 

During the training task, the helper was encouraged to locate the necessary component, ask the worker to pick it up, and provide assembly instructions to the worker. Participants were asked to gather the required components for each step (steps are listed in the manual shown in Figure \ref{fig:study}C) using a certain mode, and then assemble the components using an alternate mode. They were then asked to reflect on their experiences. We repeated this procedure for all the modes, allowing participants to gain experience with each mode for different task activities. We allowed participants to complete the final step of the task using any combination of modes they preferred. Participants were finally asked to reflect on their overall experience in all modes. If a participant avoided using a feature or used it wrongly, the experimenter reminded or corrected them regarding the system's functions. The training task took approximately 60 minutes.

The video conferencing link was disabled before Experimenter 1 went to the helper room and explained the procedure and goals for the main task to the helper. The helper was shown a completed model of the security house and had the opportunity to interact with it. Then, Experimenter 1 partially disassembled the house and set it up on the worker's table. The video conference was then resumed, and participants were given high-level directions on which panel to assemble. Participants were given the flexibility to use any (or none) of the system's modes and other features they found suitable for completing the task. We used this approach because we wanted to gain insights into how people utilized the system in a relatively realistic setting. Participants had 45--60 minutes to collaboratively work on the main task.

\subsection{Participants}
\label{sec:participants}

For the user study, we did not target any particular user group, as the scientific play kits did not require specific expertise and the system was designed for use by individuals unfamiliar with robots. We recruited 24 participants from the University of Wisconsin--Madison's campus community. Demographic information for one dyad was not collected. The remaining participants (8 female, 14 male) were aged between 18 and 69 years ($M=26.32$, $SD=10.48$). Participants had various educational backgrounds, including urban design, business, physics, engineering, and computer science. Two participants reported prior participation in robotics studies, and one dyad consisted of individuals who knew each other prior to the experiment. 

The first four out of the twelve dyads underwent a less rigorous training protocol and performed a different (but similar) task from the kit. While these four dyads were important for establishing the final protocol, we excluded them from our dataset for analysis as they followed a different procedure compared to the other eight dyads. The next two dyads followed the procedure described in \S \ref{sec:procedure}, with the only difference being that the helpers were not shown the completed model of the security house before starting the main task. The remaining six dyads strictly followed the procedure described in \S \ref{sec:procedure}. To ensure consistency and comparability within the dataset for the analysis described below, we used data from the last eight dyads that followed a similar procedure.

\subsection{Analysis}
\label{sec:analysis}
During the study, we screen-recorded the helper's interface and recorded the workspace (including the worker and the robot), resulting in $\sim$36 hours of video recordings (12 dyads*2 users*$\sim$1.5 hours). The dataset for our analysis consists of $\sim$12 hours of video recordings from eight dyads during the main task (8 dyads*2 users*$\sim$0.75 hours). This is rich multi-user, multi-modal data containing dialogue, interactions with the system, worker actions, and camera motions. 

We analyzed the videos using a deductive thematic analysis approach \cite{braun2012thematic}. The first author and a study team member were familiarized with the data through conducting all study sessions and transcribing participant conversations. Both the first author and the study team member coded all helper videos (screen-recordings) to identify relevant conversations and patterns, conducted meetings to discuss their codes and resolve any conflict, and distilled the codes in a codebook. The first author then coded all worker videos and refined the codes. Resulting themes were refined by the first author and reported after discussions with the remaining authors. ELAN\footnote{\url{https://archive.mpi.nl/tla/elan}\\ Max Planck Institute for Psycholinguistics, The Language Archive, Nijmegen, The Netherlands}\cite{crasborn2008enhanced} was used for video coding, and the collaborative whiteboard app Miro\footnote{\url{https://miro.com/}} was used to refine thematic findings.

\section{Results}
\label{sec:results}

\begin{figure}[!h]
    \vspace{-0.05in}
    \centering
    \includegraphics[width=\textwidth ]{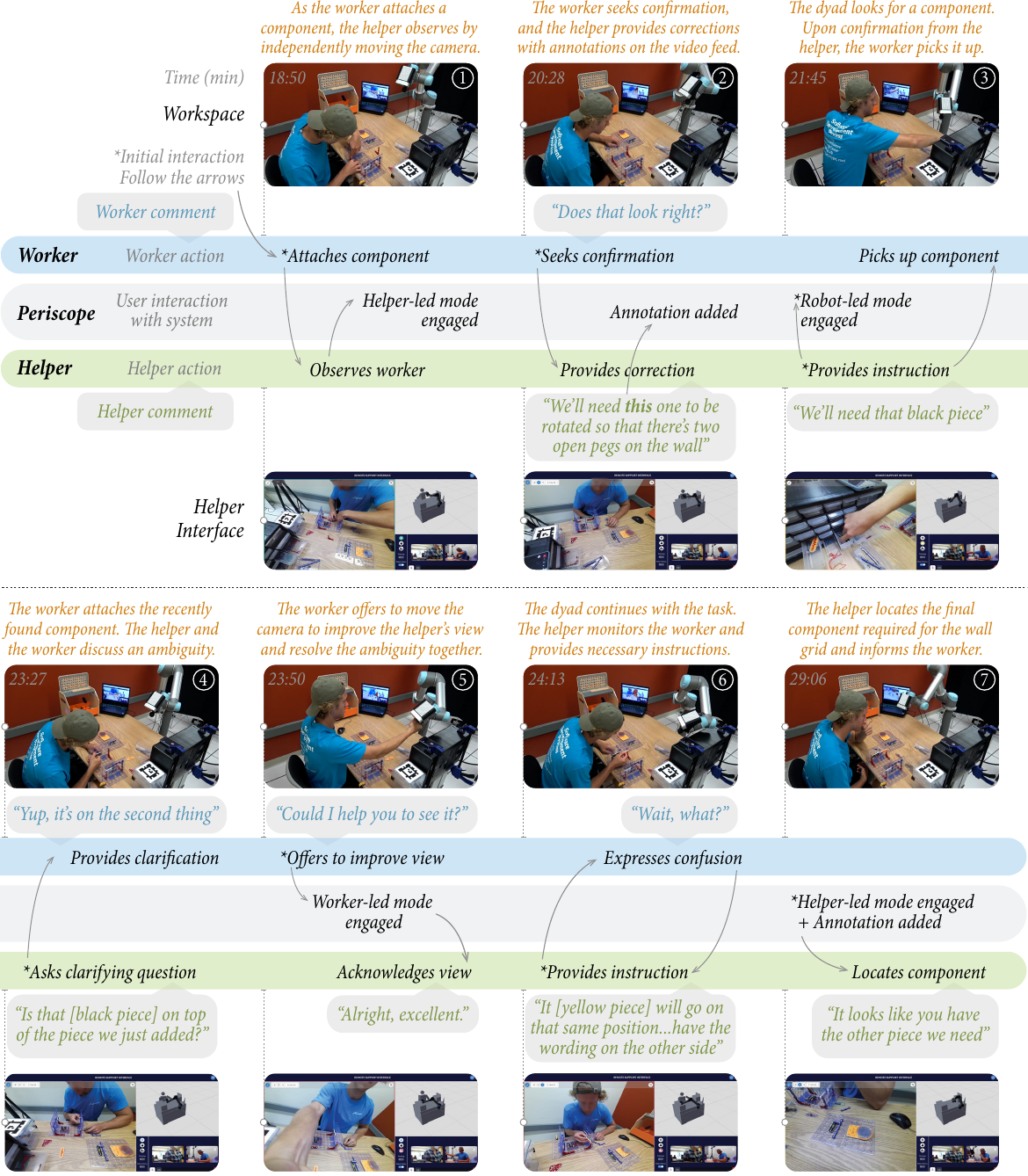}
    \caption{This example interaction from dyad \textit{D4} includes seven micro-interactions spanning 10:16 minutes, with three annotated tracks for worker actions (\textit{``Worker''}), helper actions (\textit{``Helper''}), and their interactions with the \textit{Periscope} system (\textit{``Periscope''}). Each micro-interaction has conversation snippets accompanying the \textit{``Helper''} and \textit{``Worker''} tracks, and images illustrating the worker's space and the helper's interface. Each micro-interaction starts with (*) and subsequent actions/interactions are indicated by arrows. 
    }
    \label{fig:flow}
\end{figure}

Overall, we observed that dyads frequently utilized the \textit{Periscope} system's modes and other features to establish a shared visual context that enhanced their verbal communication (e.g., see Figure~\ref{fig:flow}). 

\begin{table}[h!]
\caption{\centering Summary of use patterns identified from the analysis of video recordings of eight dyads who participated in a user study. Column 2 provides references in \S \ref{sec:results} to details about each pattern.  }
\begin{tabularx}{\textwidth} { 
   >{\hsize=.58\hsize\raggedright\arraybackslash}X 
   >{\hsize=.25\hsize\centering\arraybackslash}X 
   >{\hsize=2.1\hsize\raggedright\arraybackslash}X  }
\hline
\textbf{Feature}               & \# & \textbf{Use Pattern}                                                                                           \\ \hline
\textit{Helper-led mode} & \ref{sec:modeone}-\ref{sec:awareness}  & The helper gains awareness of the workspace.                                                              \\
\textit{}                   & \ref{sec:modeone}-\ref{sec:instructions}  & The helper provides the worker with task instructions.                                                    \\
\textit{}                   & \ref{sec:modeone}-\ref{sec:searching}  & The helper searches for something.                                                                      \\
\textit{}                   & \ref{sec:modeone}-\ref{sec:passing}  & The helper attempts to move the camera before asking the worker do it instead.                               \\ \hline
\textit{Robot-led mode}  & \ref{sec:modetwo}-\ref{sec:gather}  & The dyad gathers components for the build.                                                                \\
\textit{}                   & \ref{sec:modetwo}-\ref{sec:track}  & The helper tracks the worker's movement.                                                                  \\ \hline
\textit{Worker-led mode} & \ref{sec:modethree}-\ref{sec:sharing}  & The worker wants to share some information with the helper.                                               \\
                            & \ref{sec:modethree}-\ref{sec:anticipating}  & The worker anticipates the helper’s need for a different view.                                            \\
                            & \ref{sec:modethree}-\ref{sec:behalf}  & The worker offers to move the camera on behalf of the helper.                                              \\
\textit{}                   & \ref{sec:modethree}-\ref{sec:own}  & The helper attempts and fails to move the camera on their own.                                            \\
\textit{}                   & \ref{sec:modethree}-\ref{sec:already}  & The helper is already aware from an earlier attempt that a particular view is difficult to achieve.       \\
                            & \ref{sec:modethree}-\ref{sec:pertinent}  & The helper requests repositioning the camera that the worker had previously set up.                       \\
\textit{}                   & \ref{sec:modethree}-\ref{sec:position}  & The helper does not know where to position the camera.                                                    \\ \hline
\textit{Point}           & \ref{sec:pointing}-\ref{sec:specific}  & The helper asks the worker for a specific view.                                                           \\
                            & \ref{sec:pointing}-\ref{sec:this}  & The worker refers to something in the workspace.                                                          \\ \hline
\textit{Reset}                       & \ref{sec:reset}-\ref{sec:bookmark}  & The reset pose serves as a bookmarked pose that provides a sufficient view of the workspace with minimal effort. \\
                            & \ref{sec:reset}-\ref{sec:transition}  & The reset pose serves as an intermediate pose when transitioning from one sub-task to the next.           \\
                            & \ref{sec:reset}-\ref{sec:starting}  & The reset pose is a comfortable starting configuration for the \textit{helper-led mode}.                            \\
                            & \ref{sec:reset}-\ref{sec:failure}  & The system does not respond as expected.                                                                  \\ \hline
\textit{Annotate}                    & \ref{sec:annotate}  & The helper refers to something in the workspace.                                                          \\ \hline
\end{tabularx}
\label{table:reasons}

\bigskip

\vspace{0.in}

\caption{\centering Count and duration of use of \textit{Periscope's} features. Dyads are ranked by amount of task completion (\#1 being best). Count represents the number of times a feature was entered and then exited (or the session ended). Duration indicates the average time in seconds spent within a mode per count ($t=$ mean (stdev)).}
\begin{tabularx}{\textwidth} { 
   >{\hsize=.15\hsize\centering\arraybackslash}X 
   >{\hsize=.3\hsize\centering\arraybackslash}X
   >{\hsize=.3\hsize\centering\arraybackslash}X
   >{\hsize=.3\hsize\centering\arraybackslash}X
   >{\hsize=.15\hsize\centering\arraybackslash}X
   >{\hsize=.15\hsize\centering\arraybackslash}X
   >{\hsize=.15\hsize\centering\arraybackslash}X }
\hline
\textbf{Dyad | Rank} & \textit{\textbf{Helper-led Mode}} & \textit{\textbf{Robot-led Mode}} & \textit{\textbf{Worker-led Mode}} & \textit{\textbf{Point}} & \textit{\textbf{Reset}} & \textit{\textbf{Annotate}} \\ \hline
\textit{D1} | \#8   & 9, $t=19 (25)$                                     & 3, $t=70 (20)$                                   & 7, $t=36 (23)$                                    & 0                          & 5                       & 52                         \\ 
\textit{D2} | \#1    & 8, $t=65 (76)$                                    & 2, $t=13 (4)$                                   & 1, $t=18 (NA)$                                    & 0                          & 7                       & 23                         \\
\textit{D3} | \#3    & 7, $t=20 (13)$                                    & 6, $t=126 (122)$                                   & 8, $t=24 (7)$                                    & 6                          & 9                       & 24                         \\
\textit{D4} | \#5   & 15, $t=46 (34)$                                   & 3, $t=18 (17)$                                   & 4, $t=25 (14)$                                    & 2                          & 12                      & 29                         \\
\textit{D5} | \#7   & 2, $t=6 (7)$                                    & 0, $t=NA (NA)$                                   & 15, $t=13 (16)$                                   & 1                          & 0                       & 14                         \\
\textit{D6} | \#4   & 21, $t=44 (39)$                                   & 2, $t=18 (20)$                                   & 5, $t=22 (15)$                                    & 1                          & 12                      & 38                         \\
\textit{D7} | \#2    & 12, $t=42 (48)$                                   & 1, $t=20 (NA)$                                   & 8, $t=25 (12)$                                    & 1                          & 7                       & 36                         \\
\textit{D8} | \#6   & 8, $t=17 (8)$                                    & 3, $t=106 (87)$                                   & 10, $t=27 (19)$                                   & 0                          & 5                       & 41                         \\\hline
Total         & 82, $t=38 (41)$                                   & 20, $t=71 (85)$                                  & 58, $t=23 (16)$                                   & 11                         & 57                      & 257                        \\ \hline
\end{tabularx}
\label{table:frequency}

\end{table}

\begin{figure}[!h]
    \centering
    \includegraphics[width=\textwidth]{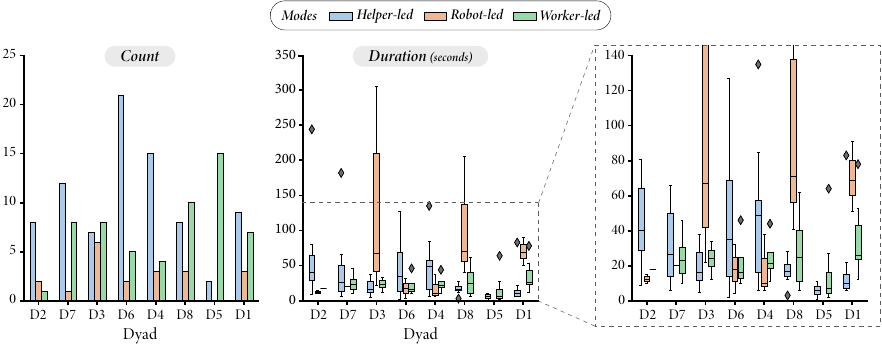}
    \caption{Visualization of the count and duration of use of \textit{Periscope's} modes. Dyads are arranged in descending order based on most to least completion of the task. (Left) Count plot depicting the number of times of use of the three modes in the data. (Center) Box plot depicting the duration of use of the three modes in the data. (Right) Zoomed-in view of the box plot depicting the duration of use of the three modes in the data with y-scale from 0 to 140.}
    \label{fig:results}
\end{figure}

All our results are summarized in Tables \ref{table:reasons} and \ref{table:frequency}, and Figure~\ref{fig:results}. Table~\ref{table:reasons} summarizes a list of use patterns for modes and other features of the system. Table~\ref{table:frequency} provides an overview of the count and duration of use of the modes and other features, and Figure~\ref{fig:results} specifically visualizes the data related to the modes. The use patterns in Table \ref{table:reasons} are nuanced interpretations of the rich multi-modal data that we analyzed. Thus, we include examples in Appendix \ref{sec:appexamples} and cite references to them in this section to offer the context of the rich interactions from which they were interpreted. Table \ref{table:frequency} includes a ranking based on the degree to which each dyad succeeded in completing the main task. We did not expect all dyads to reach completion because we deliberately designed the task to be challenging to prevent dyads from succeeding purely through verbal communication. In the rest of the section, we provide a detailed breakdown of use patterns. In \S \ref{sec:reflection}, we elaborate on the significance of these results to our design goals.

\textbf{Note:} Dyads are enumerated as \textit{D1, D2, D3, D4, D5, D6, D7, D8}. Dyad references and counts of use patterns are included in parentheses.

\subsection{Helper-led Mode Use Patterns} 
\label{sec:modeone}

We observed that this mode was used 82 times (excluding its use when the worker used the pointing gesture which we discuss separately in \S \ref{sec:pointing}). The average duration of each use was 38 seconds ($SD=41$ seconds). We observed that helpers used the mode in the following ways: targeting only \textit{(21/82)}, targeting with zoom adjustment \textit{(27/82)}, and targeting with orbit adjustment \textit{(34/82)}. This mode was the first mode that majority of the helpers used during the session \textit{(6/8)}. The remaining dyads, \textit{D3} and \textit{D5}, used the \textit{worker-led mode} as their first mode. This mode was exited when helpers opened the annotation toolbox \textit{(46/82)}, reset the camera \textit{(16/82)}, switched to the \textit{worker-led mode} \textit{(13/82)}, or switched to the \textit{robot-led mode} \textit{(5/82)}. This mode was not exited in the remaining cases \textit{(2/82)}. Instead, it was either immediately followed by another use of the same mode \textit{(1/2)} or the session ended \textit{(1/2)}. We observed four use patterns, with occasional overlaps \textit{(13/82)}:

\begin{enumerate}
    \item \textit{The helper gains awareness of the workspace (36/82):} The helper inspected various objects in the workspace in order to assess the situation. For example in \textit{D4}, as the worker attached a component onto a grid, the helper wanted to \textit{``double check that it [the component] is facing the correct way''} and moved the camera for a better view of the grid (E\ref{ex:arrow}). This category is distinct because it involves the helper gaining information from the remote workspace. 
    \label{sec:awareness}
        
    \item \textit{The helper provides the worker with task instructions (30/82):} The helper provided guidance to the worker to make progress on the task. For example in \textit{D6}, the helper moved the camera to look at the components that the worker had recently attached and instructed the worker to make modifications, \textit{“This part right here [a base support]...Okay, so you have to flip it”} (E\ref{ex:flip}). This category is distinct because it involves the helper providing information to the worker. 
    \label{sec:instructions}
    
    \item \textit{The helper searches for something (17/82):} When the helper and the worker searched for something together, they typically utilized the \textit{robot-led mode} (\S \ref{sec:modetwo}-\ref{sec:gather}), but if the helper needed to search for something independently, they used the \textit{helper-led mode}. For example in \textit{D1}, the helper explicitly switched from \textit{robot-led mode} to \textit{helper-led mode} while looking for a component, which may have been prompted by the worker not following their instructions correctly (E\ref{ex:switch}). We consider \textit{searching} to be a distinct category in which the helper mostly used targeting only or targeting with zoom adjustment \textit{(16/17)}. 
    \label{sec:searching}
        
    \item \textit{The helper attempts to move the camera before asking the worker do it instead (13/82):} If the initial attempt with the helper-led mode was not sufficient to get the desired view, the helper asked the worker to move the camera using the worker-led mode (see E\ref{ex:closer}). We will revisit this reason in \S \ref{sec:modethree}-\ref{sec:own} when discussing the use of the \textit{worker-led mode}.
    \label{sec:passing}
        
\end{enumerate}

\subsection{Robot-led Mode Use Patterns} 
\label{sec:modetwo}

This mode was used 20 times (excluding its use when the worker used the pointing gesture which we discuss separately in \S \ref{sec:pointing}). The average duration of each use was 71 seconds ($SD=85$ seconds). We observed some view adjustment by the helper \textit{(Adjust: zoom (4/20), Adjust: orbit (2/20))}. 
There were three instances of the robot losing track of the worker's hand requiring the helper to reset the robot \textit{(2/3)} or engage the \textit{helper-led mode (1/3)}. This mode was exited when helpers opened the annotation toolbox \textit{(8/20)}, switched to the \textit{helper-led mode} \textit{(8/20)}, reset the camera \textit{(3/20)}, or switched to the \textit{worker-led mode} \textit{(1/20)}. We observed two use patterns, with one overlap \textit{(1/20)}:

\begin{enumerate}
    \item  \textit{The dyad gathers components for the build (17/20):} The \textit{robot-led mode} was mostly employed to locate the required components in the organizer (see E\ref{ex:follow}). In the majority of these cases, the helper explicitly informed the worker that the tracking mode was on and that their hand was being tracked \textit{(16/17)}. In one instance \textit{(D4)}, the worker held their hand visible to the camera as if to direct the robot, prompting the helper to switch from the \textit{helper-led mode} to the \textit{robot-led mode}. In all cases, we observed that workers explicitly directed the camera by moving their hand to relevant locations \textit{(17/17)}. Additionally, when waiting for the helper to give further instructions, workers often rested their hand on the table to maintain a steady view of the relevant area \textit{(13/17)}. The \textit{robot-led mode} was most frequently used by dyads \textit{D1 (3/17)}, \textit{D3 (6/17)}, and \textit{D8 (3/17)} to find components.
    \label{sec:gather}

    \item  \textit{The helper tracks the worker's movement (4/20):} The helper used the \textit{robot-led mode} to maintain the worker's hand in view as the worker moved their hand to demonstrate or put something together (see E\ref{ex:storage}). In these instances, the worker did not explicitly direct the camera. 
    \label{sec:track}

\end{enumerate}

\subsection{Worker-led Mode Use Patterns} 
\label{sec:modethree}

We observed that this mode was used 58 times in two distinct ways: \textit{worker-initiated (22/58)} or \textit{helper-initiated (36/58)}. This split-use may be due to the design of this mode, which can be engaged either by the worker or the helper. We distinguish between the mode's initiation and engagement; initiation relates to the individual who suggests using the mode, while engagement refers to actually clicking the button. 
The average duration of each use was 23 seconds ($SD=16$ seconds). We observed some view adjustment by the helper \textit{(6/58)}. Helpers often switched to the \textit{helper-led mode} after attempting to adjust the view in this mode \textit{(4/6)}. One instance of view adjustment \textit{(D5)} required the system to be reset since the helper and the worker both attempted to move the camera at the same time, activating the robot's emergency brake. This mode was exited when helpers opened the annotation toolbox \textit{(29/58)}, switched to the \textit{helper-led mode} \textit{(10/58)}, reset the camera \textit{(6/58)}, or switched to the \textit{robot-led mode} \textit{(2/58)}. This mode was not exited in the remaining cases \textit{(11/58)}. Instead, it was either immediately followed by another use of the same mode \textit{(7/11)}, or the session ended \textit{(4/11)}.

\subsubsection{Worker-initiated:} Workers initiated this mode either directly by engaging the mode and moving the camera \textit{(12/22)} or indirectly through conversation \textit{(10/22)}, such as \textit{``Do you need me to move the camera again''} \textit{(D4)}. The former behavior, in which the worker altered the view without notifying the helper, was most prevalent in dyads \textit{D5} and \textit{D8} \textit{(11/12)}. 
Workers initiating this mode mostly engaged the mode themselves \textit{(16/22)} or the mode was already active from prior use \textit{(3/22)}. Otherwise, they asked the helper to engage the mode on their behalf \textit{(3/22)} with a phrase, such as \textit{``Do you want to move to mode 3 [\textit{worker-led mode}] and I can show it?''} \textit{(D1)}.
\label{sec:modethree-one} We observed three reasons for workers initiating the \textit{worker-led mode}:

\begin{enumerate}
    \item \textit{The worker wants to share some information with the helper (7/22):} The worker showed the helper something new in the workspace \textit{(2/7)}, a view pertinent to a query or response that they had regarding the task \textit{(4/7)} (see E\ref{ex:phototransistor}), or their progress on the task \textit{(1/7)}. Workers may \textit{(3/7)} or may not \textit{(4/7)} let the helper know that they are changing the view. 
    \label{sec:sharing}

    \item \textit{The worker anticipates the helper's need for a different view (12/22):} When a helper acknowledged the end of the current step in the process or verbalized the next step in the process (see E\ref{ex:connect}), some workers anticipated the helper's need for a different view and offered to move the camera \textit{(4/12)} or proactively moved the camera without informing the helper \textit{(8/12)}. 
    \label{sec:anticipating}

    \item \textit{The worker offers to move the camera on behalf of the helper (3/22):} When a helper expressed frustration with camera positioning, for instance, by stating, \textit{``Um...let me see if I can move the camera just a little bit''} \textit{(D3)}, some workers offered to move the camera on the helper's behalf. 
    \label{sec:behalf}
        
\end{enumerate}

We observed the least amount of initiation of this mode by the worker in dyads \textit{D2 (none)}, \textit{D6 (once)}, and \textit{D7 (none)}. Additionally, there were six instances of conflict in these dyads---\textit{D2 (1/6)}, \textit{D6 (2/6)}, \textit{D7 (3/6)}---when the worker offered to move the camera or tried to proactively move the camera for any of the reasons mentioned above, but was overruled by the helper who used the \textit{helper-led mode} to move the camera (see E\ref{ex:myself}).

\subsubsection{Helper-initiated:} Helpers initiated this mode with a verbal request to the worker to move the camera. The request was ambiguous and context-specific, yet the worker typically understood it correctly \textit{(31/36)}. For example, one helper requested, \textit{``Could you move the camera so that I'm getting like a more of a bird's eye view''} \textit{(D3)}. While the helper did not indicate which area or item should be visible, the worker showed a view of the base grid based on an earlier conversation about where the base supports would link to on the base grid. If the worker was unable to decide which view to show, there was additional conversation to clarify the request \textit{(5/36)}. When the worker moved the camera, the helper often acknowledged an adequate view \textit{(20/36)} with a phrase such as \textit{``Okay alright, that's enough''} \textit{(D6)}.
\label{sec:modethree-two} We observed four reasons for helpers initiating the \textit{worker-led mode}:

\begin{enumerate}[resume]
    \item \textit{The helper attempts and fails to move the camera on their own (13/36):} If the helper did not get the desired view using the \textit{helper-led mode}, they asked the worker to move the camera. For example in \textit{D7} (E\ref{ex:top}), the helper made an unsuccessful attempt to inspect the roof panel and gave up, saying, \textit{``I am not able to see the top panel...can you?...I need to look up to the panel.''}
    \label{sec:own}

    \item \textit{The helper is already aware from an earlier attempt that a particular view is difficult to achieve (6/36):} The helper preemptively requested the worker to move the camera (see E\ref{ex:before}) because they had previously made an effort to observe the same area but had either been successful after a protracted attempt \textit{(2/6)} or had been unsuccessful and had relied on the worker \textit{(4/6)}.  
    \label{sec:already}
    
    \item \textit{The helper requests repositioning the camera that the worker had previously set up (14/36):} After the \textit{worker-led mode} was used once, there were instances when helpers requested the view to be modified to show something that had become more pertinent (see E\ref{ex:parts}). 
    \label{sec:pertinent}
    
    \item \textit{The helper does not know where to position the camera (3/36):} Since the helper was remote, the worker was more familiar with the layout of the workspace. Thus, the first use of the \textit{worker-led mode} by three helpers \textit{(D3, D5, D6)} was for the worker to move the camera so they could look at something that was located in a place they were unfamiliar with (see E\ref{ex:workbench}). 
    \label{sec:position}

\end{enumerate}

Sometimes, helpers initiating the \textit{worker-led mode} engaged the mode \textit{(10/36)} or the mode was already active from prior use \textit{(4/36)}. Otherwise, workers engaged the mode \textit{(18/36)}. We observed four instances of conflict over mode engagement when both the helper and the worker engaged the mode, thereby canceling out each other's inputs. Additionally, there were two instances of conflict in dyad \textit{D8} when the helper said, \textit{``Can you show me...''}, and used the \textit{helper-led mode} to move the camera. This statement was misunderstood by the worker as a request to engage the \textit{worker-led mode} and move the camera, resulting in overriding the helper's mode selection.

\subsection{Other Use Patterns}
\label{sec:others}

\subsubsection{Point:} We observed 11 instances where pointing was used for two reasons: 
\label{sec:pointing}
\begin{enumerate}
    \item \textit{The helper asks the worker for a specific view (6/11):} Pointing was used explicitly by helpers in dyads \textit{D3 (5/6)} and \textit{D7 (1/6)} to request a view. For example, the helper in \textit{D3} asked the worker, \textit{``Could you point to the wall so that I can see inside it?''} (see E\ref{ex:wall}). 
    \label{sec:specific}
    
    \item \textit{The worker refers to something in the workspace (5/11):} Four workers---\textit{D3 (1/5), D4 (2/5), D5 (1/5), D6 (1/5)}---used pointing to refer to something in the workspace (see E\ref{ex:spatial}). 
    \label{sec:this}

\end{enumerate}

Only one dyad \textit{(D3)} successfully completed the interaction sequence as designed \textit{(3/11)}: worker points, helper approves, and camera provides a close-up of the worker's target. In multiple cases, the helper was unable to approve the worker's target because of a bug in the system \textit{(5/11)}. In these cases, the helper switched to the \textit{robot-led mode} to track the worker's hand \textit{(3/11)}, switched to the \textit{worker-led mode} \textit{(1/11)}, or did not take any action \textit{(1/11)}. In the remaining cases, the helper never attempted to use the \textit{helper-led mode} but directly used the \textit{robot-led mode} when the worker pointed toward something \textit{(3/11)}. Interestingly, some helpers \textit{(D3, D7)} expected the camera to align with the direction of pointing. For example, the helper in \textit{D3} stated, \textit{``Oh, it's looking at your hand and not what I want it to be looking at,''} when the view did not match their expectation of the camera aligning with the direction of pointing (see E\ref{ex:align}).

\subsubsection{Reset:} We observed 57 instances where the helper used the \textit{Reset} feature and identified four potential reasons for its use. However, due to insufficient context in the data to determine the intent behind each occurrence, we do not report the number of instances for each reason. 
\label{sec:reset}

\begin{enumerate}
    \item \textit{The reset pose serves as a bookmarked pose that provides a sufficient view of the workspace with minimal effort:} By simply clicking a button, the helper could easily obtain a reasonable view of most of the workspace (see E\ref{ex:back}).
    \label{sec:bookmark}
    
    \item \textit{The reset pose serves as an intermediate pose when transitioning from one sub-task to the next:} In many instances, the completion of a sub-task was marked by the helper using the \textit{Reset} feature (see E\ref{ex:missing} and E\ref{ex:plan}).
    \label{sec:transition}
    
    \item \textit{The reset pose is a comfortable starting configuration for the \textit{helper-led mode}:} The robot would occasionally get into an odd configuration that the helper found challenging to modify. In such cases, the helper relied on the reset feature to restore the robot to its initial configuration, with which they were familiar and comfortable working (see E\ref{ex:board}).
    \label{sec:starting}

    \item \textit{The system does not respond as expected:} Occasionally, there was a prohibitive lag between user commands and the corresponding robot motion, or the user was unable to move the camera because of issues with the robot's autonomous behaviors, such as being stuck in a collision state or losing track of the worker's hand (see E\ref{ex:lost}). In response, helpers used the \textit{Reset} feature as a way to restore the system to a functional state.
    \label{sec:failure}
    
    
\end{enumerate}

\subsubsection{Annotate:} We observed 257 instances where the helper added annotations to the view. These visual annotations were accompanied with one or more of the following words in the helper's speech: \textit{this, that, these, those, it, other, here, there, where, looks similar/like, same, thing, next, last, one, another, both, right, way, direction, across, on, top, middle, bottom, horizontal, opposite}. While we see evidence of the system facilitating referential communication, a comprehensive conversation analysis on this topic is outside the scope of this work.
\label{sec:annotate} 
\section{Discussion}





In this section, we discuss our system's effectiveness in supporting our design goals introduced in \S \ref{sec:design}, present design implications for future robotic camera systems, address the limitations of our current work, and suggest possible directions for future research. 

\subsection{Reflection on Design Goals}
\label{sec:reflection}

Table \ref{table:reflection} summarizes our reflections on the design goals, highlighting both the \textit{Periscope} system's effectiveness and limitations in supporting them. In the rest of this section, we provide detailed reflections related to each design goal.

\textbf{Note:} Whenever a statement is connected to a result in \S \ref{sec:results} or is illustrated by an example in Appendix \ref{sec:appexamples}, the relevant reference is included in parentheses.

\begin{table}[b!]
\footnotesize
\vspace{-0.1in}
\caption{\centering Summary of reflections on our design goals: ``+'' points indicate our system's effectiveness in supporting the design goals; ``–'' indicate limitations identified in supporting the design goals.}
\begin{tabularx}{\textwidth} {
   >{\hsize=.12\hsize\raggedright\arraybackslash}X 
   >{\hsize=.84\hsize\raggedright\arraybackslash}X
    }
\hline
\textbf{Design Goal} & \textbf{Reflection} \\ \hline                        
\textit{Versatility} & 
\begin{minipage}[t]{0.84\linewidth}
\raggedright
\begin{itemize}[leftmargin=*,noitemsep,topsep=0pt]
    \item[+] Participants used the system's versatility to obtain diverse and context-specific views to support various task activities. The frequent use of the system's features and consistent use patterns across dyads is encouraging, especially since participants were not compelled to use any features.
    \item[+] The shared context afforded by the system, including annotation and deixis, facilitated efficient and unambiguous communication.
    \item[--]  Certain angles and locations were inaccessible due to the choice of robotic hardware, and the system did not adequately support repeated view specifications and precise views for certain task activities.
\end{itemize}
\end{minipage}
\smallskip
\\ \hline
\textit{Intuitivity} &  
\begin{minipage}[t]{0.84\linewidth}
\raggedright
\begin{itemize}[leftmargin=*,noitemsep,topsep=0pt]
    \item[+] The frequent use of camera controls suggests that users found it worthwhile to put in the effort to acquire information through camera control. The emergence of consistent patterns in camera control usage suggests an inherent intuitiveness associated with the controls.
    \item [+] Autonomous robot behaviors were generally invisible to participants, contributing to the intuitiveness of camera controls.
    \item [--] Conversation pauses and dialogue related to camera control interrupted collaboration flow, but users took the time to obtain good views, after which interactions were smooth.
    \item[--] The system did not adequately support repeated view specifications and precise views. Additionally, the lack of autonomous behaviors in \textit{Freedrive} affected camera control by the worker, and in the \textit{Point} interaction, some users expected the camera to align with the direction of pointing.
\end{itemize}
\end{minipage}
\smallskip
\\ \hline
\textit{Dual-user Interactivity} &  
\begin{minipage}[t]{0.84\linewidth}
\raggedright
\begin{itemize}[leftmargin=*,noitemsep,topsep=0pt]
    \item[+] Helpers used the provided interactions to explore the workspace independently and in parallel with the worker's task execution. This allowed for efficient collaboration and timely interventions based on the helper's assessment of the task status without requiring constant dialogue with the worker.
    \item[+] Participants took ownership of the point of view when they had ownership of a task or relevant information, resulting in frequent transfer of view control between the helper and the worker. 
    \item [+] There were two types of shared view control: (1)~collaborative view control within a mode, where users jointly influenced the camera view through dialogue and the provided interactions, and (2)~transfer of view control when switching between modes, where responsibility for camera control was shifted between the helper, the worker, and the robot. 
    \item[--] Workers frequently moved the camera on behalf of the helper when helpers were dissatisfied with their user experience. Workers only occasionally leveraged their familiarity with the workspace to share information with the helper by using the provided interactions.
\end{itemize}
\end{minipage}
\smallskip
\\ \hline
\textit{Congruity} &  
\begin{minipage}[t]{0.84\linewidth}
\raggedright
\begin{itemize}[leftmargin=*,noitemsep,topsep=0pt]
    \item[+] Effective arbitration was facilitated by our leader-follower approach to designing the modes. Additionally, verbal negotiation between the helper
and the worker during collaborative view control helped to achieve congruity. 
    \item[--] There were instances of conflict in the data, particularly regarding the engagement of the worker-led mode, simultaneous attempts by the helper and the worker to move the camera, and a diminished role for the worker and robot during arbitration.
\end{itemize}
\end{minipage}
\smallskip
\\ \hline
\textit{Usability} &  
\begin{minipage}[t]{0.84\linewidth}
\raggedright
\begin{itemize}[leftmargin=*,noitemsep,topsep=0pt]
    \item[+] The system facilitated rich interactions and enabled remote collaboration on physical tasks.
    \item[--] Helpers expressed frustration with some latency and unresponsiveness in robot motion, especially when adjusting the view. Helpers may have different preferences for input sensitivity and direction based on their past experience and task context. Lack of transparency in some state transitions led to confusion when the robot became unresponsive in certain situations. 
    \item [--] Workers split their attention between the task space, the robot, and the shared view on the laptop. The split-attention effect was mitigated to some extent by the embodied cues provided by the robot about the shared view and the helper's focus of attention.
\end{itemize}
\end{minipage}
\smallskip
\\ \hline

\end{tabularx}

\label{table:reflection}
\end{table}

\subsubsection{Versatility:} 
\label{sec:versatility}
The frequent use of the system's features (Table \ref{table:frequency}) and consistent use patterns across dyads (Table \ref{table:reasons}) is encouraging\footnote{The duration of mode use in Table \ref{table:frequency} is more challenging to interpret than the count data, as it is possible that the user found obtaining the desired view difficult and therefore took longer, or that the user was actively accumulating information throughout the entire time. Better interpretations of duration and count data would be possible if we knew the quality of the information users acquired from every view.}, especially since participants were not compelled to use any features to move the camera. The initial configuration (which is also the pre-defined pose for the \textit{Reset} feature) offered a reasonable view of the workspace, and if the worker had brought everything into the static view or the dyads had relied mainly on verbal communication, it may have been possible to progress on the task (albeit inefficiently). However, we found that participants made use of the system's versatility to obtain diverse and context-specific views to support a variety of task activities, such as gaining awareness, providing instructions, searching and gathering components, assembling, sharing information, inspecting objects, and correcting errors.




Similar to prior work \cite{kraut2003visual, gurevich2012teleadvisor}, we saw evidence for the helper and the worker using our interface to establish a shared visual context in order to maintain awareness and ground their conversation. It should be noted that the following discussion about system \textit{versatility} is inherently linked to system \textit{usability}, which enabled effective communication between users. Annotation, in conjunction with the use of deixis (e.g., \textit{this, here, across, now, next}; see \S \ref{sec:annotate}), was the most apparent use of the shared context to achieve efficient and unambiguous communication. Additionally, dyads used the shared context to ground references of task objects (e.g., \textit{``L-shaped stuff''} for the base support in E\ref{ex:adjust} and \textit{``the blue one''} for the snap-connector in E\ref{ex:switch}), especially since they had no prior shared vocabulary for the objects. Finally, infrequent verbal communication related to some aspects of collaboration, such as monitoring comprehension, may suggest an effective use of visual information. Helpers could infer worker comprehension by watching worker actions immediately after receiving instructions, and then correct them if necessary (e.g., E\ref{ex:arrow}). 

We found some limitations in the system's versatility due to the particular robotic hardware that we used. There were some angles and locations that the robot could not be configured to show. Additionally, the system did not adequately support certain task activities such as debugging that required precise views and repeated view specifications (discussed in detail in \S \ref{sec:intuitivity}). These findings provide concrete directions for enhancing the versatility of future systems.


\subsubsection{Intuitivity:} 
\label{sec:intuitivity}

The frequent use of the system's features to move the camera (Table \ref{table:frequency}) may indicate that there were enough instances where users found it worthwhile to put in the effort to acquire information through camera control. Moreover, participants converged on particular patterns in their use of camera controls (Table \ref{table:reasons}), which could suggest that the controls had some degree of intuitivity. It is also promising that autonomous robot behaviors were generally invisible to participants. Occasionally, the robot lost track of the worker's hand and required guidance from the helper (\S \ref{sec:modetwo}) and rare robot collisions required experimenters to restart the system (\S \ref{sec:reset}-\ref{sec:failure}). Otherwise, users did not have to intervene and take responsibility for the aspects of camera control that were handled by the robot. Overall, we believe that the discussion in \S \ref{sec:versatility} of participants using the system to achieve diverse, informative, and task-relevant views is supportive of the intuitiveness of our camera controls.

Conversation pauses and dialogue about camera control in our data raise concerns that participant efforts to move the camera interrupted their flow of collaboration (e.g., E\ref{ex:arrow}, E\ref{ex:adjust}, E\ref{ex:phototransistor}, E\ref{ex:parts}, and E\ref{ex:board}). Nevertheless, helpers and workers took the time to do so in order to get a good view, after which interactions were smooth. This is illustrated in E\ref{ex:circle}, where the verbose description, \textit{``(it should attach on)...the inside of the triangle, like on the inside edge of the triangle that connects to the circle thing...Sorry...the thing...the clear thing with the circle on it''}, was replaced by the concise deictic expression, \textit{``It should attach right...here''}, after the helper took the effort to obtain a good view. While we have taken steps in the right direction with our system design, we explain cases below where our system did not adequately meet this design goal. 

\textit{Obtaining precise views:} Helpers seemed comfortable with camera control when they used targeting only or targeting with zoom (e.g., during searching; see \S \ref{sec:modeone}-\ref{sec:searching}) to set three or four of the camera's DoF. In contrast, helpers had trouble with camera control when trying to obtain views that needed precise 6-DoF camera specification, such as viewing the bottom of the roof grid (E\ref{ex:top}). 

\textit{Repeated view specifications:} Helpers were frustrated with repeatedly specifying views when they had to move away to look for and collect components before returning to finish assembly (e.g., E\ref{ex:before}). Here, the reset pose was useful on occasion since it may be used as a transitional pose when switching between sub-tasks (\S \ref{sec:reset}-\ref{sec:transition}), or as a quick way to get a sufficient view of the workspace without much effort (\S \ref{sec:reset}-\ref{sec:bookmark}).

\textit{Lack of autonomous behaviors in Freedrive:} 
We did not include any autonomous robot behaviors in our implementation of the \textit{Freedrive} interaction for the worker. However, this may have resulted in workers having too many degrees of freedom to manipulate, causing them to sometimes struggle with physically posing the robot's joints. Workers had the most trouble with keeping the camera upright and the robot colliding with itself.


\textit{Non-intuitive pointing behavior:} Some participants expected the camera to align with the direction of pointing and expressed frustration when this was not the case (e.g., E\ref{ex:wall} and E\ref{ex:align}).

\subsubsection{Dual-user Interactivity:} 
\label{sec:interactivity}
We begin with a discussion of how helpers and workers individually used their interactions. Helpers could have simply requested the worker to move the camera each time (as the helper in dyad \textit{D5} did), but most helpers extensively used the interactions provided to them and independently explored the workspace without relying on the worker (\S \ref{sec:modeone}). Additionally, this independence allowed parallel work in which the helper could move the camera as the worker was simultaneously carrying out a task (e.g., E\ref{ex:circle}). The helper could also intervene based on their assessment of the state of the task without always needing to engage the worker in a dialogue about the status (e.g., E\ref{ex:flip}). Workers used the interactions provided to them in two distinct ways. In the intended use, workers leveraged their familiarity and access to the workspace in order to share information with the helper (\S \ref{sec:modethree-one}-\ref{sec:sharing} and \S \ref{sec:modethree-two}-\ref{sec:position}). However, more frequently, workers moved the camera on behalf of the helper when they were dissatisfied with their user experience (discussed later in this subsection). Overall, when participants had ownership of a part of the task or relevant information, they took ownership of the point of view. This finding is consistent with prior work \cite{lanir2013ownership, mentis2020remotely}, but it merits further study to determine if there is a relation (and what its nature is) between the extent to which a user feels task or information ownership and the degree of camera control (e.g., 1-DoF vs 6-DoF) provided by an interaction.


An intriguing and novel outcome of participants having different degrees of camera control in each mode was the frequent transfer of control of the view between the helper and the worker both within and between modes (see mode exit details in \S \ref{sec:modeone}, \S \ref{sec:modetwo}, and \S \ref{sec:modethree}). 
Our analysis revealed that we must consider a user's influence over the view not only through the explicit use of a system feature but also through conversation, such as in the helper-initiated \textit{worker-led mode} (\S \ref{sec:modethree-two}). Influencing the view through conversation was unexpectedly frequent during the use of the \textit{robot-led mode} for gathering components, in which the helper verbally directed the worker to move their hand to modify the view (\S \ref{sec:modetwo}-\ref{sec:gather}). The worker was also mindful of this collaborative view control and exhibited unique behaviors, such as resting their hand on the table to maintain a steady view of the relevant area for the helper. In this scenario, the view is continuously, and sometimes implicitly, negotiated between the helper and the worker. Collaborative view control was also present, but infrequently and intermittently, within the \textit{helper-led mode} and the \textit{worker-led mode}. In the \textit{helper-led mode}, workers could use pointing (although only dyad \textit{D3} successfully used this feature; see \S \ref{sec:pointing}-\ref{sec:this}) and in the \textit{worker-led mode}, helpers could adjust the view themselves or ask the worker to adjust it instead (\S \ref{sec:modethree-two}-\ref{sec:pertinent}). The balance of view control in the \textit{helper-led mode} and the \textit{worker-led mode} may have been skewed disproportionately in favor of either the helper or the worker, making it less apparent than in the \textit{robot-led mode} that view control could be shared.
%


There is an explicit transfer of view control when switching from one mode to another. Users may have changed modes due to the evolving needs of the task that necessitate more or less camera control (e.g., E\ref{ex:switch}). Otherwise, users may exit a mode (in favor of another) when they were unable to acquire the desired view using the interactions provided in that mode. This was more typical with helpers requesting workers to move the camera on their behalf (\S \ref{sec:modethree-two}-\ref{sec:own}, \S \ref{sec:modethree-two}-\ref{sec:already}, \S \ref{sec:modethree-two}-\ref{sec:pertinent}), although there were also cases of the reverse (e.g., E\ref{ex:board} and E\ref{ex:lost}). Although this demonstrates the potential of dual-user interactivity to compensate for shortcomings in the system, future designs of the system should minimize this behavior.

\subsubsection{Congruity:} 
\label{sec:congruity}
The frequent transfer of view control between the helper and the worker within and between modes, which we discuss in connection to dual-user interactivity in \S \ref{sec:interactivity}, is made possible through effective arbitration. We designed arbitration mechanisms within the system to ensure congruity, but interestingly, we observed that verbal negotiation between the helper and the worker during collaborative view control (discussed in \S \ref{sec:interactivity}) also helped to achieve congruity.  
Another facet of arbitration is the role of autonomous robot behaviors in camera control. Autonomous behaviors were generally unobtrusive to participants, as discussed in connection to intuitivity in \S \ref{sec:intuitivity}, and thus contributed to effective arbitration.  

The leader-follower approach (see \S \ref{sec:modes}) that we adopted to streamline arbitration seemed to be an effective strategy, as it may have helped to establish clear roles and ownership. This approach is also linked to the concept of information ownership leading to view ownership, as discussed in \S \ref{sec:interactivity}, where the leader drives the task forward based on information they possess, and the follower follows suit. However, we observed a few instances of conflict in the data, highlighting areas where arbitration could be more effective. There were disagreements between the helper and the worker on when to engage the \textit{worker-led mode} and by whom (\S \ref{sec:modethree}). This is due to both users having the option of engaging the mode. Another source of conflict in this mode was when the helper and the worker both tried to move the camera. Finally, there were issues with the arbitration of the worker's pointing interaction, which required approval by the helper to influence the view and hence diminished the worker's authority (\S \ref{sec:pointing}). While it is promising that there were only a few instances of conflict, we recognize that we may have granted the helper excessive authority during arbitration. The worker had a diminished role in the arbitration process. This made achieving consensus more manageable, but it did not fully leverage the potential contributions that workers could make. Additionally, the robot could also play a more active role and take initiative, rather than just performing passive behaviors in support of helper and worker interactions.

\subsubsection{Usability:} 
\label{sec:usability}
The system facilitates rich interactions between the helper and the worker (illustrated through examples in Appendix \ref{sec:appexamples}) and enables dyads to remotely collaborate on physical tasks. This is promising for the usability for the system. Below, we address usability issues that provide potential for improvement in future systems. 






\textit{Latency and unresponsiveness:} All helpers expressed frustration with the delay between their commands and corresponding robot motion. Furthermore, this latency varied during the session. This was especially problematic when the robot did not immediately respond to commands for adjusting the view (orbit, shift, zoom). Helpers then gave additional commands which caused the robot to overshoot the target location and necessitated correction. 

\textit{Input sensitivity and direction:} We had defined a standard amount and direction of robot movement in response to mouse input, but helpers may have different preferences based on their past experience with other systems and the task context.

\textit{Lack of transparency in certain state transitions:} When the worker moved the camera in the \textit{worker-led mode}, robot assistance through autonomous behaviors was designed to be inactive. However, this meant that the robot might be in a collision state and unable to move for safety reasons when helpers switched to the \textit{helper-led mode} and attempted to move the camera. Since this information was not communicated to users, they assumed that the system was unresponsive and reset the robot's pose to resolve the issue. 

\textit{Split-attention effect for the worker:} The worker's interactions with the system were spatially distributed. Workers engaged \textit{Mode 3} using the interface on the laptop and then moved the robot, which could be in a different part of the workspace than the laptop. While moving the robot, the worker had to simultaneously look at three spatially distributed areas: the task space, the robot (to avoid collisions), and the shared view on the laptop. The split-attention effect seemed less of a factor (although not eliminated) when the helper modified the shared view. The position of the robot-mounted camera changes whenever helpers modified the view, providing embodied cues about the helper's focus of attention to the worker. This could help the worker in achieving joint attention without requiring them to look at the interface on the laptop.








\subsection{Design Implications}
\label{sec:implications}

\subsubsection{Modeless arbitration:} Designing arbitration mechanisms that directly leverage the helper and worker interactions (e.g., target, point, freedrive), without the need for explicit modes, could improve the \textit{intuitivity} and \textit{congruity} of the system. For example, in the current prototype, setting the camera's target as an object versus the worker's hand requires disengaging from one mode and engaging in another. With an integrated interaction system, multiple specifications could be initiated using the same input, such as clicking on the hand in the camera feed to initiate hand tracking, and clicking on an object in the feed to set it as the camera's target. We would like to acknowledge the utility of modes, as they explicitly distribute each user's and the robot's influence on the camera view, establishing clear leader and follower roles. However, exploring a modeless approach presents the possibility of introducing intermediate levels with varying degrees of influence, allowing for more nuanced interactions.  

\subsubsection{Stronger worker-centered design:} Designing the system with explicit support for workers could improve \textit{dual-user interactivity}, particularly because our system, like many other prior works, was designed in a helper-centered manner. For views that are challenging for helpers to specify remotely, incorporating complementary interactions for workers could empower them to more efficiently shape the desired view on behalf of helpers. 
Additionally, in our current design, helpers have significant authority (e.g., to switch between modes). 
Designing the system to encourage variable authority between the helper and the worker could enhance the fluidity of collaboration. For instance, the system could automatically switch to freedrive when the worker makes physical contact with the robot, and switch to remote control when the helper provides mouse input.

\subsubsection{Use pattern-based arbitration:} Designing arbitration based on the use patterns presented in Table \ref{table:reasons} has the potential to improve the \textit{versatility} of the system.
For example, in different contexts, users might require different sensitivity to their directional input when trying to adjust the view. The robot could adjust the amount of movement based on the perceived use pattern (inferred from the state of the environment and usage history). This approach could provide users with the responsiveness needed in one use case versus the precision required in another.    


\subsubsection{Expertise-based arbitration:} Designing arbitration around expertise levels could improve the \textit{intuitivity} and \textit{congruity} of the system. For example, novices may benefit from simplified camera control and a more active robot agent. As users gain expertise, the system could provide them with increased control through new interactions or new ways to parameterize interactions. 


\subsubsection{System feedback:} Providing more frequent and timely feedback to users (e.g., during state transitions) could enhance the \textit{usability} of the system and promote efficient collaboration by reducing the need for dyads to discuss system status. 
The worker may also benefit from more embodied cues that inform about the state of the system.


\subsection{Limitations}
Our work has a number of limitations that primarily stem from the design of our evaluation study. Firstly, although we have envisioned \textit{Periscope} to serve as an expert tool, our evaluation was conducted with novices. We attempted to overcome this discrepancy with extensive training until participants appeared fluent with the system. However, experts who frequently utilize video-based collaboration tools for physical tasks might provide more insight into the challenges they face day-to-day, use our system differently, and provide different feedback. Future studies could focus on expert users, explore other real-world scenarios that they might face, such as expert helpers assisting multiple workers, and apply these tools in more realistic tasks and conditions. Secondly, the setup of our study resulted in a stationary work environment where the robot arm only utilized about a third of its range of movement. Although these constraints afforded greater safety for the participant from collisions with the robot and minimized the potential for discomfort from large motions within close proximity, more research is needed to understand how our system might be used by collaborators in other workspace arrangements. Finally, as we allowed participants to freely interact with the system, determining the specific interaction capabilities of the system that contribute to user success is challenging. Future research that includes comparisons within the system (e.g., evaluating performance when only using one mode for the entire task) and with similar remote collaboration systems could provide a more comprehensive evaluation.


\subsection{Future Work}
We envision a number of potential extensions to our system that point to future research. First, in \textit{Periscope}, modes arbitrated inputs from different sources in a relatively simple fashion. While this approach was sufficient to realize an instance of robotic systems based on shared camera control, future systems that integrate more complex interactions and consider more nuanced circumstances for arbitration will require more sophisticated methods for arbitration. Planning-based approaches, program verification, and optimization-based scheduling are all promising directions that future work can consider. Second, our system assumed a very specific worker-helper setup, and other configurations, such as mutual collaboration, cross-training scenarios, and experts providing remote assistance to multiple workers, will require significant extensions to \textit{Periscope}. These are interesting and challenging scenarios that make up an exciting research space for robotic camera systems. 

We envision additional capabilities for \textit{Periscope} that will require more research. For example, while our work considered how an autonomous agent, other than the helper and the worker, might participate in arbitration, further research is needed to understand how arbitration would apply to a more, or fully, autonomous agent that controls viewpoints by taking initiative. Similarly, \textit{Periscope} can be extended to integrate a semi-autonomous agent with the ability to capture worker actions during periods of helper inattention (due to, e.g., distraction, interruption, assisting other workers) and to provide summaries of work completed. We also found the simulated 3D view of the workspace to be underutilized by collaborators and envision enhancing the capabilities of this view for input (e.g., receiving input directly through the simulation) and output (e.g., offering interactive capabilities through a head-mounted display). Finally, the robot's actions can go beyond supporting shared visual context and include also providing physical assistance to the worker, introducing telemanipulation and autonomous manipulation capabilities to \textit{Periscope}. Manipulation actions by the robot will introduce questions around safety and arbitration, which also serve as interesting avenues for future research. 



\section{Acknowledgements}

An earlier iteration of this manuscript was submitted to the \textit{ACM CHI Conference on Human Factors in Computing Systems (CHI 2023)}. We thank the anonymous reviewers for their careful reading of our manuscript and their insightful suggestions for improvement. We thank Lily Reback (supervised by Pragathi Praveena and Bengisu Cagiltay) for her contribution as Experimenter 2 in conducting the study and coding the data. We thank Haoming Meng (supervised by Pragathi Praveena and Yeping Wang) for his contribution as a software engineer in building the front-end user interfaces. We thank Bengisu Cagiltay for her contribution in organizing and coding the data. We thank Michael Hagenow for his insights and assistance in editing the paper. We thank Seth Petersen (supervised by Yeping Wang) for designing the camera mount. We thank Amy Koike for printing the camera mount. Figure~1 used art by \textit{jcomp} on Freepik. Figure~2 used art by \textit{storyset} on Freepik. 

This work was supported by a NASA University Leadership Initiative (ULI) grant awarded to the University of Wisconsin--Madison and The Boeing Company (Cooperative Agreement \#80NSSC19M0124), and an NSF award 1830242.

\textit{Author Contribution Statement}---The authors confirm contribution to the paper as follows: (1) \textit{conceptualization}: Pragathi Praveena; (2) \textit{design}: Pragathi Praveena, Emmanuel Senft; (3) \textit{implementation}: Pragathi Praveena, Yeping Wang; (4) \textit{data collection and analysis}: Pragathi Praveena; (5) \textit{writing and editing}: all authors; (6) \textit{visualization}: Pragathi Praveena, Yeping Wang, Emmanuel Senft, Bilge Mutlu; (7) \textit{supervision}: Bilge Mutlu, Michael Gleicher.


\bibliographystyle{ACM-Reference-Format}
\bibliography{periscope}


\begin{thebibliography}{95}


\ifx \showCODEN    \undefined \def \showCODEN     #1{\unskip}     \fi
\ifx \showDOI      \undefined \def \showDOI       #1{#1}\fi
\ifx \showISBNx    \undefined \def \showISBNx     #1{\unskip}     \fi
\ifx \showISBNxiii \undefined \def \showISBNxiii  #1{\unskip}     \fi
\ifx \showISSN     \undefined \def \showISSN      #1{\unskip}     \fi
\ifx \showLCCN     \undefined \def \showLCCN      #1{\unskip}     \fi
\ifx \shownote     \undefined \def \shownote      #1{#1}          \fi
\ifx \showarticletitle \undefined \def \showarticletitle #1{#1}   \fi
\ifx \showURL      \undefined \def \showURL       {\relax}        \fi
\providecommand\bibfield[2]{#2}
\providecommand\bibinfo[2]{#2}
\providecommand\natexlab[1]{#1}
\providecommand\showeprint[2][]{arXiv:#2}

\bibitem[Abi-Farraj et~al\mbox{.}(2016)]%
        {abi2016visual}
\bibfield{author}{\bibinfo{person}{Firas Abi-Farraj}, \bibinfo{person}{Nicolò
  Pedemonte}, {and} \bibinfo{person}{Paolo Robuffo~Giordano}.}
  \bibinfo{year}{2016}\natexlab{}.
\newblock \showarticletitle{A Visual-Based Shared Control Architecture for
  Remote Telemanipulation}. In \bibinfo{booktitle}{\emph{2016 IEEE/RSJ
  International Conference on Intelligent Robots and Systems (IROS)}}.
  \bibinfo{pages}{4266--4273}.
\newblock
\urldef\tempurl%
\url{https://doi.org/10.1109/IROS.2016.7759628}
\showDOI{\tempurl}


\bibitem[Adalgeirsson and Breazeal(2010)]%
        {adalgeirsson2010mebot}
\bibfield{author}{\bibinfo{person}{Sigurdur~Orn Adalgeirsson} {and}
  \bibinfo{person}{Cynthia Breazeal}.} \bibinfo{year}{2010}\natexlab{}.
\newblock \showarticletitle{MeBot: A Robotic Platform for Socially Embodied
  Telepresence}. In \bibinfo{booktitle}{\emph{2010 5th ACM/IEEE International
  Conference on Human-Robot Interaction (HRI)}}. \bibinfo{pages}{15--22}.
\newblock
\urldef\tempurl%
\url{https://doi.org/10.1109/HRI.2010.5453272}
\showDOI{\tempurl}


\bibitem[Adcock et~al\mbox{.}(2013)]%
        {adcock2013remotefusion}
\bibfield{author}{\bibinfo{person}{Matt Adcock}, \bibinfo{person}{Stuart
  Anderson}, {and} \bibinfo{person}{Bruce Thomas}.}
  \bibinfo{year}{2013}\natexlab{}.
\newblock \showarticletitle{RemoteFusion: Real Time Depth Camera Fusion for
  Remote Collaboration on Physical Tasks}. In
  \bibinfo{booktitle}{\emph{Proceedings of the 12th ACM SIGGRAPH International
  Conference on Virtual-Reality Continuum and Its Applications in Industry}}
  (Hong Kong, Hong Kong) \emph{(\bibinfo{series}{VRCAI '13})}.
  \bibinfo{publisher}{Association for Computing Machinery},
  \bibinfo{address}{New York, NY, USA}, \bibinfo{pages}{235–242}.
\newblock
\showISBNx{9781450325905}
\urldef\tempurl%
\url{https://doi.org/10.1145/2534329.2534331}
\showDOI{\tempurl}


\bibitem[Akkil et~al\mbox{.}(2016)]%
        {akkil2016gazetorch}
\bibfield{author}{\bibinfo{person}{Deepak Akkil}, \bibinfo{person}{Jobin~Mathew
  James}, \bibinfo{person}{Poika Isokoski}, {and} \bibinfo{person}{Jari
  Kangas}.} \bibinfo{year}{2016}\natexlab{}.
\newblock \showarticletitle{GazeTorch: Enabling Gaze Awareness in Collaborative
  Physical Tasks}. In \bibinfo{booktitle}{\emph{Proceedings of the 2016 CHI
  Conference Extended Abstracts on Human Factors in Computing Systems}} (San
  Jose, California, USA) \emph{(\bibinfo{series}{CHI EA '16})}.
  \bibinfo{publisher}{Association for Computing Machinery},
  \bibinfo{address}{New York, NY, USA}, \bibinfo{pages}{1151–1158}.
\newblock
\showISBNx{9781450340823}
\urldef\tempurl%
\url{https://doi.org/10.1145/2851581.2892459}
\showDOI{\tempurl}


\bibitem[Bai et~al\mbox{.}(2020)]%
        {bai2020user}
\bibfield{author}{\bibinfo{person}{Huidong Bai}, \bibinfo{person}{Prasanth
  Sasikumar}, \bibinfo{person}{Jing Yang}, {and} \bibinfo{person}{Mark
  Billinghurst}.} \bibinfo{year}{2020}\natexlab{}.
\newblock \showarticletitle{A User Study on Mixed Reality Remote Collaboration
  with Eye Gaze and Hand Gesture Sharing}. In
  \bibinfo{booktitle}{\emph{Proceedings of the 2020 CHI Conference on Human
  Factors in Computing Systems}} (Honolulu, HI, USA)
  \emph{(\bibinfo{series}{CHI '20})}. \bibinfo{publisher}{Association for
  Computing Machinery}, \bibinfo{address}{New York, NY, USA},
  \bibinfo{pages}{1–13}.
\newblock
\showISBNx{9781450367080}
\urldef\tempurl%
\url{https://doi.org/10.1145/3313831.3376550}
\showDOI{\tempurl}


\bibitem[Braun and Clarke(2012)]%
        {braun2012thematic}
\bibfield{author}{\bibinfo{person}{Virginia Braun} {and}
  \bibinfo{person}{Victoria Clarke}.} \bibinfo{year}{2012}\natexlab{}.
\newblock \showarticletitle{Thematic Analysis}.
\newblock In \bibinfo{booktitle}{\emph{APA Handbook of Research Methods in
  Psychology, Vol. 2. Research Designs: Quantitative, Qualitative,
  Neuropsychological, and Biological}},
  \bibfield{editor}{\bibinfo{person}{Harris Cooper}, \bibinfo{person}{Paul~M.
  Camic}, \bibinfo{person}{David~L. Long}, \bibinfo{person}{A.~T. Panter},
  \bibinfo{person}{David Rindskopf}, {and} \bibinfo{person}{Kenneth~J. Sher}}
  (Eds.). \bibinfo{publisher}{American Psychological Association},
  \bibinfo{pages}{57--71}.
\newblock
\urldef\tempurl%
\url{https://doi.org/10.1037/13620-004}
\showDOI{\tempurl}


\bibitem[Bruner(1995)]%
        {bruner1995joint}
\bibfield{author}{\bibinfo{person}{Jerome Bruner}.}
  \bibinfo{year}{1995}\natexlab{}.
\newblock \showarticletitle{From Joint Attention to the Meeting of Minds}.
\newblock In \bibinfo{booktitle}{\emph{Joint Attention: Its Origins and Role in
  Development}}, \bibfield{editor}{\bibinfo{person}{Carolyn Moore} {and}
  \bibinfo{person}{Patrick Dunham}} (Eds.). \bibinfo{publisher}{Erlbaum},
  \bibinfo{address}{Hillsdale, N.J.}
\newblock


\bibitem[Cao et~al\mbox{.}(2021)]%
        {openpose}
\bibfield{author}{\bibinfo{person}{Z. Cao}, \bibinfo{person}{G. Hidalgo},
  \bibinfo{person}{T. Simon}, \bibinfo{person}{S. Wei}, {and}
  \bibinfo{person}{Y. Sheikh}.} \bibinfo{year}{2021}\natexlab{}.
\newblock \showarticletitle{OpenPose: Realtime Multi-Person 2D Pose Estimation
  Using Part Affinity Fields}.
\newblock \bibinfo{journal}{\emph{IEEE Transactions on Pattern Analysis \&
  Machine Intelligence}} \bibinfo{volume}{43}, \bibinfo{number}{01}
  (\bibinfo{date}{January} \bibinfo{year}{2021}), \bibinfo{pages}{172--186}.
\newblock
\showISSN{1939-3539}
\urldef\tempurl%
\url{https://doi.org/10.1109/TPAMI.2019.2929257}
\showDOI{\tempurl}


\bibitem[Chaumette et~al\mbox{.}(2016)]%
        {chaumette2016visual}
\bibfield{author}{\bibinfo{person}{Fran{\c{c}}ois Chaumette},
  \bibinfo{person}{Seth Hutchinson}, {and} \bibinfo{person}{Peter Corke}.}
  \bibinfo{year}{2016}\natexlab{}.
\newblock \showarticletitle{Visual Servoing}.
\newblock In \bibinfo{booktitle}{\emph{Springer Handbook of Robotics}},
  \bibfield{editor}{\bibinfo{person}{Bruno Siciliano} {and}
  \bibinfo{person}{Oussama Khatib}} (Eds.). \bibinfo{publisher}{Springer
  International Publishing}, \bibinfo{address}{Cham},
  \bibinfo{pages}{841--866}.
\newblock
\showISBNx{978-3-319-32552-1}
\urldef\tempurl%
\url{https://doi.org/10.1007/978-3-319-32552-1_34}
\showDOI{\tempurl}


\bibitem[Choi and Kwak(2017)]%
        {choi2017identity}
\bibfield{author}{\bibinfo{person}{Jung~Ju Choi} {and}
  \bibinfo{person}{Sonya~S. Kwak}.} \bibinfo{year}{2017}\natexlab{}.
\newblock \showarticletitle{Who Is This?: Identity and Presence in
  Robot-Mediated Communication}.
\newblock \bibinfo{journal}{\emph{Cognitive Systems Research}}
  \bibinfo{volume}{43} (\bibinfo{year}{2017}), \bibinfo{pages}{174--189}.
\newblock
\showISSN{1389-0417}
\urldef\tempurl%
\url{https://doi.org/10.1016/j.cogsys.2016.07.006}
\showDOI{\tempurl}


\bibitem[Christianson et~al\mbox{.}(1996)]%
        {christianson1996declarative}
\bibfield{author}{\bibinfo{person}{David~B. Christianson},
  \bibinfo{person}{Sean~E. Anderson}, \bibinfo{person}{Li-wei He},
  \bibinfo{person}{David~H. Salesin}, \bibinfo{person}{Daniel~S. Weld}, {and}
  \bibinfo{person}{Michael~F. Cohen}.} \bibinfo{year}{1996}\natexlab{}.
\newblock \showarticletitle{Declarative Camera Control for Automatic
  Cinematography}. In \bibinfo{booktitle}{\emph{Proceedings of the Thirteenth
  National Conference on Artificial Intelligence - Volume 1}} (Portland,
  Oregon) \emph{(\bibinfo{series}{AAAI'96})}. \bibinfo{publisher}{AAAI Press},
  \bibinfo{pages}{148–155}.
\newblock
\showISBNx{026251091X}


\bibitem[Christie et~al\mbox{.}(2008)]%
        {christie2008camera}
\bibfield{author}{\bibinfo{person}{Marc Christie}, \bibinfo{person}{Patrick
  Olivier}, {and} \bibinfo{person}{Jean-Marie Normand}.}
  \bibinfo{year}{2008}\natexlab{}.
\newblock \showarticletitle{Camera Control in Computer Graphics}.
\newblock \bibinfo{journal}{\emph{Computer Graphics Forum}}
  \bibinfo{volume}{27}, \bibinfo{number}{8}, \bibinfo{pages}{2197--2218}.
\newblock
\urldef\tempurl%
\url{https://doi.org/10.1111/j.1467-8659.2008.01181.x}
\showDOI{\tempurl}


\bibitem[Clark and Marshall(1981)]%
        {clark1981definite}
\bibfield{author}{\bibinfo{person}{Herbert~H. Clark} {and}
  \bibinfo{person}{Catherine~R. Marshall}.} \bibinfo{year}{1981}\natexlab{}.
\newblock \showarticletitle{Definite Knowledge and Mutual Knowledge}.
\newblock In \bibinfo{booktitle}{\emph{Elements of Discourse Understanding}},
  \bibfield{editor}{\bibinfo{person}{Aravind~K. Joshi},
  \bibinfo{person}{Bonnie~L. Webber}, {and} \bibinfo{person}{Ivan~A. Sag}}
  (Eds.). \bibinfo{publisher}{Cambridge, UK: Cambridge University Press},
  \bibinfo{pages}{10--63}.
\newblock


\bibitem[Crasborn and Sloetjes(2008)]%
        {crasborn2008enhanced}
\bibfield{author}{\bibinfo{person}{Onno Crasborn} {and} \bibinfo{person}{Han
  Sloetjes}.} \bibinfo{year}{2008}\natexlab{}.
\newblock \showarticletitle{Enhanced ELAN Functionality for Sign Language
  Corpora}. In \bibinfo{booktitle}{\emph{6th International Conference on
  Language Resources and Evaluation (LREC 2008)/3rd Workshop on the
  Representation and Processing of Sign Languages: Construction and
  Exploitation of Sign Language Corpora}}. \bibinfo{pages}{39--43}.
\newblock


\bibitem[Daly-Jones et~al\mbox{.}(1998)]%
        {daly1998some}
\bibfield{author}{\bibinfo{person}{Owen Daly-Jones}, \bibinfo{person}{Andrew
  Monk}, {and} \bibinfo{person}{Leon Watts}.} \bibinfo{year}{1998}\natexlab{}.
\newblock \showarticletitle{Some Advantages of Video Conferencing over
  High-Quality Audio Conferencing: Fluency and Awareness of Attentional Focus}.
\newblock \bibinfo{journal}{\emph{International Journal of Human-Computer
  Studies}} \bibinfo{volume}{49}, \bibinfo{number}{1} (\bibinfo{year}{1998}),
  \bibinfo{pages}{21--58}.
\newblock
\showISSN{1071-5819}
\urldef\tempurl%
\url{https://doi.org/10.1006/ijhc.1998.0195}
\showDOI{\tempurl}


\bibitem[Dragan and Srinivasa(2013)]%
        {dragan2013policy}
\bibfield{author}{\bibinfo{person}{Anca~D. Dragan} {and}
  \bibinfo{person}{Siddhartha~S. Srinivasa}.} \bibinfo{year}{2013}\natexlab{}.
\newblock \showarticletitle{A Policy-Blending Formalism for Shared Control}.
\newblock \bibinfo{journal}{\emph{The International Journal of Robotics
  Research}} \bibinfo{volume}{32}, \bibinfo{number}{7} (\bibinfo{year}{2013}),
  \bibinfo{pages}{790--805}.
\newblock
\urldef\tempurl%
\url{https://doi.org/10.1177/0278364913490324}
\showDOI{\tempurl}


\bibitem[Druta et~al\mbox{.}(2021)]%
        {druta2021review}
\bibfield{author}{\bibinfo{person}{Romina Druta}, \bibinfo{person}{Cristian
  Druta}, \bibinfo{person}{Paul Negirla}, {and} \bibinfo{person}{Ioan Silea}.}
  \bibinfo{year}{2021}\natexlab{}.
\newblock \showarticletitle{A Review on Methods and Systems for Remote
  Collaboration}.
\newblock \bibinfo{journal}{\emph{Applied Sciences}} \bibinfo{volume}{11},
  \bibinfo{number}{21} (\bibinfo{year}{2021}), \bibinfo{pages}{10035}.
\newblock
\urldef\tempurl%
\url{https://doi.org/10.3390/app112110035}
\showDOI{\tempurl}


\bibitem[Endsley(1995)]%
        {endsley1995measurement}
\bibfield{author}{\bibinfo{person}{Mica~R. Endsley}.}
  \bibinfo{year}{1995}\natexlab{}.
\newblock \showarticletitle{Toward a Theory of Situation Awareness in Dynamic
  Systems}.
\newblock \bibinfo{journal}{\emph{Human Factors}} \bibinfo{volume}{37},
  \bibinfo{number}{1} (\bibinfo{year}{1995}), \bibinfo{pages}{32--64}.
\newblock
\urldef\tempurl%
\url{https://doi.org/10.1518/001872095779049543}
\showDOI{\tempurl}


\bibitem[Feick et~al\mbox{.}(2018)]%
        {feick2018perspective}
\bibfield{author}{\bibinfo{person}{Martin Feick}, \bibinfo{person}{Terrance
  Mok}, \bibinfo{person}{Anthony Tang}, \bibinfo{person}{Lora Oehlberg}, {and}
  \bibinfo{person}{Ehud Sharlin}.} \bibinfo{year}{2018}\natexlab{}.
\newblock \showarticletitle{Perspective on and Re-Orientation of Physical
  Proxies in Object-Focused Remote Collaboration}. In
  \bibinfo{booktitle}{\emph{Proceedings of the 2018 CHI Conference on Human
  Factors in Computing Systems}} (Montreal QC, Canada)
  \emph{(\bibinfo{series}{CHI '18})}. \bibinfo{publisher}{Association for
  Computing Machinery}, \bibinfo{address}{New York, NY, USA},
  \bibinfo{pages}{1–13}.
\newblock
\showISBNx{9781450356206}
\urldef\tempurl%
\url{https://doi.org/10.1145/3173574.3173855}
\showDOI{\tempurl}


\bibitem[Flor(1998)]%
        {flor1998side}
\bibfield{author}{\bibinfo{person}{Nick~V Flor}.}
  \bibinfo{year}{1998}\natexlab{}.
\newblock \showarticletitle{Side-by-side Collaboration: A Case Study}.
\newblock \bibinfo{journal}{\emph{International Journal of Human-Computer
  Studies}} \bibinfo{volume}{49}, \bibinfo{number}{3} (\bibinfo{year}{1998}),
  \bibinfo{pages}{201--222}.
\newblock
\showISSN{1071-5819}
\urldef\tempurl%
\url{https://doi.org/10.1006/ijhc.1998.0203}
\showDOI{\tempurl}


\bibitem[Fussell et~al\mbox{.}(2000)]%
        {fussell2000coordination}
\bibfield{author}{\bibinfo{person}{Susan~R. Fussell},
  \bibinfo{person}{Robert~E. Kraut}, {and} \bibinfo{person}{Jane Siegel}.}
  \bibinfo{year}{2000}\natexlab{}.
\newblock \showarticletitle{Coordination of Communication: Effects of Shared
  Visual Context on Collaborative Work}. In
  \bibinfo{booktitle}{\emph{Proceedings of the 2000 ACM Conference on Computer
  Supported Cooperative Work}} (Philadelphia, Pennsylvania, USA)
  \emph{(\bibinfo{series}{CSCW '00})}. \bibinfo{publisher}{Association for
  Computing Machinery}, \bibinfo{address}{New York, NY, USA},
  \bibinfo{pages}{21–30}.
\newblock
\showISBNx{1581132220}
\urldef\tempurl%
\url{https://doi.org/10.1145/358916.358947}
\showDOI{\tempurl}


\bibitem[Fussell et~al\mbox{.}(2003a)]%
        {fussell2003effects}
\bibfield{author}{\bibinfo{person}{Susan~R. Fussell},
  \bibinfo{person}{Leslie~D. Setlock}, {and} \bibinfo{person}{Robert~E.
  Kraut}.} \bibinfo{year}{2003}\natexlab{a}.
\newblock \showarticletitle{Effects of Head-Mounted and Scene-Oriented Video
  Systems on Remote Collaboration on Physical Tasks}. In
  \bibinfo{booktitle}{\emph{Proceedings of the SIGCHI Conference on Human
  Factors in Computing Systems}} (Ft. Lauderdale, Florida, USA)
  \emph{(\bibinfo{series}{CHI '03})}. \bibinfo{publisher}{Association for
  Computing Machinery}, \bibinfo{address}{New York, NY, USA},
  \bibinfo{pages}{513–520}.
\newblock
\showISBNx{1581136307}
\urldef\tempurl%
\url{https://doi.org/10.1145/642611.642701}
\showDOI{\tempurl}


\bibitem[Fussell et~al\mbox{.}(2003b)]%
        {fussell2003helpers}
\bibfield{author}{\bibinfo{person}{Susan~R. Fussell},
  \bibinfo{person}{Leslie~D. Setlock}, {and} \bibinfo{person}{Elizabeth~M.
  Parker}.} \bibinfo{year}{2003}\natexlab{b}.
\newblock \showarticletitle{Where Do Helpers Look? Gaze Targets during
  Collaborative Physical Tasks}. In \bibinfo{booktitle}{\emph{CHI '03 Extended
  Abstracts on Human Factors in Computing Systems}} (Ft. Lauderdale, Florida,
  USA) \emph{(\bibinfo{series}{CHI EA '03})}. \bibinfo{publisher}{Association
  for Computing Machinery}, \bibinfo{address}{New York, NY, USA},
  \bibinfo{pages}{768–769}.
\newblock
\showISBNx{1581136374}
\urldef\tempurl%
\url{https://doi.org/10.1145/765891.765980}
\showDOI{\tempurl}


\bibitem[Fussell et~al\mbox{.}(2003c)]%
        {fussell2003assessing}
\bibfield{author}{\bibinfo{person}{Susan~R. Fussell},
  \bibinfo{person}{Leslie~D. Setlock}, \bibinfo{person}{Elizabeth~M. Parker},
  {and} \bibinfo{person}{Jie Yang}.} \bibinfo{year}{2003}\natexlab{c}.
\newblock \showarticletitle{Assessing the Value of a Cursor Pointing Device for
  Remote Collaboration on Physical Tasks}. In \bibinfo{booktitle}{\emph{CHI '03
  Extended Abstracts on Human Factors in Computing Systems}} (Ft. Lauderdale,
  Florida, USA) \emph{(\bibinfo{series}{CHI EA '03})}.
  \bibinfo{publisher}{Association for Computing Machinery},
  \bibinfo{address}{New York, NY, USA}, \bibinfo{pages}{788–789}.
\newblock
\showISBNx{1581136374}
\urldef\tempurl%
\url{https://doi.org/10.1145/765891.765992}
\showDOI{\tempurl}


\bibitem[Fussell et~al\mbox{.}(2004)]%
        {fussell2004gestures}
\bibfield{author}{\bibinfo{person}{Susan~R. Fussell},
  \bibinfo{person}{Leslie~D. Setlock}, \bibinfo{person}{Jie Yang},
  \bibinfo{person}{Jiazhi Ou}, \bibinfo{person}{Elizabeth Mauer}, {and}
  \bibinfo{person}{Adam D.~I. Kramer}.} \bibinfo{year}{2004}\natexlab{}.
\newblock \showarticletitle{Gestures Over Video Streams to Support Remote
  Collaboration on Physical Tasks}.
\newblock \bibinfo{journal}{\emph{Human–Computer Interaction}}
  \bibinfo{volume}{19}, \bibinfo{number}{3} (\bibinfo{year}{2004}),
  \bibinfo{pages}{273--309}.
\newblock
\urldef\tempurl%
\url{https://doi.org/10.1207/s15327051hci1903\_3}
\showDOI{\tempurl}


\bibitem[Gauglitz et~al\mbox{.}(2012)]%
        {gauglitz2012integrating}
\bibfield{author}{\bibinfo{person}{Steffen Gauglitz}, \bibinfo{person}{Cha
  Lee}, \bibinfo{person}{Matthew Turk}, {and} \bibinfo{person}{Tobias
  H\"{o}llerer}.} \bibinfo{year}{2012}\natexlab{}.
\newblock \showarticletitle{Integrating the Physical Environment into Mobile
  Remote Collaboration}. In \bibinfo{booktitle}{\emph{Proceedings of the 14th
  International Conference on Human-Computer Interaction with Mobile Devices
  and Services}} (San Francisco, California, USA)
  \emph{(\bibinfo{series}{MobileHCI '12})}. \bibinfo{publisher}{Association for
  Computing Machinery}, \bibinfo{address}{New York, NY, USA},
  \bibinfo{pages}{241–250}.
\newblock
\showISBNx{9781450311052}
\urldef\tempurl%
\url{https://doi.org/10.1145/2371574.2371610}
\showDOI{\tempurl}


\bibitem[Gauglitz et~al\mbox{.}(2014)]%
        {gauglitz2014touch}
\bibfield{author}{\bibinfo{person}{Steffen Gauglitz}, \bibinfo{person}{Benjamin
  Nuernberger}, \bibinfo{person}{Matthew Turk}, {and} \bibinfo{person}{Tobias
  H\"{o}llerer}.} \bibinfo{year}{2014}\natexlab{}.
\newblock \showarticletitle{In Touch with the Remote World: Remote
  Collaboration with Augmented Reality Drawings and Virtual Navigation}. In
  \bibinfo{booktitle}{\emph{Proceedings of the 20th ACM Symposium on Virtual
  Reality Software and Technology}} (Edinburgh, Scotland)
  \emph{(\bibinfo{series}{VRST '14})}. \bibinfo{publisher}{Association for
  Computing Machinery}, \bibinfo{address}{New York, NY, USA},
  \bibinfo{pages}{197–205}.
\newblock
\showISBNx{9781450332538}
\urldef\tempurl%
\url{https://doi.org/10.1145/2671015.2671016}
\showDOI{\tempurl}


\bibitem[Gaver(1992)]%
        {gaver1992affordances}
\bibfield{author}{\bibinfo{person}{William~W. Gaver}.}
  \bibinfo{year}{1992}\natexlab{}.
\newblock \showarticletitle{The Affordances of Media Spaces for Collaboration}.
  In \bibinfo{booktitle}{\emph{Proceedings of the 1992 ACM Conference on
  Computer-Supported Cooperative Work}} (Toronto, Ontario, Canada)
  \emph{(\bibinfo{series}{CSCW '92})}. \bibinfo{publisher}{Association for
  Computing Machinery}, \bibinfo{address}{New York, NY, USA},
  \bibinfo{pages}{17–24}.
\newblock
\showISBNx{0897915429}
\urldef\tempurl%
\url{https://doi.org/10.1145/143457.371596}
\showDOI{\tempurl}


\bibitem[Gaver et~al\mbox{.}(1993)]%
        {gaver1993one}
\bibfield{author}{\bibinfo{person}{William~W. Gaver}, \bibinfo{person}{Abigail
  Sellen}, \bibinfo{person}{Christian Heath}, {and} \bibinfo{person}{Paul
  Luff}.} \bibinfo{year}{1993}\natexlab{}.
\newblock \showarticletitle{One is Not Enough: Multiple Views in a Media
  Space}. In \bibinfo{booktitle}{\emph{Proceedings of the INTERACT '93 and CHI
  '93 Conference on Human Factors in Computing Systems}} (Amsterdam, The
  Netherlands) \emph{(\bibinfo{series}{CHI '93})}.
  \bibinfo{publisher}{Association for Computing Machinery},
  \bibinfo{address}{New York, NY, USA}, \bibinfo{pages}{335–341}.
\newblock
\showISBNx{0897915755}
\urldef\tempurl%
\url{https://doi.org/10.1145/169059.169268}
\showDOI{\tempurl}


\bibitem[Gleicher and Witkin(1992)]%
        {gleicher1992through}
\bibfield{author}{\bibinfo{person}{Michael Gleicher} {and}
  \bibinfo{person}{Andrew Witkin}.} \bibinfo{year}{1992}\natexlab{}.
\newblock \showarticletitle{Through-the-Lens Camera Control}. In
  \bibinfo{booktitle}{\emph{Proceedings of the 19th Annual Conference on
  Computer Graphics and Interactive Techniques}}
  \emph{(\bibinfo{series}{SIGGRAPH '92})}. \bibinfo{publisher}{Association for
  Computing Machinery}, \bibinfo{address}{New York, NY, USA},
  \bibinfo{pages}{331–340}.
\newblock
\showISBNx{0897914791}
\urldef\tempurl%
\url{https://doi.org/10.1145/133994.134088}
\showDOI{\tempurl}


\bibitem[Gupta et~al\mbox{.}(2016)]%
        {gupta2016do}
\bibfield{author}{\bibinfo{person}{Kunal Gupta}, \bibinfo{person}{Gun~A. Lee},
  {and} \bibinfo{person}{Mark Billinghurst}.} \bibinfo{year}{2016}\natexlab{}.
\newblock \showarticletitle{Do You See What I See? The Effect of Gaze Tracking
  on Task Space Remote Collaboration}.
\newblock \bibinfo{journal}{\emph{IEEE Transactions on Visualization and
  Computer Graphics}} \bibinfo{volume}{22}, \bibinfo{number}{11}
  (\bibinfo{year}{2016}), \bibinfo{pages}{2413--2422}.
\newblock
\urldef\tempurl%
\url{https://doi.org/10.1109/TVCG.2016.2593778}
\showDOI{\tempurl}


\bibitem[Gurevich et~al\mbox{.}(2012)]%
        {gurevich2012teleadvisor}
\bibfield{author}{\bibinfo{person}{Pavel Gurevich}, \bibinfo{person}{Joel
  Lanir}, \bibinfo{person}{Benjamin Cohen}, {and} \bibinfo{person}{Ran Stone}.}
  \bibinfo{year}{2012}\natexlab{}.
\newblock \showarticletitle{TeleAdvisor: A Versatile Augmented Reality Tool for
  Remote Assistance}. In \bibinfo{booktitle}{\emph{Proceedings of the SIGCHI
  Conference on Human Factors in Computing Systems}} (Austin, Texas, USA)
  \emph{(\bibinfo{series}{CHI '12})}. \bibinfo{publisher}{Association for
  Computing Machinery}, \bibinfo{address}{New York, NY, USA},
  \bibinfo{pages}{619–622}.
\newblock
\showISBNx{9781450310154}
\urldef\tempurl%
\url{https://doi.org/10.1145/2207676.2207763}
\showDOI{\tempurl}


\bibitem[Higuch et~al\mbox{.}(2016)]%
        {higuch2016can}
\bibfield{author}{\bibinfo{person}{Keita Higuch}, \bibinfo{person}{Ryo
  Yonetani}, {and} \bibinfo{person}{Yoichi Sato}.}
  \bibinfo{year}{2016}\natexlab{}.
\newblock \showarticletitle{Can Eye Help You? Effects of Visualizing Eye
  Fixations on Remote Collaboration Scenarios for Physical Tasks}. In
  \bibinfo{booktitle}{\emph{Proceedings of the 2016 CHI Conference on Human
  Factors in Computing Systems}} (San Jose, California, USA)
  \emph{(\bibinfo{series}{CHI '16})}. \bibinfo{publisher}{Association for
  Computing Machinery}, \bibinfo{address}{New York, NY, USA},
  \bibinfo{pages}{5180–5190}.
\newblock
\showISBNx{9781450333627}
\urldef\tempurl%
\url{https://doi.org/10.1145/2858036.2858438}
\showDOI{\tempurl}


\bibitem[Hutchinson et~al\mbox{.}(1996)]%
        {hutchinson1996tutorial}
\bibfield{author}{\bibinfo{person}{S. Hutchinson}, \bibinfo{person}{G.D.
  Hager}, {and} \bibinfo{person}{P.I. Corke}.} \bibinfo{year}{1996}\natexlab{}.
\newblock \showarticletitle{A Tutorial on Visual Servo Control}.
\newblock \bibinfo{journal}{\emph{IEEE Transactions on Robotics and
  Automation}} \bibinfo{volume}{12}, \bibinfo{number}{5}
  (\bibinfo{year}{1996}), \bibinfo{pages}{651--670}.
\newblock
\urldef\tempurl%
\url{https://doi.org/10.1109/70.538972}
\showDOI{\tempurl}


\bibitem[Jiang and Arkin(2015)]%
        {jiang2015mixed}
\bibfield{author}{\bibinfo{person}{Shu Jiang} {and} \bibinfo{person}{Ronald~C.
  Arkin}.} \bibinfo{year}{2015}\natexlab{}.
\newblock \showarticletitle{Mixed-Initiative Human-Robot Interaction:
  Definition, Taxonomy, and Survey}. In \bibinfo{booktitle}{\emph{2015 IEEE
  International Conference on Systems, Man, and Cybernetics}}.
  \bibinfo{pages}{954--961}.
\newblock
\urldef\tempurl%
\url{https://doi.org/10.1109/SMC.2015.174}
\showDOI{\tempurl}


\bibitem[Johnson et~al\mbox{.}(2015a)]%
        {johnson2015handheld}
\bibfield{author}{\bibinfo{person}{Steven Johnson}, \bibinfo{person}{Madeleine
  Gibson}, {and} \bibinfo{person}{Bilge Mutlu}.}
  \bibinfo{year}{2015}\natexlab{a}.
\newblock \showarticletitle{Handheld or Handsfree? Remote Collaboration via
  Lightweight Head-Mounted Displays and Handheld Devices}. In
  \bibinfo{booktitle}{\emph{Proceedings of the 18th ACM Conference on Computer
  Supported Cooperative Work \& Social Computing}} (Vancouver, BC, Canada)
  \emph{(\bibinfo{series}{CSCW '15})}. \bibinfo{publisher}{Association for
  Computing Machinery}, \bibinfo{address}{New York, NY, USA},
  \bibinfo{pages}{1825–1836}.
\newblock
\showISBNx{9781450329224}
\urldef\tempurl%
\url{https://doi.org/10.1145/2675133.2675176}
\showDOI{\tempurl}


\bibitem[Johnson et~al\mbox{.}(2015b)]%
        {johnson2015can}
\bibfield{author}{\bibinfo{person}{Steven Johnson}, \bibinfo{person}{Irene
  Rae}, \bibinfo{person}{Bilge Mutlu}, {and} \bibinfo{person}{Leila Takayama}.}
  \bibinfo{year}{2015}\natexlab{b}.
\newblock \showarticletitle{Can You See Me Now? How Field of View Affects
  Collaboration in Robotic Telepresence}. In
  \bibinfo{booktitle}{\emph{Proceedings of the 33rd Annual ACM Conference on
  Human Factors in Computing Systems}} (Seoul, Republic of Korea)
  \emph{(\bibinfo{series}{CHI '15})}. \bibinfo{publisher}{Association for
  Computing Machinery}, \bibinfo{address}{New York, NY, USA},
  \bibinfo{pages}{2397–2406}.
\newblock
\showISBNx{9781450331456}
\urldef\tempurl%
\url{https://doi.org/10.1145/2702123.2702526}
\showDOI{\tempurl}


\bibitem[Kasahara et~al\mbox{.}(2014)]%
        {kasahara2014livesphere}
\bibfield{author}{\bibinfo{person}{Shunichi Kasahara}, \bibinfo{person}{Shohei
  Nagai}, {and} \bibinfo{person}{Jun Rekimoto}.}
  \bibinfo{year}{2014}\natexlab{}.
\newblock \showarticletitle{LiveSphere: Immersive Experience Sharing with 360
  Degrees Head-Mounted Cameras}. In \bibinfo{booktitle}{\emph{Adjunct
  Proceedings of the 27th Annual ACM Symposium on User Interface Software and
  Technology}} (Honolulu, Hawaii, USA) \emph{(\bibinfo{series}{UIST '14
  Adjunct})}. \bibinfo{publisher}{Association for Computing Machinery},
  \bibinfo{address}{New York, NY, USA}, \bibinfo{pages}{61–62}.
\newblock
\showISBNx{9781450330688}
\urldef\tempurl%
\url{https://doi.org/10.1145/2658779.2659114}
\showDOI{\tempurl}


\bibitem[Kehl et~al\mbox{.}(2017)]%
        {kehl2017ssd}
\bibfield{author}{\bibinfo{person}{Wadim Kehl}, \bibinfo{person}{Fabian
  Manhardt}, \bibinfo{person}{Federico Tombari}, \bibinfo{person}{Slobodan
  Ilic}, {and} \bibinfo{person}{Nassir Navab}.}
  \bibinfo{year}{2017}\natexlab{}.
\newblock \showarticletitle{SSD-6D: Making RGB-Based 3D Detection and 6D Pose
  Estimation Great Again}. In \bibinfo{booktitle}{\emph{Proceedings of the IEEE
  International Conference on Computer Vision (ICCV)}}.
\newblock


\bibitem[Kenwright(2015)]%
        {kenwright2015generic}
\bibfield{author}{\bibinfo{person}{Benjamin Kenwright}.}
  \bibinfo{year}{2015}\natexlab{}.
\newblock \showarticletitle{Generic Convex Collision Detection Using Support
  Mapping}.
\newblock \bibinfo{journal}{\emph{Technical Report}} (\bibinfo{year}{2015}).
\newblock


\bibitem[Kim et~al\mbox{.}(2013)]%
        {kim2013comparing}
\bibfield{author}{\bibinfo{person}{Seungwon Kim}, \bibinfo{person}{Gun~A. Lee},
  {and} \bibinfo{person}{Nobuchika Sakata}.} \bibinfo{year}{2013}\natexlab{}.
\newblock \showarticletitle{Comparing Pointing and Drawing for Remote
  Collaboration}. In \bibinfo{booktitle}{\emph{2013 IEEE International
  Symposium on Mixed and Augmented Reality (ISMAR)}}. \bibinfo{pages}{1--6}.
\newblock
\urldef\tempurl%
\url{https://doi.org/10.1109/ISMAR.2013.6671833}
\showDOI{\tempurl}


\bibitem[Kirk et~al\mbox{.}(2007)]%
        {kirk2007turn}
\bibfield{author}{\bibinfo{person}{David Kirk}, \bibinfo{person}{Tom Rodden},
  {and} \bibinfo{person}{Dana\"{e}~Stanton Fraser}.}
  \bibinfo{year}{2007}\natexlab{}.
\newblock \showarticletitle{Turn It This Way: Grounding Collaborative Action
  with Remote Gestures}. In \bibinfo{booktitle}{\emph{Proceedings of the SIGCHI
  Conference on Human Factors in Computing Systems}} (San Jose, California,
  USA) \emph{(\bibinfo{series}{CHI '07})}. \bibinfo{publisher}{Association for
  Computing Machinery}, \bibinfo{address}{New York, NY, USA},
  \bibinfo{pages}{1039–1048}.
\newblock
\showISBNx{9781595935939}
\urldef\tempurl%
\url{https://doi.org/10.1145/1240624.1240782}
\showDOI{\tempurl}


\bibitem[Kiselev et~al\mbox{.}(2014)]%
        {kiselev2014effect}
\bibfield{author}{\bibinfo{person}{Andrey Kiselev}, \bibinfo{person}{Annica
  Kristoffersson}, {and} \bibinfo{person}{Amy Loutfi}.}
  \bibinfo{year}{2014}\natexlab{}.
\newblock \showarticletitle{The Effect of Field of View on Social Interaction
  in Mobile Robotic Telepresence Systems}. In
  \bibinfo{booktitle}{\emph{Proceedings of the 2014 ACM/IEEE International
  Conference on Human-Robot Interaction}} (Bielefeld, Germany)
  \emph{(\bibinfo{series}{HRI '14})}. \bibinfo{publisher}{Association for
  Computing Machinery}, \bibinfo{address}{New York, NY, USA},
  \bibinfo{pages}{214–215}.
\newblock
\showISBNx{9781450326582}
\urldef\tempurl%
\url{https://doi.org/10.1145/2559636.2559799}
\showDOI{\tempurl}


\bibitem[Kratz and Rabelo~Ferriera(2016)]%
        {kratz2016immersed}
\bibfield{author}{\bibinfo{person}{Sven Kratz} {and} \bibinfo{person}{Fred
  Rabelo~Ferriera}.} \bibinfo{year}{2016}\natexlab{}.
\newblock \showarticletitle{Immersed Remotely: Evaluating the Use of Head
  Mounted Devices for Remote Collaboration in Robotic Telepresence}. In
  \bibinfo{booktitle}{\emph{2016 25th IEEE International Symposium on Robot and
  Human Interactive Communication (RO-MAN)}}. \bibinfo{pages}{638--645}.
\newblock
\urldef\tempurl%
\url{https://doi.org/10.1109/ROMAN.2016.7745185}
\showDOI{\tempurl}


\bibitem[Kraut et~al\mbox{.}(2003)]%
        {kraut2003visual}
\bibfield{author}{\bibinfo{person}{Robert~E. Kraut}, \bibinfo{person}{Susan~R.
  Fussell}, {and} \bibinfo{person}{Jane Siegel}.}
  \bibinfo{year}{2003}\natexlab{}.
\newblock \showarticletitle{Visual Information as a Conversational Resource in
  Collaborative Physical Tasks}.
\newblock \bibinfo{journal}{\emph{Human–Computer Interaction}}
  \bibinfo{volume}{18}, \bibinfo{number}{1-2} (\bibinfo{year}{2003}),
  \bibinfo{pages}{13--49}.
\newblock
\urldef\tempurl%
\url{https://doi.org/10.1207/S15327051HCI1812\_2}
\showDOI{\tempurl}


\bibitem[Kurata et~al\mbox{.}(2004)]%
        {kurate2004remote}
\bibfield{author}{\bibinfo{person}{T. Kurata}, \bibinfo{person}{N. Sakata},
  \bibinfo{person}{M. Kourogi}, \bibinfo{person}{H. Kuzuoka}, {and}
  \bibinfo{person}{M. Billinghurst}.} \bibinfo{year}{2004}\natexlab{}.
\newblock \showarticletitle{Remote Collaboration Using a Shoulder-Worn Active
  Camera/Laser}. In \bibinfo{booktitle}{\emph{Eighth International Symposium on
  Wearable Computers}}, Vol.~\bibinfo{volume}{1}. \bibinfo{pages}{62--69}.
\newblock
\urldef\tempurl%
\url{https://doi.org/10.1109/ISWC.2004.37}
\showDOI{\tempurl}


\bibitem[Kuzuoka et~al\mbox{.}(1994)]%
        {kuzuoka1994gesturecam}
\bibfield{author}{\bibinfo{person}{Hideaki Kuzuoka}, \bibinfo{person}{Toshio
  Kosuge}, {and} \bibinfo{person}{Masatomo Tanaka}.}
  \bibinfo{year}{1994}\natexlab{}.
\newblock \showarticletitle{GestureCam: A Video Communication System for
  Sympathetic Remote Collaboration}. In \bibinfo{booktitle}{\emph{Proceedings
  of the 1994 ACM Conference on Computer Supported Cooperative Work}} (Chapel
  Hill, North Carolina, USA) \emph{(\bibinfo{series}{CSCW '94})}.
  \bibinfo{publisher}{Association for Computing Machinery},
  \bibinfo{address}{New York, NY, USA}, \bibinfo{pages}{35–43}.
\newblock
\showISBNx{0897916891}
\urldef\tempurl%
\url{https://doi.org/10.1145/192844.192866}
\showDOI{\tempurl}


\bibitem[Kuzuoka et~al\mbox{.}(2000)]%
        {kuzuoka2000gestureman}
\bibfield{author}{\bibinfo{person}{Hideaki Kuzuoka}, \bibinfo{person}{Shinya
  Oyama}, \bibinfo{person}{Keiichi Yamazaki}, \bibinfo{person}{Kenji Suzuki},
  {and} \bibinfo{person}{Mamoru Mitsuishi}.} \bibinfo{year}{2000}\natexlab{}.
\newblock \showarticletitle{GestureMan: A Mobile Robot That Embodies a Remote
  Instructor's Actions}. In \bibinfo{booktitle}{\emph{Proceedings of the 2000
  ACM Conference on Computer Supported Cooperative Work}} (Philadelphia,
  Pennsylvania, USA) \emph{(\bibinfo{series}{CSCW '00})}.
  \bibinfo{publisher}{Association for Computing Machinery},
  \bibinfo{address}{New York, NY, USA}, \bibinfo{pages}{155–162}.
\newblock
\showISBNx{1581132220}
\urldef\tempurl%
\url{https://doi.org/10.1145/358916.358986}
\showDOI{\tempurl}


\bibitem[Lanir et~al\mbox{.}(2013)]%
        {lanir2013ownership}
\bibfield{author}{\bibinfo{person}{Joel Lanir}, \bibinfo{person}{Ran Stone},
  \bibinfo{person}{Benjamin Cohen}, {and} \bibinfo{person}{Pavel Gurevich}.}
  \bibinfo{year}{2013}\natexlab{}.
\newblock \showarticletitle{Ownership and Control of Point of View in Remote
  Assistance}. In \bibinfo{booktitle}{\emph{Proceedings of the SIGCHI
  Conference on Human Factors in Computing Systems}} (Paris, France)
  \emph{(\bibinfo{series}{CHI '13})}. \bibinfo{publisher}{Association for
  Computing Machinery}, \bibinfo{address}{New York, NY, USA},
  \bibinfo{pages}{2243–2252}.
\newblock
\showISBNx{9781450318990}
\urldef\tempurl%
\url{https://doi.org/10.1145/2470654.2481309}
\showDOI{\tempurl}


\bibitem[Lee et~al\mbox{.}(2017)]%
        {lee2017mixed}
\bibfield{author}{\bibinfo{person}{Gun~A. Lee}, \bibinfo{person}{Theophilus
  Teo}, \bibinfo{person}{Seungwon Kim}, {and} \bibinfo{person}{Mark
  Billinghurst}.} \bibinfo{year}{2017}\natexlab{}.
\newblock \showarticletitle{Mixed Reality Collaboration through Sharing a Live
  Panorama}.
\newblock In \bibinfo{booktitle}{\emph{SIGGRAPH Asia 2017 Mobile Graphics \&
  Interactive Applications}}. \bibinfo{publisher}{Association for Computing
  Machinery}, \bibinfo{address}{New York, NY, USA}.
\newblock
\showISBNx{9781450354103}
\urldef\tempurl%
\url{https://doi.org/10.1145/3132787.3139203}
\showDOI{\tempurl}


\bibitem[Levinson(2006)]%
        {levinson2004deixis}
\bibfield{author}{\bibinfo{person}{Stephen~C. Levinson}.}
  \bibinfo{year}{2006}\natexlab{}.
\newblock \showarticletitle{Deixis}.
\newblock In \bibinfo{booktitle}{\emph{The Handbook of Pragmatics}},
  \bibfield{editor}{\bibinfo{person}{Laurence~R. Horn} {and}
  \bibinfo{person}{Gregory Ward}} (Eds.).
\newblock
\urldef\tempurl%
\url{https://doi.org/10.1002/9780470756959.ch5}
\showDOI{\tempurl}


\bibitem[Li et~al\mbox{.}(2022)]%
        {li2022asteroids}
\bibfield{author}{\bibinfo{person}{Jiannan Li}, \bibinfo{person}{Maur\'{\i}cio
  Sousa}, \bibinfo{person}{Chu Li}, \bibinfo{person}{Jessie Liu},
  \bibinfo{person}{Yan Chen}, \bibinfo{person}{Ravin Balakrishnan}, {and}
  \bibinfo{person}{Tovi Grossman}.} \bibinfo{year}{2022}\natexlab{}.
\newblock \showarticletitle{ASTEROIDS: Exploring Swarms of Mini-Telepresence
  Robots for Physical Skill Demonstration}. In
  \bibinfo{booktitle}{\emph{Proceedings of the 2022 CHI Conference on Human
  Factors in Computing Systems}} (New Orleans, LA, USA)
  \emph{(\bibinfo{series}{CHI '22})}. \bibinfo{publisher}{Association for
  Computing Machinery}, \bibinfo{address}{New York, NY, USA}, Article
  \bibinfo{articleno}{111}, \bibinfo{numpages}{14}~pages.
\newblock
\showISBNx{9781450391573}
\urldef\tempurl%
\url{https://doi.org/10.1145/3491102.3501927}
\showDOI{\tempurl}


\bibitem[Losey et~al\mbox{.}(2018)]%
        {losey2018review}
\bibfield{author}{\bibinfo{person}{David~P. Losey}, \bibinfo{person}{Craig~G.
  McDonald}, \bibinfo{person}{Edoardo Battaglia}, {and}
  \bibinfo{person}{Marcia~K. O'Malley}.} \bibinfo{year}{2018}\natexlab{}.
\newblock \showarticletitle{A Review of Intent Detection, Arbitration, and
  Communication Aspects of Shared Control for Physical Human–Robot
  Interaction}.
\newblock \bibinfo{journal}{\emph{ASME Applied Mechanics Reviews}}
  \bibinfo{volume}{70}, \bibinfo{number}{1} (\bibinfo{date}{January}
  \bibinfo{year}{2018}), \bibinfo{pages}{010804}.
\newblock
\urldef\tempurl%
\url{https://doi.org/10.1115/1.4039145}
\showDOI{\tempurl}


\bibitem[Lugaresi et~al\mbox{.}(2019)]%
        {lugaresi2019mediapipe}
\bibfield{author}{\bibinfo{person}{Camillo Lugaresi}, \bibinfo{person}{Jiuqiang
  Tang}, \bibinfo{person}{Hadon Nash}, \bibinfo{person}{Chris McClanahan},
  \bibinfo{person}{Esha Uboweja}, \bibinfo{person}{Michael Hays},
  \bibinfo{person}{Fan Zhang}, \bibinfo{person}{Chuo-Ling Chang},
  \bibinfo{person}{Ming~Guang Yong}, \bibinfo{person}{Juhyun Lee},
  {et~al\mbox{.}}} \bibinfo{year}{2019}\natexlab{}.
\newblock \showarticletitle{Mediapipe: A Framework for Building Perception
  Pipelines}.
\newblock \bibinfo{journal}{\emph{arXiv preprint arXiv:1906.08172}}
  (\bibinfo{year}{2019}).
\newblock


\bibitem[Macharet and Florencio(2012)]%
        {macharet2012collaborative}
\bibfield{author}{\bibinfo{person}{Douglas~G. Macharet} {and}
  \bibinfo{person}{Dinei~A. Florencio}.} \bibinfo{year}{2012}\natexlab{}.
\newblock \showarticletitle{A Collaborative Control System for Telepresence
  Robots}. In \bibinfo{booktitle}{\emph{2012 IEEE/RSJ International Conference
  on Intelligent Robots and Systems}}. \bibinfo{pages}{5105--5111}.
\newblock
\urldef\tempurl%
\url{https://doi.org/10.1109/IROS.2012.6385705}
\showDOI{\tempurl}


\bibitem[Machino et~al\mbox{.}(2006)]%
        {machino2006remote}
\bibfield{author}{\bibinfo{person}{T. Machino}, \bibinfo{person}{S. Iwaki},
  \bibinfo{person}{H. Kawata}, \bibinfo{person}{Y. Yanagihara},
  \bibinfo{person}{Y. Nanjo}, {and} \bibinfo{person}{K.-i. Shimokura}.}
  \bibinfo{year}{2006}\natexlab{}.
\newblock \showarticletitle{Remote-Collaboration System Using Mobile Robot with
  Camera and Projector}. In \bibinfo{booktitle}{\emph{Proceedings 2006 IEEE
  International Conference on Robotics and Automation, 2006. ICRA 2006.}}
  \bibinfo{pages}{4063--4068}.
\newblock
\urldef\tempurl%
\url{https://doi.org/10.1109/ROBOT.2006.1642326}
\showDOI{\tempurl}


\bibitem[Malyuta et~al\mbox{.}(2020)]%
        {Malyuta2019}
\bibfield{author}{\bibinfo{person}{Dmytro Malyuta}, \bibinfo{person}{Christoph
  Brommer}, \bibinfo{person}{David Hentzen}, \bibinfo{person}{Thomas Stastny},
  \bibinfo{person}{Roland Siegwart}, {and} \bibinfo{person}{Roland Brockers}.}
  \bibinfo{year}{2020}\natexlab{}.
\newblock \showarticletitle{Long-Duration Fully Autonomous Operation of
  Rotorcraft Unmanned Aerial Systems for Remote-Sensing Data Acquisition}.
\newblock \bibinfo{journal}{\emph{Journal of Field Robotics}}
  \bibinfo{volume}{37} (\bibinfo{year}{2020}), \bibinfo{pages}{137--157}.
\newblock
\urldef\tempurl%
\url{https://doi.org/10.1002/rob.21898}
\showDOI{\tempurl}


\bibitem[Marques et~al\mbox{.}(2022)]%
        {marques2022remote}
\bibfield{author}{\bibinfo{person}{Bruno Marques}, \bibinfo{person}{Susana
  Silva}, \bibinfo{person}{João Alves}, {and} \bibinfo{person}{et al.}}
  \bibinfo{year}{2022}\natexlab{}.
\newblock \showarticletitle{Remote Collaboration in Maintenance Contexts Using
  Augmented Reality: Insights from a Participatory Process}.
\newblock \bibinfo{journal}{\emph{International Journal of Interactive Design
  and Manufacturing}}  \bibinfo{volume}{16} (\bibinfo{year}{2022}),
  \bibinfo{pages}{419--438}.
\newblock
\urldef\tempurl%
\url{https://doi.org/10.1007/s12008-021-00798-6}
\showDOI{\tempurl}


\bibitem[Meng et~al\mbox{.}(2023)]%
        {meng2023demo}
\bibfield{author}{\bibinfo{person}{Haoming Meng}, \bibinfo{person}{Yeping
  Wang}, \bibinfo{person}{Pragathi Praveena}, \bibinfo{person}{Michael
  Gleicher}, {and} \bibinfo{person}{Bilge Mutlu}.}
  \bibinfo{year}{2023}\natexlab{}.
\newblock \showarticletitle{Demonstrating Periscope: A Robotic Camera System to
  Support Remote Physical Collaboration}. In
  \bibinfo{booktitle}{\emph{Companion Publication of the 2023 Conference on
  Computer Supported Cooperative Work and Social Computing}}
  \emph{(\bibinfo{series}{CSCW'23 Companion})}. \bibinfo{publisher}{Association
  for Computing Machinery}, \bibinfo{address}{New York, NY, USA},
  \bibinfo{numpages}{4}~pages.
\newblock
\urldef\tempurl%
\url{https://doi.org/10.1145/3584931.3607493}
\showDOI{\tempurl}


\bibitem[Mentis et~al\mbox{.}(2020)]%
        {mentis2020remotely}
\bibfield{author}{\bibinfo{person}{Helena~M. Mentis}, \bibinfo{person}{Yuanyuan
  Feng}, \bibinfo{person}{Azin Semsar}, {and} \bibinfo{person}{Todd~A.
  Ponsky}.} \bibinfo{year}{2020}\natexlab{}.
\newblock \showarticletitle{Remotely Shaping the View in Surgical
  Telementoring}. In \bibinfo{booktitle}{\emph{Proceedings of the 2020 CHI
  Conference on Human Factors in Computing Systems}} (Honolulu, HI, USA)
  \emph{(\bibinfo{series}{CHI '20})}. \bibinfo{publisher}{Association for
  Computing Machinery}, \bibinfo{address}{New York, NY, USA},
  \bibinfo{pages}{1–14}.
\newblock
\showISBNx{9781450367080}
\urldef\tempurl%
\url{https://doi.org/10.1145/3313831.3376622}
\showDOI{\tempurl}


\bibitem[Nicolis et~al\mbox{.}(2018)]%
        {nicolis2018occlusion}
\bibfield{author}{\bibinfo{person}{Davide Nicolis}, \bibinfo{person}{Marco
  Palumbo}, \bibinfo{person}{Andrea~Maria Zanchettin}, {and}
  \bibinfo{person}{Paolo Rocco}.} \bibinfo{year}{2018}\natexlab{}.
\newblock \showarticletitle{Occlusion-Free Visual Servoing for the Shared
  Autonomy Teleoperation of Dual-Arm Robots}.
\newblock \bibinfo{journal}{\emph{IEEE Robotics and Automation Letters}}
  \bibinfo{volume}{3}, \bibinfo{number}{2} (\bibinfo{year}{2018}),
  \bibinfo{pages}{796--803}.
\newblock
\urldef\tempurl%
\url{https://doi.org/10.1109/LRA.2018.2792143}
\showDOI{\tempurl}


\bibitem[Oda et~al\mbox{.}(2015)]%
        {oda2015virtual}
\bibfield{author}{\bibinfo{person}{Ohan Oda}, \bibinfo{person}{Carmine
  Elvezio}, \bibinfo{person}{Mengu Sukan}, \bibinfo{person}{Steven Feiner},
  {and} \bibinfo{person}{Barbara Tversky}.} \bibinfo{year}{2015}\natexlab{}.
\newblock \showarticletitle{Virtual Replicas for Remote Assistance in Virtual
  and Augmented Reality}. In \bibinfo{booktitle}{\emph{Proceedings of the 28th
  Annual ACM Symposium on User Interface Software \& Technology}} (Charlotte,
  NC, USA) \emph{(\bibinfo{series}{UIST '15})}. \bibinfo{publisher}{Association
  for Computing Machinery}, \bibinfo{address}{New York, NY, USA},
  \bibinfo{pages}{405–415}.
\newblock
\showISBNx{9781450337793}
\urldef\tempurl%
\url{https://doi.org/10.1145/2807442.2807497}
\showDOI{\tempurl}


\bibitem[Onishi et~al\mbox{.}(2016)]%
        {onishi2016embodiment}
\bibfield{author}{\bibinfo{person}{Yuya Onishi}, \bibinfo{person}{Kazuaki
  Tanaka}, {and} \bibinfo{person}{Hideyuki Nakanishi}.}
  \bibinfo{year}{2016}\natexlab{}.
\newblock \showarticletitle{Embodiment of Video-Mediated Communication Enhances
  Social Telepresence}. In \bibinfo{booktitle}{\emph{Proceedings of the Fourth
  International Conference on Human Agent Interaction}} (Biopolis, Singapore)
  \emph{(\bibinfo{series}{HAI '16})}. \bibinfo{publisher}{Association for
  Computing Machinery}, \bibinfo{address}{New York, NY, USA},
  \bibinfo{pages}{171–178}.
\newblock
\showISBNx{9781450345088}
\urldef\tempurl%
\url{https://doi.org/10.1145/2974804.2974826}
\showDOI{\tempurl}


\bibitem[Otsuki et~al\mbox{.}(2018)]%
        {otsuki2018effects}
\bibfield{author}{\bibinfo{person}{Mai Otsuki}, \bibinfo{person}{Keita
  Maruyama}, \bibinfo{person}{Hideaki Kuzuoka}, {and} \bibinfo{person}{Yusuke
  SUZUKI}.} \bibinfo{year}{2018}\natexlab{}.
\newblock \showarticletitle{Effects of Enhanced Gaze Presentation on Gaze
  Leading in Remote Collaborative Physical Tasks}. In
  \bibinfo{booktitle}{\emph{Proceedings of the 2018 CHI Conference on Human
  Factors in Computing Systems}} (Montreal QC, Canada)
  \emph{(\bibinfo{series}{CHI '18})}. \bibinfo{publisher}{Association for
  Computing Machinery}, \bibinfo{address}{New York, NY, USA},
  \bibinfo{pages}{1–11}.
\newblock
\showISBNx{9781450356206}
\urldef\tempurl%
\url{https://doi.org/10.1145/3173574.3173942}
\showDOI{\tempurl}


\bibitem[Palmer et~al\mbox{.}(2007)]%
        {palmer2007annotating}
\bibfield{author}{\bibinfo{person}{Doug Palmer}, \bibinfo{person}{Matt Adcock},
  \bibinfo{person}{Jocelyn Smith}, \bibinfo{person}{Matthew Hutchins},
  \bibinfo{person}{Chris Gunn}, \bibinfo{person}{Duncan Stevenson}, {and}
  \bibinfo{person}{Ken Taylor}.} \bibinfo{year}{2007}\natexlab{}.
\newblock \showarticletitle{Annotating with Light for Remote Guidance}. In
  \bibinfo{booktitle}{\emph{Proceedings of the 19th Australasian Conference on
  Computer-Human Interaction: Entertaining User Interfaces}} (Adelaide,
  Australia) \emph{(\bibinfo{series}{OZCHI '07})}.
  \bibinfo{publisher}{Association for Computing Machinery},
  \bibinfo{address}{New York, NY, USA}, \bibinfo{pages}{103–110}.
\newblock
\showISBNx{9781595938725}
\urldef\tempurl%
\url{https://doi.org/10.1145/1324892.1324911}
\showDOI{\tempurl}


\bibitem[Piumsomboon et~al\mbox{.}(2018)]%
        {piumsomboon2018mini}
\bibfield{author}{\bibinfo{person}{Thammathip Piumsomboon},
  \bibinfo{person}{Gun~A. Lee}, \bibinfo{person}{Jonathon~D. Hart},
  \bibinfo{person}{Barrett Ens}, \bibinfo{person}{Robert~W. Lindeman},
  \bibinfo{person}{Bruce~H. Thomas}, {and} \bibinfo{person}{Mark
  Billinghurst}.} \bibinfo{year}{2018}\natexlab{}.
\newblock \showarticletitle{Mini-Me: An Adaptive Avatar for Mixed Reality
  Remote Collaboration}. In \bibinfo{booktitle}{\emph{Proceedings of the 2018
  CHI Conference on Human Factors in Computing Systems}} (Montreal QC, Canada)
  \emph{(\bibinfo{series}{CHI '18})}. \bibinfo{publisher}{Association for
  Computing Machinery}, \bibinfo{address}{New York, NY, USA},
  \bibinfo{pages}{1–13}.
\newblock
\showISBNx{9781450356206}
\urldef\tempurl%
\url{https://doi.org/10.1145/3173574.3173620}
\showDOI{\tempurl}


\bibitem[Piumsomboon et~al\mbox{.}(2019)]%
        {piumsomboon2019on}
\bibfield{author}{\bibinfo{person}{Thammathip Piumsomboon},
  \bibinfo{person}{Gun~A. Lee}, \bibinfo{person}{Andrew Irlitti},
  \bibinfo{person}{Barrett Ens}, \bibinfo{person}{Bruce~H. Thomas}, {and}
  \bibinfo{person}{Mark Billinghurst}.} \bibinfo{year}{2019}\natexlab{}.
\newblock \showarticletitle{On the Shoulder of the Giant: A Multi-Scale Mixed
  Reality Collaboration with 360 Video Sharing and Tangible Interaction}. In
  \bibinfo{booktitle}{\emph{Proceedings of the 2019 CHI Conference on Human
  Factors in Computing Systems}} (Glasgow, Scotland Uk)
  \emph{(\bibinfo{series}{CHI '19})}. \bibinfo{publisher}{Association for
  Computing Machinery}, \bibinfo{address}{New York, NY, USA},
  \bibinfo{pages}{1–17}.
\newblock
\showISBNx{9781450359702}
\urldef\tempurl%
\url{https://doi.org/10.1145/3290605.3300458}
\showDOI{\tempurl}


\bibitem[Rae et~al\mbox{.}(2014)]%
        {rae2014bodies}
\bibfield{author}{\bibinfo{person}{Irene Rae}, \bibinfo{person}{Bilge Mutlu},
  {and} \bibinfo{person}{Leila Takayama}.} \bibinfo{year}{2014}\natexlab{}.
\newblock \showarticletitle{Bodies in Motion: Mobility, Presence, and Task
  Awareness in Telepresence}. In \bibinfo{booktitle}{\emph{Proceedings of the
  SIGCHI Conference on Human Factors in Computing Systems}} (Toronto, Ontario,
  Canada) \emph{(\bibinfo{series}{CHI '14})}. \bibinfo{publisher}{Association
  for Computing Machinery}, \bibinfo{address}{New York, NY, USA},
  \bibinfo{pages}{2153–2162}.
\newblock
\showISBNx{9781450324731}
\urldef\tempurl%
\url{https://doi.org/10.1145/2556288.2557047}
\showDOI{\tempurl}


\bibitem[Rae et~al\mbox{.}(2013a)]%
        {rae2013body}
\bibfield{author}{\bibinfo{person}{Irene Rae}, \bibinfo{person}{Leila
  Takayama}, {and} \bibinfo{person}{Bilge Mutlu}.}
  \bibinfo{year}{2013}\natexlab{a}.
\newblock \showarticletitle{In-Body Experiences: Embodiment, Control, and Trust
  in Robot-Mediated Communication}. In \bibinfo{booktitle}{\emph{Proceedings of
  the SIGCHI Conference on Human Factors in Computing Systems}} (Paris, France)
  \emph{(\bibinfo{series}{CHI '13})}. \bibinfo{publisher}{Association for
  Computing Machinery}, \bibinfo{address}{New York, NY, USA},
  \bibinfo{pages}{1921–1930}.
\newblock
\showISBNx{9781450318990}
\urldef\tempurl%
\url{https://doi.org/10.1145/2470654.2466253}
\showDOI{\tempurl}


\bibitem[Rae et~al\mbox{.}(2013b)]%
        {rae2013influence}
\bibfield{author}{\bibinfo{person}{Irene Rae}, \bibinfo{person}{Leila
  Takayama}, {and} \bibinfo{person}{Bilge Mutlu}.}
  \bibinfo{year}{2013}\natexlab{b}.
\newblock \showarticletitle{The Influence of Height in Robot-Mediated
  Communication}. In \bibinfo{booktitle}{\emph{2013 8th ACM/IEEE International
  Conference on Human-Robot Interaction (HRI)}}. \bibinfo{pages}{1--8}.
\newblock
\urldef\tempurl%
\url{https://doi.org/10.1109/HRI.2013.6483495}
\showDOI{\tempurl}


\bibitem[Rae et~al\mbox{.}(2015)]%
        {rae2015framework}
\bibfield{author}{\bibinfo{person}{Irene Rae}, \bibinfo{person}{Gina Venolia},
  \bibinfo{person}{John~C. Tang}, {and} \bibinfo{person}{David Molnar}.}
  \bibinfo{year}{2015}\natexlab{}.
\newblock \showarticletitle{A Framework for Understanding and Designing
  Telepresence}. In \bibinfo{booktitle}{\emph{Proceedings of the 18th ACM
  Conference on Computer Supported Cooperative Work \& Social Computing}}
  (Vancouver, BC, Canada) \emph{(\bibinfo{series}{CSCW '15})}.
  \bibinfo{publisher}{Association for Computing Machinery},
  \bibinfo{address}{New York, NY, USA}, \bibinfo{pages}{1552–1566}.
\newblock
\showISBNx{9781450329224}
\urldef\tempurl%
\url{https://doi.org/10.1145/2675133.2675141}
\showDOI{\tempurl}


\bibitem[Rakita et~al\mbox{.}(2017)]%
        {rakita2017motion}
\bibfield{author}{\bibinfo{person}{Daniel Rakita}, \bibinfo{person}{Bilge
  Mutlu}, {and} \bibinfo{person}{Michael Gleicher}.}
  \bibinfo{year}{2017}\natexlab{}.
\newblock \showarticletitle{A Motion Retargeting Method for Effective
  Mimicry-Based Teleoperation of Robot Arms}. In
  \bibinfo{booktitle}{\emph{Proceedings of the 2017 ACM/IEEE International
  Conference on Human-Robot Interaction}} (Vienna, Austria)
  \emph{(\bibinfo{series}{HRI '17})}. \bibinfo{publisher}{Association for
  Computing Machinery}, \bibinfo{address}{New York, NY, USA},
  \bibinfo{pages}{361–370}.
\newblock
\showISBNx{9781450343367}
\urldef\tempurl%
\url{https://doi.org/10.1145/2909824.3020254}
\showDOI{\tempurl}


\bibitem[Rakita et~al\mbox{.}(2018)]%
        {rakita2018autonomous}
\bibfield{author}{\bibinfo{person}{Daniel Rakita}, \bibinfo{person}{Bilge
  Mutlu}, {and} \bibinfo{person}{Michael Gleicher}.}
  \bibinfo{year}{2018}\natexlab{}.
\newblock \showarticletitle{An Autonomous Dynamic Camera Method for Effective
  Remote Teleoperation}. In \bibinfo{booktitle}{\emph{Proceedings of the 2018
  ACM/IEEE International Conference on Human-Robot Interaction}} (Chicago, IL,
  USA) \emph{(\bibinfo{series}{HRI '18})}. \bibinfo{publisher}{Association for
  Computing Machinery}, \bibinfo{address}{New York, NY, USA},
  \bibinfo{pages}{325–333}.
\newblock
\showISBNx{9781450349536}
\urldef\tempurl%
\url{https://doi.org/10.1145/3171221.3171279}
\showDOI{\tempurl}


\bibitem[Rakita et~al\mbox{.}(2019)]%
        {rakita2019remote}
\bibfield{author}{\bibinfo{person}{Daniel Rakita}, \bibinfo{person}{Bilge
  Mutlu}, {and} \bibinfo{person}{Michael Gleicher}.}
  \bibinfo{year}{2019}\natexlab{}.
\newblock \showarticletitle{Remote Telemanipulation with Adapting Viewpoints in
  Visually Complex Environments}.
\newblock \bibinfo{journal}{\emph{Robotics: Science and Systems XV}}
  (\bibinfo{year}{2019}).
\newblock


\bibitem[Rakita et~al\mbox{.}(2021)]%
        {rakita2021collisionik}
\bibfield{author}{\bibinfo{person}{Daniel Rakita}, \bibinfo{person}{Haochen
  Shi}, \bibinfo{person}{Bilge Mutlu}, {and} \bibinfo{person}{Michael
  Gleicher}.} \bibinfo{year}{2021}\natexlab{}.
\newblock \showarticletitle{CollisionIK: A Per-Instant Pose Optimization Method
  for Generating Robot Motions with Environment Collision Avoidance}. In
  \bibinfo{booktitle}{\emph{2021 IEEE International Conference on Robotics and
  Automation (ICRA)}}. \bibinfo{pages}{9995--10001}.
\newblock
\urldef\tempurl%
\url{https://doi.org/10.1109/ICRA48506.2021.9561505}
\showDOI{\tempurl}


\bibitem[Ranjan et~al\mbox{.}(2007)]%
        {ranjan2007dynamic}
\bibfield{author}{\bibinfo{person}{Abhishek Ranjan}, \bibinfo{person}{Jeremy~P.
  Birnholtz}, {and} \bibinfo{person}{Ravin Balakrishnan}.}
  \bibinfo{year}{2007}\natexlab{}.
\newblock \showarticletitle{Dynamic Shared Visual Spaces: Experimenting with
  Automatic Camera Control in a Remote Repair Task}. In
  \bibinfo{booktitle}{\emph{Proceedings of the SIGCHI Conference on Human
  Factors in Computing Systems}} (San Jose, California, USA)
  \emph{(\bibinfo{series}{CHI '07})}. \bibinfo{publisher}{Association for
  Computing Machinery}, \bibinfo{address}{New York, NY, USA},
  \bibinfo{pages}{1177–1186}.
\newblock
\showISBNx{9781595935939}
\urldef\tempurl%
\url{https://doi.org/10.1145/1240624.1240802}
\showDOI{\tempurl}


\bibitem[Rasmussen and Huang(2019)]%
        {rasmussen2019scenecam}
\bibfield{author}{\bibinfo{person}{Troels~Ammitsbøl Rasmussen} {and}
  \bibinfo{person}{Weidong Huang}.} \bibinfo{year}{2019}\natexlab{}.
\newblock \showarticletitle{SceneCam: Improving Multi-camera Remote
  Collaboration using Augmented Reality}. In \bibinfo{booktitle}{\emph{2019
  IEEE International Symposium on Mixed and Augmented Reality Adjunct
  (ISMAR-Adjunct)}}. \bibinfo{pages}{28--33}.
\newblock
\urldef\tempurl%
\url{https://doi.org/10.1109/ISMAR-Adjunct.2019.00023}
\showDOI{\tempurl}


\bibitem[Sabet et~al\mbox{.}(2021)]%
        {sabet2021designing}
\bibfield{author}{\bibinfo{person}{Mehrnaz Sabet}, \bibinfo{person}{Mania
  Orand}, {and} \bibinfo{person}{David W.~McDonald}.}
  \bibinfo{year}{2021}\natexlab{}.
\newblock \showarticletitle{Designing Telepresence Drones to Support
  Synchronous, Mid-Air Remote Collaboration: An Exploratory Study}. In
  \bibinfo{booktitle}{\emph{Proceedings of the 2021 CHI Conference on Human
  Factors in Computing Systems}} (Yokohama, Japan) \emph{(\bibinfo{series}{CHI
  '21})}. \bibinfo{publisher}{Association for Computing Machinery},
  \bibinfo{address}{New York, NY, USA}, Article \bibinfo{articleno}{450},
  \bibinfo{numpages}{17}~pages.
\newblock
\showISBNx{9781450380966}
\urldef\tempurl%
\url{https://doi.org/10.1145/3411764.3445041}
\showDOI{\tempurl}


\bibitem[Sakashita et~al\mbox{.}(2022)]%
        {sakashita2022remotecode}
\bibfield{author}{\bibinfo{person}{Mose Sakashita}, \bibinfo{person}{E.~Andy
  Ricci}, \bibinfo{person}{Jatin Arora}, {and} \bibinfo{person}{Fran\c{c}ois
  Guimbreti\`{e}re}.} \bibinfo{year}{2022}\natexlab{}.
\newblock \showarticletitle{RemoteCoDe: Robotic Embodiment for Enhancing
  Peripheral Awareness in Remote Collaboration Tasks}.
\newblock \bibinfo{journal}{\emph{Proc. ACM Hum.-Comput. Interact.}}
  \bibinfo{volume}{6}, \bibinfo{number}{CSCW1}, Article \bibinfo{articleno}{63}
  (\bibinfo{date}{apr} \bibinfo{year}{2022}), \bibinfo{numpages}{22}~pages.
\newblock
\urldef\tempurl%
\url{https://doi.org/10.1145/3512910}
\showDOI{\tempurl}


\bibitem[Sasikumar et~al\mbox{.}(2019)]%
        {sasikumar2019wearable}
\bibfield{author}{\bibinfo{person}{Prasanth Sasikumar}, \bibinfo{person}{Lei
  Gao}, \bibinfo{person}{Huidong Bai}, {and} \bibinfo{person}{Mark
  Billinghurst}.} \bibinfo{year}{2019}\natexlab{}.
\newblock \showarticletitle{Wearable RemoteFusion: A Mixed Reality Remote
  Collaboration System with Local Eye Gaze and Remote Hand Gesture Sharing}. In
  \bibinfo{booktitle}{\emph{2019 IEEE International Symposium on Mixed and
  Augmented Reality Adjunct (ISMAR-Adjunct)}}. \bibinfo{pages}{393--394}.
\newblock
\urldef\tempurl%
\url{https://doi.org/10.1109/ISMAR-Adjunct.2019.000-3}
\showDOI{\tempurl}


\bibitem[Sch\"{a}fer et~al\mbox{.}(2022)]%
        {schafer2021survey}
\bibfield{author}{\bibinfo{person}{Alexander Sch\"{a}fer},
  \bibinfo{person}{Gerd Reis}, {and} \bibinfo{person}{Didier Stricker}.}
  \bibinfo{year}{2022}\natexlab{}.
\newblock \showarticletitle{A Survey on Synchronous Augmented, Virtual,
  AndMixed Reality Remote Collaboration Systems}.
\newblock \bibinfo{journal}{\emph{ACM Comput. Surv.}} \bibinfo{volume}{55},
  \bibinfo{number}{6}, Article \bibinfo{articleno}{116} (\bibinfo{date}{dec}
  \bibinfo{year}{2022}), \bibinfo{numpages}{27}~pages.
\newblock
\showISSN{0360-0300}
\urldef\tempurl%
\url{https://doi.org/10.1145/3533376}
\showDOI{\tempurl}


\bibitem[Senft et~al\mbox{.}(2022)]%
        {senft2022drone}
\bibfield{author}{\bibinfo{person}{Emmanuel Senft}, \bibinfo{person}{Michael
  Hagenow}, \bibinfo{person}{Pragathi Praveena}, \bibinfo{person}{Robert
  Radwin}, \bibinfo{person}{Michael Zinn}, \bibinfo{person}{Michael Gleicher},
  {and} \bibinfo{person}{Bilge Mutlu}.} \bibinfo{year}{2022}\natexlab{}.
\newblock \showarticletitle{A Method For Automated Drone Viewpoints to Support
  Remote Robot Manipulation}. In \bibinfo{booktitle}{\emph{2022 IEEE/RSJ
  International Conference on Intelligent Robots and Systems (IROS)}}.
  \bibinfo{pages}{7704--7711}.
\newblock
\urldef\tempurl%
\url{https://doi.org/10.1109/IROS47612.2022.9982063}
\showDOI{\tempurl}


\bibitem[Sirkin and Ju(2012)]%
        {sirkin2012consistency}
\bibfield{author}{\bibinfo{person}{David Sirkin} {and} \bibinfo{person}{Wendy
  Ju}.} \bibinfo{year}{2012}\natexlab{}.
\newblock \showarticletitle{Consistency in Physical and On-Screen Action
  Improves Perceptions of Telepresence Robots}. In
  \bibinfo{booktitle}{\emph{Proceedings of the Seventh Annual ACM/IEEE
  International Conference on Human-Robot Interaction}} (Boston, Massachusetts,
  USA) \emph{(\bibinfo{series}{HRI '12})}. \bibinfo{publisher}{Association for
  Computing Machinery}, \bibinfo{address}{New York, NY, USA},
  \bibinfo{pages}{57–64}.
\newblock
\showISBNx{9781450310635}
\urldef\tempurl%
\url{https://doi.org/10.1145/2157689.2157699}
\showDOI{\tempurl}


\bibitem[Sodhi et~al\mbox{.}(2013)]%
        {sodhi2013bethere}
\bibfield{author}{\bibinfo{person}{Rajinder~S. Sodhi},
  \bibinfo{person}{Brett~R. Jones}, \bibinfo{person}{David Forsyth},
  \bibinfo{person}{Brian~P. Bailey}, {and} \bibinfo{person}{Giuliano
  Maciocci}.} \bibinfo{year}{2013}\natexlab{}.
\newblock \showarticletitle{BeThere: 3D Mobile Collaboration with Spatial
  Input}. In \bibinfo{booktitle}{\emph{Proceedings of the SIGCHI Conference on
  Human Factors in Computing Systems}} (Paris, France)
  \emph{(\bibinfo{series}{CHI '13})}. \bibinfo{publisher}{Association for
  Computing Machinery}, \bibinfo{address}{New York, NY, USA},
  \bibinfo{pages}{179–188}.
\newblock
\showISBNx{9781450318990}
\urldef\tempurl%
\url{https://doi.org/10.1145/2470654.2470679}
\showDOI{\tempurl}


\bibitem[Speicher et~al\mbox{.}(2018)]%
        {speicher2018360anywhere}
\bibfield{author}{\bibinfo{person}{Maximilian Speicher},
  \bibinfo{person}{Jingchen Cao}, \bibinfo{person}{Ao Yu},
  \bibinfo{person}{Haihua Zhang}, {and} \bibinfo{person}{Michael Nebeling}.}
  \bibinfo{year}{2018}\natexlab{}.
\newblock \showarticletitle{360Anywhere: Mobile Ad-Hoc Collaboration in Any
  Environment Using 360 Video and Augmented Reality}.
\newblock \bibinfo{journal}{\emph{Proc. ACM Hum.-Comput. Interact.}}
  \bibinfo{volume}{2}, \bibinfo{number}{EICS}, Article \bibinfo{articleno}{9}
  (\bibinfo{date}{jun} \bibinfo{year}{2018}), \bibinfo{numpages}{20}~pages.
\newblock
\urldef\tempurl%
\url{https://doi.org/10.1145/3229091}
\showDOI{\tempurl}


\bibitem[Stahl et~al\mbox{.}(2018)]%
        {stahl2018social}
\bibfield{author}{\bibinfo{person}{Christoph Stahl}, \bibinfo{person}{Dimitra
  Anastasiou}, {and} \bibinfo{person}{Thibaud Latour}.}
  \bibinfo{year}{2018}\natexlab{}.
\newblock \showarticletitle{Social Telepresence Robots: The Role of Gesture for
  Collaboration over a Distance}. In \bibinfo{booktitle}{\emph{Proceedings of
  the 11th PErvasive Technologies Related to Assistive Environments
  Conference}} (Corfu, Greece) \emph{(\bibinfo{series}{PETRA '18})}.
  \bibinfo{publisher}{Association for Computing Machinery},
  \bibinfo{address}{New York, NY, USA}, \bibinfo{pages}{409–414}.
\newblock
\showISBNx{9781450363907}
\urldef\tempurl%
\url{https://doi.org/10.1145/3197768.3203180}
\showDOI{\tempurl}


\bibitem[Tang(1991)]%
        {tang1991findings}
\bibfield{author}{\bibinfo{person}{John~C. Tang}.}
  \bibinfo{year}{1991}\natexlab{}.
\newblock \showarticletitle{Findings from Observational Studies of
  Collaborative Work}.
\newblock \bibinfo{journal}{\emph{International Journal of Man-Machine
  Studies}} \bibinfo{volume}{34}, \bibinfo{number}{2} (\bibinfo{year}{1991}),
  \bibinfo{pages}{143--160}.
\newblock
\showISSN{0020-7373}
\urldef\tempurl%
\url{https://doi.org/10.1016/0020-7373(91)90039-A}
\showDOI{\tempurl}
\newblock
\shownote{Special Issue: Computer-supported Cooperative Work and Groupware.
  Part 1}.


\bibitem[Tecchia et~al\mbox{.}(2012)]%
        {tecchia20123d}
\bibfield{author}{\bibinfo{person}{Franco Tecchia}, \bibinfo{person}{Leila
  Alem}, {and} \bibinfo{person}{Weidong Huang}.}
  \bibinfo{year}{2012}\natexlab{}.
\newblock \showarticletitle{3D Helping Hands: A Gesture Based MR System for
  Remote Collaboration}. In \bibinfo{booktitle}{\emph{Proceedings of the 11th
  ACM SIGGRAPH International Conference on Virtual-Reality Continuum and Its
  Applications in Industry}} (Singapore, Singapore)
  \emph{(\bibinfo{series}{VRCAI '12})}. \bibinfo{publisher}{Association for
  Computing Machinery}, \bibinfo{address}{New York, NY, USA},
  \bibinfo{pages}{323–328}.
\newblock
\showISBNx{9781450318259}
\urldef\tempurl%
\url{https://doi.org/10.1145/2407516.2407590}
\showDOI{\tempurl}


\bibitem[Teo et~al\mbox{.}(2019)]%
        {teo2019mixed}
\bibfield{author}{\bibinfo{person}{Theophilus Teo}, \bibinfo{person}{Louise
  Lawrence}, \bibinfo{person}{Gun~A. Lee}, \bibinfo{person}{Mark Billinghurst},
  {and} \bibinfo{person}{Matt Adcock}.} \bibinfo{year}{2019}\natexlab{}.
\newblock \showarticletitle{Mixed Reality Remote Collaboration Combining 360
  Video and 3D Reconstruction}. In \bibinfo{booktitle}{\emph{Proceedings of the
  2019 CHI Conference on Human Factors in Computing Systems}} (Glasgow,
  Scotland Uk) \emph{(\bibinfo{series}{CHI '19})}.
  \bibinfo{publisher}{Association for Computing Machinery},
  \bibinfo{address}{New York, NY, USA}, \bibinfo{pages}{1–14}.
\newblock
\showISBNx{9781450359702}
\urldef\tempurl%
\url{https://doi.org/10.1145/3290605.3300431}
\showDOI{\tempurl}


\bibitem[Thoravi~Kumaravel et~al\mbox{.}(2019)]%
        {thoravi2019loki}
\bibfield{author}{\bibinfo{person}{Balasaravanan Thoravi~Kumaravel},
  \bibinfo{person}{Fraser Anderson}, \bibinfo{person}{George Fitzmaurice},
  \bibinfo{person}{Bjoern Hartmann}, {and} \bibinfo{person}{Tovi Grossman}.}
  \bibinfo{year}{2019}\natexlab{}.
\newblock \showarticletitle{Loki: Facilitating Remote Instruction of Physical
  Tasks Using Bi-Directional Mixed-Reality Telepresence}. In
  \bibinfo{booktitle}{\emph{Proceedings of the 32nd Annual ACM Symposium on
  User Interface Software and Technology}} (New Orleans, LA, USA)
  \emph{(\bibinfo{series}{UIST '19})}. \bibinfo{publisher}{Association for
  Computing Machinery}, \bibinfo{address}{New York, NY, USA},
  \bibinfo{pages}{161–174}.
\newblock
\showISBNx{9781450368162}
\urldef\tempurl%
\url{https://doi.org/10.1145/3332165.3347872}
\showDOI{\tempurl}


\bibitem[Vartiainen et~al\mbox{.}(2015)]%
        {vartiainen2015expert}
\bibfield{author}{\bibinfo{person}{Elina Vartiainen}, \bibinfo{person}{Veronika
  Domova}, {and} \bibinfo{person}{Marcus Englund}.}
  \bibinfo{year}{2015}\natexlab{}.
\newblock \showarticletitle{Expert on Wheels: An Approach to Remote
  Collaboration}. In \bibinfo{booktitle}{\emph{Proceedings of the 3rd
  International Conference on Human-Agent Interaction}} (Daegu, Kyungpook,
  Republic of Korea) \emph{(\bibinfo{series}{HAI '15})}.
  \bibinfo{publisher}{Association for Computing Machinery},
  \bibinfo{address}{New York, NY, USA}, \bibinfo{pages}{49–54}.
\newblock
\showISBNx{9781450335270}
\urldef\tempurl%
\url{https://doi.org/10.1145/2814940.2814943}
\showDOI{\tempurl}


\bibitem[Villanueva et~al\mbox{.}(2021)]%
        {villanueva2021robotar}
\bibfield{author}{\bibinfo{person}{Ana~M Villanueva}, \bibinfo{person}{Ziyi
  Liu}, \bibinfo{person}{Zhengzhe Zhu}, \bibinfo{person}{Xin Du},
  \bibinfo{person}{Joey Huang}, \bibinfo{person}{Kylie~A Peppler}, {and}
  \bibinfo{person}{Karthik Ramani}.} \bibinfo{year}{2021}\natexlab{}.
\newblock \showarticletitle{RobotAR: An Augmented Reality Compatible
  Teleconsulting Robotics Toolkit for Augmented Makerspace Experiences}. In
  \bibinfo{booktitle}{\emph{Proceedings of the 2021 CHI Conference on Human
  Factors in Computing Systems}} (Yokohama, Japan) \emph{(\bibinfo{series}{CHI
  '21})}. \bibinfo{publisher}{Association for Computing Machinery},
  \bibinfo{address}{New York, NY, USA}, Article \bibinfo{articleno}{477},
  \bibinfo{numpages}{13}~pages.
\newblock
\showISBNx{9781450380966}
\urldef\tempurl%
\url{https://doi.org/10.1145/3411764.3445726}
\showDOI{\tempurl}


\bibitem[Wang et~al\mbox{.}(2019)]%
        {wang2019head}
\bibfield{author}{\bibinfo{person}{Peng Wang}, \bibinfo{person}{Shusheng
  Zhang}, \bibinfo{person}{Xiaoliang Bai}, \bibinfo{person}{Mark Billinghurst},
  \bibinfo{person}{Weiping He}, \bibinfo{person}{Shuxia Wang},
  \bibinfo{person}{Xiaokun Zhang}, \bibinfo{person}{Jiaxiang Du}, {and}
  \bibinfo{person}{Yongxing Chen}.} \bibinfo{year}{2019}\natexlab{}.
\newblock \showarticletitle{Head Pointer or Eye Gaze: Which Helps More in MR
  Remote Collaboration?}. In \bibinfo{booktitle}{\emph{2019 IEEE Conference on
  Virtual Reality and 3D User Interfaces (VR)}}. \bibinfo{pages}{1219--1220}.
\newblock
\urldef\tempurl%
\url{https://doi.org/10.1109/VR.2019.8798024}
\showDOI{\tempurl}


\bibitem[Xiao et~al\mbox{.}(2021)]%
        {xiao2020usage}
\bibfield{author}{\bibinfo{person}{Chun Xiao}, \bibinfo{person}{Weidong Huang},
  {and} \bibinfo{person}{Mark Billinghurst}.} \bibinfo{year}{2021}\natexlab{}.
\newblock \showarticletitle{Usage and Effect of Eye Tracking in Remote
  Guidance}. In \bibinfo{booktitle}{\emph{Proceedings of the 32nd Australian
  Conference on Human-Computer Interaction}} (Sydney, NSW, Australia)
  \emph{(\bibinfo{series}{OzCHI '20})}. \bibinfo{publisher}{Association for
  Computing Machinery}, \bibinfo{address}{New York, NY, USA},
  \bibinfo{pages}{622–628}.
\newblock
\showISBNx{9781450389754}
\urldef\tempurl%
\url{https://doi.org/10.1145/3441000.3441051}
\showDOI{\tempurl}


\bibitem[Zhang et~al\mbox{.}(2019)]%
        {zhang2019lightbee}
\bibfield{author}{\bibinfo{person}{Xujing Zhang}, \bibinfo{person}{Sean
  Braley}, \bibinfo{person}{Calvin Rubens}, \bibinfo{person}{Timothy Merritt},
  {and} \bibinfo{person}{Roel Vertegaal}.} \bibinfo{year}{2019}\natexlab{}.
\newblock \showarticletitle{LightBee: A Self-Levitating Light Field Display for
  Hologrammatic Telepresence}. In \bibinfo{booktitle}{\emph{Proceedings of the
  2019 CHI Conference on Human Factors in Computing Systems}} (Glasgow,
  Scotland Uk) \emph{(\bibinfo{series}{CHI '19})}.
  \bibinfo{publisher}{Association for Computing Machinery},
  \bibinfo{address}{New York, NY, USA}, \bibinfo{pages}{1–10}.
\newblock
\showISBNx{9781450359702}
\urldef\tempurl%
\url{https://doi.org/10.1145/3290605.3300242}
\showDOI{\tempurl}


\end{thebibliography}

\appendix

\section{Technical Details}
\label{sec:apptechnical}

\subsection{System Overview}

The system is built on the Robot Operating System (ROS)\footnote{\url{https://www.ros.org/}}, which enables communication between system components and real-time control of the robot arm. In our prototype, we mount an Azure Kinect camera\footnote{\url{https://azure.microsoft.com/en-us/services/kinect-dk/}} on a Universal Robot UR5 collaborative robot arm.\footnote{\url{https://www.universal-robots.com/products/ur5-robot/}} 
The camera provides both color images, which the users can view, and depth data for use in computer vision algorithms.
The color and depth data have a resolution of 2048x1536 and 512x512, respectively.

We used the React framework\footnote{\url{https://reactjs.org/}} for the front-end interface and it connects to the back-end ROS server using \texttt{roslibjs}\footnote{\url{http://wiki.ros.org/roslibjs}}. Visual feedback on the camera feed for input commands and annotations is implemented using React Conva\footnote{\url{https://konvajs.org/docs/react/index.html}}. The 3D view, built on \texttt{ros3djs}\footnote{\url{http://wiki.ros.org/ros3djs}}, shows a simulated visualization of the robot and its surrounding objects in \texttt{threejs}\footnote{\url{https://threejs.org/}} and updates their states in real-time from the back-end ROS server. We use Dolby's API\footnote{\url{https://dolby.io/}} for video conferencing services. The control panel consists of buttons that interact with the ROS back-end. 

\subsection{Motion Generation}
We cast the real-time motion generation problem in a constrained multiple-objective optimization structure. Most user interactions and autonomous behaviors are formulated as objectives.

\begin{equation}
\label{eq:optimization}
\begin{aligned}
    \mathbf{q} = & \argminA_\mathbf{q}
    \sum_{i=1}^N w_i * f(\chi_i(\mathbf{q}))
     \\
    s.t. \,\,\, & l_i \leq q_i \leq u_i, \,\,\, \forall i
\end{aligned}
\end{equation}

Here, $\mathbf{q} \in \mathbb{R}^n$ is the configuration of an $n$-joint robot. $l_i$ and $r_i$ are the upper and lower bounds of the $i$-th robot joint. $N$ is the total number of objectives and $w_i$ is the weight of the $i$-th objective $\chi_i(\mathbf{q})$. 
$f$ is the Groove function introduced by Rakita et al. \cite{rakita2017motion} that normalizes objective values for accommodating multiple-objectives.

The forward kinematics function $\mathbf{\Psi}$ calculates the camera pose given a joint configuration. Forward kinematics functions $\mathbf{\Psi}_p(\mathbf{q}), \mathbf{\Psi}_R(\mathbf{q}), \mathbf{\Psi}_q(\mathbf{q})$ represent the position, rotation matrix, and quaternion of the camera for joint configuration $\mathbf{q}$, respectively.

The optimized joint configuration $\mathbf{q}$ is sent to the robot arm using its native programming language, URScript. URScript additionally has commands that directly support \textit{Reset} and \textit{Freedrive}.

\subsection{Helper Interactions}

\subsubsection{Target}

To point the camera towards a target, we adapt the ``look-at'' objective from prior work \cite{rakita2018autonomous}.
\begin{equation}
    \chi_{\text{set\_target}}(\mathbf{q}) = dist(\mathbf{t}, \mathbf{v})
\end{equation}

\noindent Here, function $dist()$ returns the orthogonal distance between a target position $\mathbf{t} \in \mathbb{R}^3$ and a unit vector $\mathbf{v} \in \mathbb{R}^3$ that indicates the view direction. 

\subsubsection{Adjust} 

To move the camera according to directional inputs, the objective is:

\begin{equation}
    \chi_{\text{adjust}}(\mathbf{q}) = || \mathbf{\Psi}_p(\mathbf{q}_{t-1}) + \mathbf{\Delta} - \mathbf{\Psi}_p(\mathbf{q}_t) ||_2
\end{equation}

\noindent Here, $\mathbf{q}_t$ and $\mathbf{q}_{t-1}$ are the robot joint configuration at time $t$ and $t-1$.  $\mathbf{\Delta} \in \mathbb{R}^3$ is an offset signal. 

\subsubsection{Reset}
To move the camera to a pre-defined starting configuration, the pre-defined joint configuration is sent to the robot arm via URScript.

\subsubsection{Annotate}  The front-end canvas accepts input signals and overlays a pin/rectangle/arrow graphic depending on user selection of shape and subsequent movement.

\subsection{Worker Interactions}

\subsubsection{Point} \label{sec:point}

Pointing detection is built upon the open-source MediaPipe solution \cite{lugaresi2019mediapipe}\footnote{\url{https://google.github.io/mediapipe/solutions/hands.html}}, in which a hand pose is represented by 21 2D landmarks. To detect a pointing gesture, an algorithm checks if the distance from the base of the worker's thumb to the worker's index fingertip is larger than the base of the thumb to all the other fingertips. With pointing being detected, the pointing slider in the control panel is enabled for the helper. If the helper chooses to turn it on, the target of the camera $\mathbf{t}$ is set to the position of the index fingertip in the robot frame.

\subsubsection{Direct.} \label{sec:direct}
As described in \S \ref{sec:point}, we detect landmarks of the worker's hand using MediaPipe. The landmarks are converted from the camera's frame of reference to the robot's frame of reference. The average position of 5 landmarks on the worker's right hand (wrist, base of all fingers) are used as the target $\mathbf{t}$. 

\subsubsection{Freedrive.} 
The robot is switched to freedrive directly via URScript. In freedrive, the robot can be manually moved by the worker into a desired pose. The robot arm senses the forces applied to it and moves in the direction of the force as if it is being pushed or pulled by the user.

\subsection{Autonomous Behaviors}

\subsubsection{Keep distance}
To maintain a specified distance between the camera and a target point, we use an objective from prior work \cite{rakita2018autonomous}:  

\begin{equation}
    \chi_{\text{dist}}(\mathbf{q}) = ||\mathbf{t} - \mathbf{\Psi}_p(\mathbf{q}) ||_2 - d
\end{equation}

\noindent Here, $\mathbf{t} \in \mathbb{R}^3$ is the target position and $d$ is the specified distance. 

\subsubsection{Keep upright}
We adapt an objective that keeps the camera upright from prior work \cite{rakita2018autonomous}.

\begin{equation}
    \chi_{\text{lookat}}(\mathbf{q}) = \left(\mathbf{\Psi}_R(\mathbf{q}) [0,1,0]^\intercal\right) \cdot [0,0,1]^\intercal
\end{equation}

\noindent To keep the camera upright, the camera's ``left'' axis ($y$ axis in our system) should be orthogonal to the vertical axis $[0,0,1]$ in the world frame.

\subsubsection{Track hand} Same as \S \ref{sec:direct}.

\subsubsection{Avoid jerky motion} 

To avoid large and jittery camera motions, both joint motion and camera motion smoothness objective are included in the optimization formulation. Prior work \cite{rakita2018autonomous} assigns equal weights to all robot joints in the
joint motion smoothness objectives. However, the joint that is closer to the robot's base leads to larger camera motion, so we apply higher penalty to these joints. Consequently, the robot has more tendency to make fine movements. 
In our notation, a joint that is closer to the robot's base has a lower index. In our system, the objectives that minimizes joint velocity, acceleration, and jerk are:
\begin{equation}
    \mathbf{\chi}_v(\mathbf{q}) = \sqrt{\sum_i^n (n-i+1) \dot{q_i}^2} ,\,\,\,\,\,\,
    \mathbf{\chi}_a(\mathbf{q}) = \sqrt{\sum_i^n (n-i+1) \ddot{q_i}^2}
    ,\,\,\,\,\,\,
    \mathbf{\chi}_j(\mathbf{q}) = \sqrt{\sum_i^n (n-i+1) \dddot{q_i}^2}
\end{equation}

We use the same objective as prior work \cite{rakita2018autonomous} to minimize the velocity of the camera.

\begin{equation}
    \mathbf{\chi}_\text{ee\_vel}(\mathbf{q}) = || \mathbf{\Psi}_p(\mathbf{q}_t) - \mathbf{\Psi}_p(\mathbf{q}_{t-1}) ||_2
\end{equation}

Although joint limits are set as inequality constraints in our formulation (Equation \ref{eq:optimization}), we also add an objective to keep solutions away from joint limits. 
\begin{equation}
    \chi_{\text{joint\_limits}}(\mathbf{q}) = \sum_{i=1}^n 0.05 \left(\frac{(q_i - l_i)/(u_i-l_i)-0.5}{0.45}\right)^{50}
\end{equation}

\noindent Here, $q_i$,  $l_i$ and $u_i$ are the angle, lower, and upper limit of the $i$-th joint, respectively.

\subsubsection{Avoid collisions} 

We use collision avoidance methods from prior work \cite{rakita2021collisionik} to prevent collisions between the robot arm and the objects in the environment including the worker. These methods allow collision avoidance with both static objects as well as dynamic objects such as the worker. We use the same methods to prevent collisions between the links of the robot arm (self-collisions).
In prior work \cite{rakita2021collisionik}, each robot link $\mathbf{I}_i$ and environment object $\mathbf{e} \in \mathcal{A}$ is wrapped in convex hull shapes. The distance between two convex hull shapes $\textit{dist}()$ is computed using  a Support Mapping method \cite{kenwright2015generic}.
\begin{equation}
    \chi_{\text{self\_collision}}(\mathbf{q}) = \sum_{i=1}^{m-2} \sum_{j=i+2}^m \frac{(5 \epsilon)^2} {{dist}\left(\mathbf{l}_i(\mathbf{q}), \mathbf{l}_j(\mathbf{q})\right)^2} 
\end{equation}

\begin{equation}
    \chi_{\text{env\_collision}}(\mathbf{q}) = \sum_{\mathbf{e} \in \mathcal{A}} \sum_{i=1}^{m} \frac{(5 \epsilon)^2}{{dist}\left(\mathbf{l}_i(\mathbf{q}), \mathbf{e}\right)^2} 
\end{equation}

Here, $m$ is the total number of robot links and  $\epsilon$ is a scalar value that signifies the cutoff distance between collision and non-collision. For both self- and environment collision, we set $\epsilon$ as 0.02. 

To detect the worker's body positions for collision avoidance, we use the open-source OpenPose\cite{openpose} system to extract human body poses from RGB images. Human body poses are represented as 25 keypoints in the RGB image, which we then map to 3D using depth data. Since the depth data can be noisy, we use a median filter to get smooth and stable body keypoints. With these stable 3D keypoints, we create convex hull spaces (e.g., spheres, cuboids) around body parts for robot collision avoidance. The body parts are also visualized in the 3D view panel in the front-end interface.

To detect dynamic object positions, we use AR tags \cite{Malyuta2019} to identify the poses of dynamic objects in the environment. In future iterations, other vision-based pose estimation technologies (e.g., SSD-6D \cite{kehl2017ssd}) could replace the AR tags.

\section{Examples}
\label{sec:appexamples}

\begin{example}[D4]\label{ex:arrow}
    When the worker attached a component on the grid, the helper said, \textit{``Okay...so let me just double check that it [the component] is facing the correct way,''} and moved the camera to get a better view of the grid. While moving the camera, the helper continued the conversation, \textit{``Is the arrow...[the worker indicates the direction of the arrow with their hand]...okay...if the arrow is pointing to the right, then it's in the correct spot.''} Finally, the helper completed the camera movement to get a view of the component and confirmed, \textit{``Yeah, that looks correct to me.''}
\end{example}

\begin{example}[D6]\label{ex:flip}
    While attaching the wall grid to the base grid, the worker asked the helper, \textit{``Am I doing it correct so far?''} The helper replied, \textit{``Yeah, you are doing it correct, yeah...''} while moving the camera to get a better view of the grid. However, after getting a better look at the grid and the recently added components, the helper said, \textit{``Wait, just hold on a minute now...,''} and instructed the worker to make modifications, \textit{``This part right here [a base support]...Okay, so you have to flip it.''} 
\end{example}

\begin{example}[D3]\label{ex:circle}
    The helper struggled with wiring instructions, \textit{``Oh, um...it should attach on the...here [adds annotation]...as well as on the inside of the triangle, like on the inside edge of the triangle that connects to the circle thing...Sorry...the thing...the clear thing with the circle on it,''} and stated, \textit{``I wish I could like look, but I don't think there's a way to get inside the house...maybe if I do this...''} The helper moved the camera and remarked, \textit{``Okay, I see it...sort of...,''} and instructed the worker with an annotation, \textit{``It should attach right...here [adds annotation].''}
\end{example}

\begin{example}[D2]\label{ex:adjust}
    The helper took time to set up the view and prefaced the process by stating, \textit{``Sorry...I need to adjust the camera first. This is not a very comfortable viewing angle for me.''} After moving the camera for a few seconds, the helper continued, \textit{``Okay, this is nice [acknowledging the view]...So first you want to fix...this L-shaped stuff [base supports]...like here [adds annotation] and here [adds annotation].''}
\end{example}

\begin{example}[D1]\label{ex:switch}
    In the \textit{robot-led mode} where the robot was tracking the worker's hand, the worker asked, \textit{``Which drawer do you want me to open up here?''} The worker moved toward a drawer, and the helper responded, \textit{``We don't need the blue one...we need to find us more...''} The worker then independently moved their hand to a different location in the workspace, changing the view unexpectedly for the helper. Frustrated, the helper switched to the \textit{helper-led mode} and said, \textit{``Okay, hold on... I will open mode 1 [\textit{helper-led mode}]...I almost find it,''} and instructed the worker to pick up the required component, \textit{``We have to pick up the red one.''}
\end{example}

\begin{example}[D8]\label{ex:closer}
    When trying to view the vertical wall grid, the helper first moved the camera with the \textit{helper-led mode}. The helper was mostly successful in getting a good view of the grid but finally asked the worker, \textit{``Can you move the camera a little bit closer in Mode 3 [\textit{worker-led mode}] so that I can see it better?''} 
\end{example}

\begin{example}[D3]\label{ex:follow}
    During the search for the last component on the wall grid, the helper stated to the worker, \textit{``I'm gonna have it [the robot] follow your hand, and you're going to start opening drawers again...[worker moves hand]...one above it...[worker moves hand]...nope not that...[worker moves hand]...one above it...Maybe take that out and show me?...[worker brings the component closer to the camera]...Yeah, that's what we're looking for.''}
\end{example}

\begin{example}[D4]\label{ex:storage}
    The worker remarked, \textit{``Oh, here it is''}, before picking up and presenting to the helper a storage box that the dyad was looking for. The helper engaged the \textit{robot-led mode} in response to the worker's remark to track the worker's movement.
\end{example}

\begin{example}[D6]\label{ex:phototransistor}
    While the helper was adjusting the camera view, the worker examined the circuit and noticed a potential error. The worker said, \textit{``I also think that the phototransistor might be upside down... (Helper: Is it?) ...I can show you''}, before engaging the \textit{worker-led mode} to show the helper the phototransistor component. 
\end{example}   

\begin{example}[D5]\label{ex:connect}
    As the helper explained the next step with the instruction, \textit{``You can connect that to the second one''}, the worker proactively changed the view to the assembly area, providing a clear view of the connections.
\end{example}

\begin{example}[D2]\label{ex:myself}
    Upon noticing the helper's difficulty in adjusting the view using the \textit{helper-led mode}, the worker offered their assistance, saying, \textit{``You can turn on mode 3 [\textit{worker-led mode}], and I'll help you adjust the camera.''} The helper turned down the offer, stating, \textit{``Um...I think I can adjust the camera myself.''}
\end{example}

\begin{example}[D7]\label{ex:top}
    The helper attempted to move the camera to inspect the roof panel but faced challenges and eventually gave up, admitting, \textit{``Actually, I am not able to see the top panel...can you?...I need to look up to the panel.''} In response, the worker engaged the \textit{worker-led mode} to move the camera and show the roof panel.
\end{example}

\begin{example}[D4]\label{ex:before}
    The helper made a brief attempt to use the \textit{helper-led mode} to view the wall grid, a view that the helper had managed to achieve previously after considerable effort. Following the brief attempt, the helper requested the worker, \textit{``Do you mind manually moving the camera? So kind of in the same spot that we had it before?''}
\end{example}  

\begin{example}[D5]\label{ex:parts}
During the assembly process, the helper consistently relied on the worker to move the camera while looking at different aspects in the workspace, such as the organizer, components, and assembly area. When they were ready to progress to the next step in the process, the helper requested, \textit{``The parts that I had you collect from the organizer...can you show me that?''} 
\end{example}

\begin{example}[D6]\label{ex:workbench}
To access the instructions located on the side of the workbench, the helper asked the worker, \textit{``Could you just guide me towards the side of the workbench?''}
\end{example}

\begin{example}[D3]\label{ex:wall}
    In order to view the inside of a wall grid, the helper asked the worker, \textit{``Could you point to the wall so that I can see inside it?''} The helper seemed to anticipate that the camera would align with the direction of the worker's pointing gesture.
\end{example}

\begin{example}[D6]\label{ex:spatial}
    When the dyad was searching for storage areas where the required component may be located, the worker remarked, \textit{``There are some drawers over here [pointing gesture]''}.
\end{example}

\begin{example}[D3]\label{ex:align}
The helper stated, \textit{``Oh, it's looking at your hand and not what I want it to be looking at,''} when the view did not match their expectation of the camera aligning with the direction of pointing.
\end{example}

\begin{example}[D6]\label{ex:back}
The helper said, \textit{``Could you just take me back...wait I'll just take myself back,''} and reset the camera to view the assembly area.
\end{example} 

\begin{example}[D3]\label{ex:missing}
    After gathering the necessary components, the helper said, \textit{``So I'm going to reset the camera...and that's the two parts that are missing,''} and proceeded to give instructions to the worker for the assembly.
\end{example} 

\begin{example}[D4]\label{ex:plan}
    When the helper was instructed by the experimenter to begin the next step, the helper responded with, \textit{``Okay, let me reset the camera,''} and proceeded with the planning for the next step.
\end{example} 

\begin{example}[D8]\label{ex:board}
    The helper initiated the \textit{worker-led mode} by asking the worker, \textit{``Can you show me the board [wall grid] again?''} The worker moved the camera, but the view was not adequate. The helper remarked, \textit{``Okay, let me reset and come back again,''} reset the camera, and then used the \textit{helper-led mode} to view the grid. 
\end{example} 

\begin{example}[D1]\label{ex:lost}
    The helper engaged the \textit{robot-led mode} and stated, \textit{``Okay, we are in mode 2 [robot-led mode] now. Did the robot detect you?''} The worker replied, \textit{``No. You might have to reset it.''} The helper proceeded to reset the robot and engaged the \textit{robot-led mode} again. There were no further issues with the robot tracking the worker's hand.
\end{example}

\end{document}